%% file: sel_best_few.tex
\pdfoutput=1
\documentclass[mnsc]{informs4}

\RequirePackage{tgtermes}
\RequirePackage{newtxtext}
\RequirePackage{newtxmath}
\RequirePackage{bm}
\makeatletter
\DeclareFontEncoding{LS1}{}{}
\DeclareFontSubstitution{LS1}{stix}{m}{n}
\DeclareSymbolFont{STIXlargesymbols}{LS1}{stix}{m}{it}
\renewcommand{\widetilde}{\mathaccent"0\hexnumber@\symSTIXlargesymbols 9B }
\makeatother
\newlength{\DSnat}
\newlength{\DSstr}
\newlength{\DSshr}
\newlength{\DSbelow}
\renewcommand{\Dskips}[4]{%
  \setlength{\DSnat}{#1em}\setlength{\DSnat}{0.8\DSnat}%
  \setlength{\DSstr}{#2em}\setlength{\DSstr}{0.8\DSstr}%
  \setlength{\DSshr}{#3em}\setlength{\DSshr}{0.8\DSshr}%
  \setlength{\DSbelow}{#4em}\setlength{\DSbelow}{0.8\DSbelow}%
  \setlength{\abovedisplayskip}{1\DSnat plus 1\DSstr minus 1\DSshr}%
  \setlength{\abovedisplayshortskip}{0pt plus 1\DSstr}%
  \setlength{\belowdisplayshortskip}{1\DSbelow plus 1\DSstr minus 1\DSstr}%
  \setlength{\belowdisplayskip}{\abovedisplayskip}}
\OneAndAHalfSpacedXI

\usepackage{mathtools}
\usepackage{microtype}
\usepackage{graphicx}
\usepackage{amsmath,amssymb,amsfonts,mathtools}
\usepackage{array}
\usepackage{booktabs}
\usepackage{siunitx}
\usepackage{enumitem}
\usepackage{algorithm}
\usepackage[noend]{algpseudocode}
\usepackage{etoolbox}
\AtBeginEnvironment{algorithm}{\SingleSpacedXI}
\AtBeginEnvironment{algorithmic}{\footnotesize}

\algrenewcommand{\algorithmicdo}{}
\algrenewcommand{\algorithmicthen}{}
\usepackage{tikz}
\usetikzlibrary{arrows.meta,positioning}
\usepackage{natbib}
 \bibpunct[, ]{(}{)}{,}{a}{}{,}%
 \def\bibfont{\small}%
\usepackage{xurl}
\usepackage[hidelinks]{hyperref}
\usepackage[capitalise]{cleveref}
\usepackage{etoc}

\EquationsNumberedThrough
\TheoremsNumberedThrough
\def\theARTICLETOPLEFT{}
\def\theARTICLETOPRIGHT{}
\makeatletter
\def\theARTICLEABSTRACT{%
  \vspace*{18pt}%
  \begin{minipage}[t]{\textwidth}\parindent1em\ABSfont
    \noindent\theABSTRACT\endgraf
    \vskip5pt
    \theKEYWORDS
    \noindent\hrulefill
  \end{minipage}%
  \vspace*{0pt}%
}
\makeatother

\let\AppendixHeadSize=\normalsize
\let\AppendixFontSize=\normalsize
\let\AppendixSectionHeadSize=\normalsize

\makeatletter
\def\theRRHFirstLine{}%
\def\theLRHFirstLine{}%
\def\theRRHSecondLine{}%
\def\theLRHSecondLine{}%
\makeatother

\def\ExampleTextFont{\thit}
\theoremstyle{TH}%
\newtheorem{approximation}{Approximation}
\crefname{assumption}{Assumption}{Assumptions}
\Crefname{assumption}{Assumption}{Assumptions}
\crefname{approximation}{Approximation}{Approximations}
\Crefname{approximation}{Approximation}{Approximations}
\crefname{section}{\S}{\S\S}
\Crefname{section}{\S}{\S\S}
\crefformat{section}{\S#2#1#3}
\Crefformat{section}{\S#2#1#3}
\crefformat{subsection}{\S#2#1#3}
\Crefformat{subsection}{\S#2#1#3}
\crefformat{subsubsection}{\S#2#1#3}
\Crefformat{subsubsection}{\S#2#1#3}

\AtBeginEnvironment{APPENDICES}{%
  \numberwithin{equation}{section}%
  \numberwithin{theorem}{section}%
  \numberwithin{lemma}{section}%
  \numberwithin{proposition}{section}%
  \numberwithin{corollary}{section}%
  \numberwithin{approximation}{section}%
  \numberwithin{assumption}{section}%
  \numberwithin{remark}{section}%
  \numberwithin{algorithm}{section}%
  \numberwithin{figure}{section}%
  \numberwithin{table}{section}%
}

\newcommand{\R}{\mathbb{R}}
\newcommand{\E}{\mathbb{E}}
\newcommand{\Pp}{\mathbb{P}}
\newcommand{\Var}{\mathbb{V}}
\newcommand{\Cov}{\mathrm{Cov}}
\newcommand{\diag}{\mathrm{diag}}
\DeclareFontFamily{U}{dsrom}{}
\DeclareFontShape{U}{dsrom}{m}{n}{<-> dsrom10}{}
\newcommand{\1}{\mathord{\text{\usefont{U}{dsrom}{m}{n}\selectfont 1}}}

\newsavebox{\FDbox}
\newlength{\FDhalf}
\newlength{\FDshift}
\newlength{\FDup}
\newlength{\FDdown}
\NewDocumentCommand{\FitDelims}{mmm}{\mathinner{\mathpalette\FitDelimsBody{{#1}{#2}{#3}}}}
\NewDocumentCommand{\FitParens}{m}{\FitDelims{(}{)}{#1}}
\newcommand{\FDmidplain}[1]{\mathord{#1}}
\let\FitMid\FDmidplain
\newcommand{\FitDelimsBody}[2]{\FitDelimsDo#1#2}
\newcommand{\FitDelimsDo}[4]{%
  \sbox{\FDbox}{$\mathsurround=0pt \let\FitMid\FDmidplain #1#4$}%
  \setlength{\FDhalf}{\ht\FDbox}%
  \addtolength{\FDhalf}{\dp\FDbox}%
  \divide\FDhalf by 2
  \setlength{\FDup}{\FDhalf}%
  \addtolength{\FDup}{\fontdimen22\textfont2}%
  \setlength{\FDdown}{\FDhalf}%
  \addtolength{\FDdown}{-\fontdimen22\textfont2}%
  \setlength{\FDshift}{\ht\FDbox}%
  \addtolength{\FDshift}{-\FDup}%
  \renewcommand{\FitMid}[1]{\FDdelim{\left##1}{\right.}}%
  \mathopen{\FDdelim{\left#2}{\right.}}#4\mathclose{\FDdelim{\left.}{\right#3}}}
\newcommand{\FDdelim}[2]{%
  \raisebox{\FDshift}{$\delimiterfactor=1000 \delimitershortfall=0pt \nulldelimiterspace=0pt #1\vrule height\FDup depth\FDdown width 0pt#2$}}

\RUNAUTHOR{Goldberger and Rudi}
\RUNTITLE{Selecting the Best Few with Human Evaluators}
\def\TITLEfont{\fs.17.21.\rm}
\TITLE{Selecting the Best Few with Human Evaluators: Adaptively Combining Ratings and Pairwise Comparisons}
\ARTICLEAUTHORS{%
\AUTHOR{Motti Goldberger}
\AFF{Yale School of Management, \EMAIL{motti.goldberger@yale.edu}}
\AUTHOR{Nils Rudi}
\AFF{Yale School of Management, \EMAIL{nils.rudi@yale.edu}}
}

\ABSTRACT{This paper considers a decision-maker tasked with identifying the $k$ best items from a large pool of candidates, such as job applicants, LLMs, award nominees, or new product ideas. The decision-maker learns about the items' quality by soliciting human evaluations. Each evaluation can be costly, so the decision-maker aims to use the limited evaluation budget efficiently by allocating evaluations adaptively. A key consideration is the form of evaluation to solicit, which should be something evaluators are comfortable providing. Two popular forms are ratings and pairwise comparisons, and we allow for both, as they may be especially useful in combination given the limited budget. Ratings can be used early on to get a coarse picture of which items are good and which are not. As that picture forms, comparisons can refine the ordering of the good items, narrowing in on the ultimate top-$k$ selection. To solve the allocation problem, we develop an adaptive algorithm built from two allocation policies. The first, our primary contribution, is based on a surrogate objective that is naturally forward-looking, valuing evaluations that help inform what to evaluate in future rounds, not just those that directly improve the selection decision. The decision rule is interpretable, and we prove that it is asymptotically consistent. Late in the evaluation process, when being forward-looking is less productive, we construct a myopic policy and a criterion for switching to it. In simulation, the method gives insight into how to combine these two types of evaluations, and beats the (limited) existing benchmarks.} 

\newif\ifHalmosDone
\let\origendproof\endproof
\def\endproof{\ifHalmosDone\else\Halmos\fi\global\HalmosDonefalse\origendproof}
\newcommand{\Halmoshere}{\tag*{$\square$}\global\HalmosDonetrue}
\newcommand{\noHalmos}{\global\HalmosDonetrue}

\begin{document}
\etocdepthtag.toc{main}
\maketitle

\section{Introduction}\label{sec:intro}

This paper considers a decision-maker tasked with identifying the $k$ best items from a large pool of candidates. This is done using data the decision-maker sources sequentially from human evaluators, where the adaptive choice of upcoming evaluations is informed by what has been learned so far. When modeling such human judgment, a key decision is the form of the evaluation input. While one may wish for some sophisticated multidimensional input, providing it is often beyond the scope, time, and ability for evaluators, which is why inputs such as thumbs-up/thumbs-down and swiping left or right have become popular. Rather than restrict the decision-maker to soliciting one type of input, we allow for the combination of thumbs-up/thumbs-down ratings and pairwise comparisons, which, as we will discuss, can be complementary in our problem. This model can form a basis for many applications, including selecting the best few among LLMs, job applicants, university applicants, financial analysts, award nominees, promotion candidates, new product ideas, and sports draft prospects.

While problems of this type are clearly important, they also tend to be difficult in two distinct ways. One challenge is the \emph{analysis and algorithm design}. The aim is a sequential evaluation allocation policy that is both close to optimal and structured enough to yield insight and earn practitioners' trust. The other difficulty is the \emph{modeling}, which connects the application to the analysis and sets the stage for how useful that analysis will be. First, we position the problem in the literature and walk through the modeling decisions involved. We then discuss how we arrived at our solution approach.

The problem of identifying the best few items by allocating evaluation resources is central to the \emph{ranking and selection} and \emph{best-arm identification} literatures, where what we refer to as evaluations are typically called samples. Unlike the classical work on multi-armed bandits \citep{Robbins1952}, where rewards are collected while learning, here the only concern is with the quality of the terminal selection, the setting known as \emph{pure exploration} \citep{BubeckMunosStoltz2009}. Much of this literature focuses on identifying the best (top-$1$) item \citep{ChenLin2000,FrazierPowell2008,ChenRyzhov2023}. We consider the more general problem of top-$k$ identification \citep{ChenHeFuLee2008,ZhangPengZhangZhou2023,Liu2024}.

In this literature, evaluating (sampling) an item typically means directly observing a noisy realization of its utility (quality), which we will refer to as untransformed evaluations. Such untransformed input is natural in applications such as simulation optimization and hyperparameter tuning, where the observed output is itself the object being optimized. In our setting, the utility of each item is latent, since the value of including each item in the chosen set is never evaluated directly. Instead, each item's quality is observed indirectly through evaluations that take the form of ratings or pairwise comparisons, both of which transform the latent utilities into coarse ordinal outcomes.

To highlight why we do not rely on adapting an established algorithm, we highlight three common modeling and analysis decisions in the ranking and selection literature. We discuss how each decision relates to our problem and, when relevant, how we overcome the additional challenges posed by differences in our setting.

The first modeling decision concerns the terminal objective: how we grade an evaluation allocation policy based on the $k$ items it ultimately selects. The two most common choices are the probability of correct selection (PCS) \citep{ChenLin2000,GlynnJuneja2004,li2025greedy} and the expected opportunity cost (EOC) \citep{ChickInoue2001,ChickBrankeSchmidt2010,GaoChen2015}. PCS is the probability that the selected set is exactly the true top-$k$, so it treats every incorrect selection equally. EOC is the expected shortfall in utility between the selected set and the true top-$k$ set, making bigger selection mistakes more costly. PCS suits highly specialized or high-stakes decisions where any deviation from the best possible selection is considered a failure. For more common tasks, where a serious mistake can be far costlier than a marginal one, EOC can be the more natural fit. Since EOC is more aligned with the applications we consider, we adopt it in this work.

A second decision concerns how the allocation problem is formulated, with two common formulations at opposite ends of the spectrum. One is static allocation, where the decision variable is the total number of evaluations assigned to each item. The other is dynamic learning, where evaluations are chosen one at a time and each outcome is observed before the next evaluation is chosen. In the static formulation, writing $\boldsymbol{N}=(N_1,\ldots,N_I)$ for the vector of evaluations assigned to items $i=1,\ldots,I$ and $R$ for the total evaluation budget, the allocation problem is
\begin{equation*}
    \min_{\boldsymbol{N}}\ \mathrm{EOC}(\boldsymbol{N})
    \qquad\text{s.t.}\qquad
    \sum_{i=1}^{I} N_i \leq R .
\end{equation*}
This allocation problem faces two main difficulties. The objective $\mathrm{EOC}(\boldsymbol{N})$, and likewise $\mathrm{PCS}(\boldsymbol{N})$, generally does not have a closed form that is amenable to analysis. Additionally, the problem is an integer program, which typically makes it intractable to consider the full solution space at scale. In the fully dynamic formulation, one evaluation is chosen at a time, avoiding the combinatorial complexity of the static problem. At each decision point, the state consists of the current posterior distribution over the items' utilities and the remaining budget, and the action is the next evaluation. The state transition updates the posterior based on the outcome and deducts the evaluation's cost from the remaining budget. This problem is formulated as a dynamic program which is intractable because the combination of possible evaluations, outcomes, transitions, and future states grows rapidly with the horizon \citep{FrazierPowell2008,HongFanLuo2021,ZhangPengZhangZhou2023}.

Our base formulation is fully dynamic with evaluations chosen sequentially. However, allocating one evaluation at a time can have logistical obstacles. There may be considerable overhead in assigning and conducting evaluations, which can favor giving an evaluator a \emph{batch} of tasks at once. We therefore also consider a hybrid approach in \cref{sec:batching}, where a batch of evaluations is chosen in each round. This sits between the two formulations and inherits both difficulties. Within a round, the allocation is an integer program, each linked through the between-round dynamics.

The complexity of the problem under both formulations has given rise to a variety of solution approaches. Before turning to our own, we discuss the three most relevant well-studied families, all developed in the untransformed evaluation setting. The first is a family of algorithms known as Optimal Computing Budget Allocation (OCBA), originating with \citet{ChenChen1997,ChenLin2000,GlynnJuneja2004}. At a high level, the algorithm loops through three steps: (i) take the items' current mean estimates as if they were the ground truth utilities, (ii) choose the next evaluation(s) by determining the static allocation that would maximize PCS (or minimize EOC), and (iii) update the estimates of the means based on the outcome. However, step (ii) is intractable due to the complexities of the static allocation problem discussed above. To overcome this, it is common to consider the asymptotic regime where $R \to \infty$ and find the allocation proportions that (approximately) optimize the resulting problem. Eventually, the estimated means will converge to the true means, and thus the evaluation proportions will converge to the (approximately) optimal proportions. The majority of the OCBA literature studies the top-$1$ setting with PCS, with extensions to top-$1$ EOC \citep{HeChick2007,GaoChen2017}, as well as to top-$k$ PCS \citep{ChenHeFuLee2008,GaoChen2015a,ZhangLee2016} and top-$k$ EOC \citep{GaoChen2015}. Such approaches tend to perform better in large-budget settings. In fact, for small budgets, it has been demonstrated that such allocation procedures may even perform worse than equal allocation \citep{PengChenFuHu2018,Liu2024}.

A second family of algorithms considers the dynamic formulation restricted to one-step look-ahead policies, see \citet{PowellRyzhov2012} for an overview. At each round, the decision-maker chooses the evaluation that (approximately) maximizes the expected value of information (EVI), the improvement in the quality of the terminal selection that results from one additional evaluation. For top-$1$ identification, \citet{ChickInoue2001,ChickBrankeSchmidt2010} derive policies that approximately maximize EVI for both the PCS and EOC objectives. Recently, \citet{ZhangPengZhangZhou2023} developed a top-$k$ identification policy that maximizes an approximation of the EVI for PCS. Assuming evaluations follow Gaussian distributions with known variances, \citet{FrazierPowell2008} derive the exact one-step look-ahead policy that maximizes the EVI for EOC, known as the knowledge gradient (KG) policy. In our setting, however, KG's myopia can be particularly detrimental with binary evaluations \citep{GrovesBranke2019,LiYucesan2024}. We return to this point in \cref{subsec:kgdiff}.

A third related family of algorithms is based on the expected improvement (EI) approach \citep{JonesSchonlauWelch1998}, which originated in the Bayesian Optimization literature. Various works have studied its application to top-$1$ identification \citep{Ryzhov2016,QinKlabjan2017,ChenRyzhov2019}. In its simplest form, EI scores each item by how much its utility, in expectation, exceeds that of the current best candidate and evaluates the item with the highest score. EI's simplicity is appealing, and although it does not explicitly look ahead, it still reflects future dynamics: evaluating items with high potential to improve on the current best can help inform which evaluations to conduct in future rounds. However, EI does not naturally extend to reflect alternative evaluation forms with different information contents and costs. Furthermore, it chooses which \emph{single} item to evaluate next, whereas our evaluation alternatives include choosing a \emph{pair} to compare.

Finally, the third modeling decision concerns the form of the evaluations themselves. In psychometrics, the field of measurement of psychological data, the two most common forms of input for human feedback are ratings, where an evaluation assigns a categorical score to a particular item, and rankings, where a set of items is ordered relative to one another \citep{Thurstone1927,Likert1932,Torgerson1958Scaling}. We consider each in its parsimonious form---binary thumbs-up/thumbs-down ratings and pairwise comparisons---both among the most common input choices. Restricting to these simple versions of ordinal feedback keeps evaluators' cognitive load low \citep{RevillaSarisKrosnick2014Categories,KutscherEtAl2024RatingScales}. The two input forms have distinct pros and cons in our setting. Ratings can be cheaper and easier to provide than pairwise comparisons \citep{WhiteEtAl2024RatingBasedRL}, and a few of them can be an efficient way to establish the ballpark utility of an item. However, ratings are not efficient at distinguishing the relative order of close items, which is essential for differentiating between top-$k$ candidates. Additionally, a rating (say, a thumbs-up) can carry a different meaning to different evaluators, resulting in potential ambiguity when aggregating and interpreting them. Pairwise comparisons, in contrast, only contain \emph{relative} information about item utilities, and while their interpretation is unambiguous, their usefulness depends heavily on which items are paired. The number of possible pairings is quadratic in the number of items, so starting from an uninformed prior would result in many unhelpful pairings early on. For example, we might compare the items that turn out to be the two worst. Even if that comparison provides a strong signal about which of the two is better, it does little to inform us about which items are in the top-$k$. However, once we do have a clear sense of which items are top-$k$ contenders, comparisons can be far more effective than ratings at pinning down their relative ordering \citep{LiZhangLi2018RatingRankingTopk}. It follows that the two evaluation forms can be especially useful when combined \citep{FageotEtAl2026}. Intuitively, ratings can be used more in early rounds to get a coarse picture of which items are ``good'' and which are not, and then as that picture forms, comparisons can gradually take over, refining the relative ordering of the good items and narrowing in on the ultimate top-$k$ selection. Our model allows the two evaluation types to differ in accuracy and cost (effort), following the empirical findings of \citet{PerezOrtizMantiuk2019UnifiedQualityScale} and \citet{WhiteEtAl2024RatingBasedRL}.

The two evaluation forms can also interact in non-intuitive ways. Consider a simple example with four items, $A$, $B$, $C$, and $D$, whose utilities are independent ex-ante. With ratings alone, a rating of $B$ informs its utility relative to the other items but leaves our beliefs about their utilities unchanged. Now suppose we observe the comparisons $A \succ B$ and $C \succ D$, followed by $B \succ C$. Intuitively, the third comparison suggests that $A$ is preferred to $D$ even though neither was involved in the comparison. The mechanism by which our model reflects this is that comparisons create dependency in the joint posterior over the utilities: the third comparison raises $B$'s estimate and thereby also lifts $A$, while lowering $C$'s estimate and thereby also lowering $D$. This dependency also changes what we learn from a rating. A subsequent rating of $B$ can now influence our posterior beliefs about $A$, $C$, and $D$. In fact, with a more complex comparison history, a rating of $B$ can even reverse our estimated relative ranking of two non-evaluated items, such as $A$ and $D$. Thus, the value of an evaluation for informing the top-$k$ set may not be apparent from what it tells us about the evaluated items alone. Including such dependency in the joint posterior is an opportunity, as it leads to better discrimination between items and hence a more accurate top-$k$ choice. But this added accuracy does not come for free: reflecting the dependency in the dynamic choice of evaluations adds complexity.

Work related to top-$k$ selection tasks in which evaluations are, like ours, binary pairwise comparisons (and possibly also ratings) is scarce in the literature, especially with small budgets. In this setting, one common approach to model evaluations is to assign each possible \emph{evaluation} its own parameter. For pairwise comparisons, this means a separate parameter specification for each pair of items, representing the probability that one item is preferred to the other. Although the outcomes are binary, this model is still in the untransformed evaluation setting: the outcomes remain a direct noisy realization of the underlying parameter. An alternative approach is the latent-utility model of our setting, where each \emph{item} has a single value, its utility, that generates the comparisons (and ratings).

In the ranking and selection literature, \citet{PriekuleMeisel2017,GrovesBranke2019,LiYucesan2024} study top-$1$ and top-$k$ selection with binary pairwise comparisons under the first evaluation model. Their methodology does not transfer to our problem, as their more flexible model requires learning the parameters of $I(I\!-\!1)/2$ item pairs, which is unrealistic in small-budget settings. More fundamentally, they only consider pairwise comparisons, and it is not clear how one could combine ratings and comparisons under this model.

In the crowdsourcing literature, these problems are more commonly studied under latent-utility models. \citet{chen2016crowdranking} develop an approximate KG algorithm for learning the full ranking of items (as opposed to the top-$k$) from pairwise comparisons. \citet{LiZhangLi2018RatingRankingTopk,PerezOrtizMantiuk2019UnifiedQualityScale,FageotEtAl2026} develop a unified statistical model for combining pairwise comparisons and ratings. \citet{FageotEtAl2026} show empirically that combining ratings and pairwise comparisons can improve estimation of the latent utilities. More directly related to our problem, \citet{LiZhangLi2018RatingRankingTopk} also give a heuristic approach for choosing evaluations for top-$k$ selection, a slightly modified version of which we use as a benchmark.

Finally, top-$k$ identification from pairwise comparisons is well studied in the dueling-bandit and active-ranking literature. Usually set in the fixed-confidence regime where sampling continues until a target confidence is reached (as opposed to a fixed budget $R$), these algorithms tend to require large numbers of evaluations---far beyond the budgets we consider. See, e.g., \citet{Mohajer2017,Ren2018,ChenLiMao2018,heckel2019activeranking,SahaGopalan2019,Ren2020,SahaGopalan2020,GoldbergerRudi2026} for recent developments and \citet{Bengs2021JMLR} for a review.

Instead of explaining our evaluation allocation approach directly, we find it easier to lay out where it came from. Our starting point was the KG policy primarily because, among the existing methods, it is well suited for small-budget problems and can naturally accommodate the combination of evaluation types---our key initial motivation for undertaking this research. However, the KG policy allocates evaluations in each round as if the final selection would immediately follow. Except for the last evaluation, that is not the actual task. The task is to prepare for the terminal selection, and it may be that the best evaluation to do next is decided by how it may inform later evaluation choices. To illustrate how much KG misses out on future dynamics, consider that in its eyes any evaluation whose outcome cannot change the current top-$k$ set has no value at all.

Ideally, we would incorporate the full dynamics into the evaluation choices, but based on the existing literature, anything beyond solving a sequence of state-dependent, static problems seems out of reach. Our key insight leading to our approach came from flipping the perspective from the positive focus on what the decision-maker wants to achieve---choosing the best $k$ items---to a negative perspective---how to reduce errors. There are two ways to err ex-post: including an item that does not belong (\emph{inclusion error}) and excluding one that does (\emph{exclusion error}). Ex-ante, items have an expected \emph{mismatch loss}: the expected magnitude of an inclusion error for items in the current top-$k$ and an exclusion error for those not in it. We then construct a per-round \emph{surrogate} objective that values uncertainty reduction by how much it reduces the expected mismatch loss. We contrast this with the KG policy that adopts the terminal top-$k$ EOC as each round's objective. It uses the current uncertainty to assess possible changes in the membership of the top-$k$ \emph{posterior means}. Our method instead holds the current means fixed and values an evaluation by how much it is expected to reduce \emph{uncertainty}. The terminal top-$k$ decision then drives the evaluation choices only implicitly by providing a measure of where uncertainty reduction is more valuable. This surrogate is naturally forward-looking. It seeks to reduce uncertainty about the items currently near the top-$k$ cutoff without requiring any direct change to the current top-$k$ set. Doing so pays off in future rounds, since the more effective we are at singling out items near the top-$k$ cutoff, the more effectively we can direct the remaining evaluations.

The difficulty is that mismatch loss does not admit an analysis-friendly expression. The magnitude of an item's error depends on where it stands relative to the $k$th or $(\!k\!+\!1\!)$th best of the remaining items, so computing an item's mismatch loss requires an expectation with respect to the joint utility distribution and an order statistic of the other $I\!-\!1$ items, where even the latter itself is intractable. Our resolution is to replace this random top-$k$ threshold with a simpler threshold that is recalibrated each round, such that, in expectation, $k$ items' utilities will exceed it. With this threshold, an item's mismatch loss depends only on the marginal distribution of its utility relative to it. This separability gives a clean and interpretable way to account for an evaluation's effect on \emph{all} items, those being directly evaluated and, via covariance, those that are not. The sum of these mismatch losses is our surrogate objective, and in each round, the policy chooses the evaluation whose reduction of this loss per evaluation cost is largest.

We also point out a potential trap that arises when combining the two evaluation types, in part because we initially fell into it. Ratings contain information about the average utility of the items, while comparisons do not. Learning about this average reduces the variance of each item's utility but does not help distinguish the top-$k$, since shifting all utilities together leaves their ordering unchanged. A surrogate that rewards this reduction in variance would therefore unfairly favor ratings over comparisons. We discuss this issue in \cref{sec:surrogate} and show that our surrogate does not reward learning about the average utility, so ratings and comparisons are on equal footing.

As the horizon approaches its end, being forward-looking becomes less valuable, and this strength of the surrogate can become a weakness. In this regime, the myopia of the KG policy would be desirable. We therefore develop the KG policy for our problem, with a switching criterion for the \emph{endgame}, the final rounds where its myopia outperforms the surrogate.

We support our method through both analysis and simulation. In the analysis, although the surrogate objective was constructed from analytical intuition rather than derived, we show that it can be obtained from an upper bound on the EOC, which rigorously justifies and refines our choice of the top-$k$ threshold. We also prove that the resulting policy that allocates by minimizing the surrogate in each round is asymptotically consistent. In simulation, we show that the allocations this policy produces are intuitive and insightful, and that it substantially outperforms existing benchmarks.

The remainder of the paper is organized as follows: \Cref{sec:model} presents the model and defines the objective. \Cref{sec:surrogate} develops the surrogate objective, which \Cref{sec:greedy} uses to construct the sequential evaluation procedure and establish its asymptotic consistency. \Cref{sec:endgame} develops the knowledge gradient policy for the endgame and the switching rule. \Cref{sec:batching} considers the batch setting. \Cref{sec:numerics} reports the results of the numerical experiments, and \Cref{sec:conc} gives concluding remarks.

\section{Model and objective}\label{sec:model}
This section defines the generative data model for pairwise comparisons and ratings, introduces the sequential evaluation process and formulates the optimization problem as a dynamic program. We then demonstrate that the KG policy is not well-suited for our setting, which motivates the surrogate objective of \cref{sec:surrogate}.

\subsection{Generative model}
\label{sec:model:gen}
We consider $I$ items with latent utilities $\boldsymbol{\theta}=(\theta_1,\ldots,\theta_I)\in\R^I$, which represent the values of selecting each item. We adopt a Bayesian framework and place a Gaussian prior on the utility vector, $\boldsymbol{\theta}\;\sim\;N(\boldsymbol{\mu}_0,\,\Sigma_0)$ where $\Sigma_0$ is positive definite. When preliminary information distinguishes the items, $\boldsymbol{\mu}_0$ and~$\Sigma_0$ can reflect it.  When no such information is available, or when items are people and a prior that discriminates raises ethical or legal concerns, an exchangeable prior ($\boldsymbol{\mu}_0=\boldsymbol{0}$, $\Sigma_0=s\operatorname{Id}_I$) treats all items symmetrically.

Evaluators can provide two types of feedback, \emph{pairwise comparisons} and \emph{ratings}. A \emph{pairwise comparison} of items $i$ and $j$ produces a binary outcome: one item is preferred to the other. The event that an evaluator prefers $i$ to $j$ is denoted by $i \succ j$, and its probability is determined by $\theta_i - \theta_j + \epsilon_1$, where $\epsilon_1 \sim N(0,\sigma_1^2)$. Thus, the evaluator's ability to distinguish between the two items depends on the difference in their utilities and on $\epsilon_1$, which captures imperfections in pairwise evaluations. The subscript $1$ is used because item $i$ is compared to $1$ other item. A \emph{rating} of item $i$ produces a binary `thumbs-up' or `thumbs-down' outcome, where item $i$ is implicitly compared to some utility threshold $\theta_0$. Note that $\theta_0$ is not a part of $\boldsymbol{\theta}$---we use this notation to facilitate unified expressions for pairwise comparisons and ratings. Similar to the pairwise comparison model, the event that an evaluator ranks $i$ above the threshold is denoted by $i \succ 0$ and its probability is determined by $\theta_i - \theta_0 + \epsilon_0$, where $\epsilon_0 \sim N(0,\sigma_0^2)$. Here $\epsilon_0$ captures two sources of noise: the evaluator's ability to estimate item $i$'s utility and the uncertainty about where the latent threshold $\theta_0$ lies. The subscript $0$ is used because item $i$ is compared to $0$ items.

The set of possible evaluations is given by $\mathcal I := \FitDelims{\{}{\}}{(i,j) : 0 \leq j < i \leq I }.$ Define $\jmath = \min(j,1)$, which is $0 \ (1)$ when an evaluation is a rating (comparison). For an evaluation $(i,j) \in \mathcal I$, let $X_{ij}$ denote the outcome where
\begin{equation*}
X_{ij} = \begin{cases}
1 & \text{if $i \succ j$}\\
-1 & \text{if $i \prec j$,}
\end{cases}
\end{equation*}
and, conditioned on the evaluator-noise term $\epsilon_\jmath$, $\Pp(X_{ij} = 1 \mid \boldsymbol{\theta},\epsilon_\jmath) = g(\theta_i - \theta_j +\epsilon_\jmath),$ where $g:\R\to(0,1)$ is some link function. For the remainder of the paper, unless specified otherwise, we consider the probit link function, commonly used to model human ordinal rating data \citep{LiddellKruschke2018OrdinalProbit}, i.e., $g = \Phi$, where $\Phi$ is the standard normal CDF and $\phi$ its density. Thus,
\begin{equation*}
\Pp(X_{ij} = 1 \mid \boldsymbol{\theta}) = \E_{\epsilon_\jmath}\!\FitDelims{[}{]}{\Phi\FitParens{\theta_i - \theta_j + \epsilon_\jmath }}
=  \Phi\FitParens{\frac{1}{\sqrt{1 + \sigma_{\jmath}^2}}(\theta_i - \theta_j)}
= \Phi\FitParens{\beta_\jmath(\theta_i - \theta_j)},
\end{equation*}
where $\beta_\jmath := \frac{1}{\sqrt{1 + \sigma_\jmath^2}} \in (0,1]$, for $\jmath \in \{0,1\}$, captures the evaluator precision when providing a rating or a pairwise comparison. As shown in \citet{PerezOrtizMantiuk2019UnifiedQualityScale}, a single latent-utility model for ratings and pairwise comparisons can fit human data well as long as the two evaluation types are allowed to differ in signal quality, which we capture through $\beta_0$ and $\beta_1$. In addition to variation in signal quality, there may also be heterogeneity in evaluator accuracy and bias. If these differences were known, our model would capture this by using different values of $\theta_0$, $\beta_0$, and $\beta_1$ for each evaluator. Because they are generally unknown, we instead use common parameters across evaluators, effectively absorbing evaluator-level variation into the noise terms rather than learning evaluator-specific accuracy or bias from past evaluations. This is a standard simplification in the literature, and models using it have been shown to still fit human data well \citep{PerezOrtizMantiuk2019UnifiedQualityScale}. Learning evaluator-specific parameters could further inform the allocation, but would introduce an additional inference problem that we leave for future work. In the numerical experiments of \cref{sec:numerics}, we stress-test this assumption by using synthetic data where accuracy and bias differ across evaluators.

One may question the feasibility of calibrating these parameters in practice. \Cref{rem:calibration} shows that after fixing one for identification, the rest can be calibrated from three simple, interpretable quantities.
\begin{remark}\label{rem:calibration}
The binary data alone cannot identify the latent utility scale and the observation accuracy parameters simultaneously: for any $c>0$, replacing $(\boldsymbol{\theta},\theta_0,\beta_0,\beta_1)$ with $(c\boldsymbol{\theta},c\theta_0,\beta_0/c,\beta_1/c)$ leaves every outcome probability $\Phi\!\FitParens{\beta_\jmath(\theta_i-\theta_j)}$ unchanged. Thus, one parameter can (and must) be fixed, and we fix the scale of $\boldsymbol{\theta}$ by choosing the prior variance $s$. The remaining parameters $\beta_1,\beta_0,\theta_0$ can then be found given the following more interpretable quantities, which we formally derive in the Appendix:
\begin{enumerate}[label=(\roman*)]
    \item The probability that a comparison of two randomly chosen items agrees with the true ordering,
    \vspace{-3pt}
    \begin{equation*}
    \frac12 + \frac{1}{\pi} \arcsin\!\FitParens{ \frac{\beta_1\sqrt{2s}}{\sqrt{\cramped{1+2s\beta_1^2}}} },
    \end{equation*}
    \vspace{-15pt}
    
    \noindent which determines $\beta_1$.
    \item The probability that a rating of a randomly chosen item is correct,
    \vspace{-3pt}
    \begin{equation*}
    1-\Phi\!\FitParens{\frac{\theta_0}{\sqrt{s}}}
    -\Phi\!\FitParens{\frac{\beta_0\theta_0}{\sqrt{1+s\beta_0^2}}}
    +2\,\Phi_2\!\FitParens{
    \frac{\theta_0}{\sqrt{s}},
    \frac{\beta_0\theta_0}{\sqrt{1+s\beta_0^2}};
    \frac{\beta_0\sqrt{s}}{\sqrt{1+s\beta_0^2}}
    },
    \end{equation*}
    \vspace{-15pt}
    
    \noindent where $\Phi_2(\cdot,\cdot;\rho)$ is the standard bivariate normal CDF with correlation $\rho$.
    \item The expected proportion of items that receive a `thumbs-up' rating,
    \vspace{-3pt}
    \begin{equation*}
    \Phi\!\FitParens{-\frac{\beta_0\theta_0}{\sqrt{1+s\beta_0^2}}},
    \end{equation*}
    \vspace{-15pt}
    
    \noindent which together with (ii) jointly determines $\beta_0$ and $\theta_0$.
\end{enumerate}
\end{remark}
\subsection{Sequential evaluation process}
\label{sec:model:seq}
Building on the generative model, we next describe how the sequential decision process unfolds. In each round $t=1,2,\ldots$, the decision-maker chooses an evaluation $(i_t,j_t)\in\mathcal I$, a rating or a pairwise comparison, and observes its outcome $X_t\in\{-1,1\}$ before choosing the next evaluation. Let $r_0>0$ and $r_1>0$ be the per-evaluation resource cost of ratings and comparisons, respectively, where the resource could be monetary or capacity (time it takes to conduct an evaluation). Let $R$ denote the total evaluation budget and $R_t$ the budget left at the start of round $t$, so $R_1=R$. An evaluation $(i,j)$ is affordable in round $t$ when $r_\jmath\le R_t$. After conducting the chosen evaluation, its cost is deducted from the remaining budget, $R_{t+1}=R_t-r_{\jmath_t},$ where $\jmath_t=\min(j_t,1)$. The process continues until no evaluation is affordable. Let $T$ denote the (random) total number of evaluations conducted, so $ T:=\min\FitDelims{\{}{\}}{t\ge 0:R_{t+1}<\min(r_0,r_1)}$.

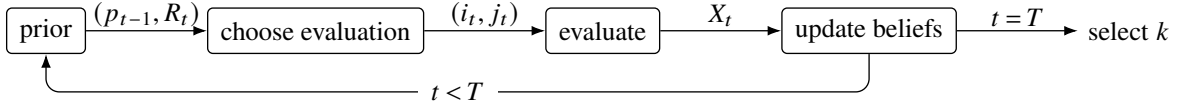
\begin{figure}[!htb]
\centering
\begin{tikzpicture}[
    box/.style={rectangle, draw, rounded corners=2pt, line width=0.5pt,
                minimum height=0.6cm, align=center, font=\small,
                inner xsep=5pt, inner ysep=2pt},
    endpoint/.style={font=\small, align=center},
    arr/.style={-{Latex[length=1.8mm, width=1.3mm]}, line width=0.5pt},
]
    \node[box] (prior) {prior};
    \node[box, right=1.6cm of prior] (choice) {choose evaluation};
    \node[box, right=1.6cm of choice] (eval) {evaluate};
    \node[box, right=1.6cm of eval] (update) {update beliefs};
    \node[endpoint, right=1.6cm of update] (select) {select~$k$};

    \draw[arr] (prior) -- (choice) node[font=\small, above, midway, inner ysep=0pt,yshift=1.5pt] {$(p_{t-1},R_t)$};
    \draw[arr] (choice) -- (eval) node[font=\small, above, midway, inner ysep=0pt,yshift=1.5pt] {$(i_t,j_t)$};
    \draw[arr] (eval) -- (update) node[font=\small, above, midway, inner ysep=0pt,yshift=1.5pt] {$X_t$};
    \draw[arr] (update) -- (select) node[font=\small, above, midway, inner ysep=0pt,yshift=1.5pt] {$t=T$};

    \draw[arr, rounded corners=5pt] (update.south) -- ++(0,-0.5) coordinate (loop-r) -| (prior.south);

    \path (loop-r) -- (loop-r -| prior.south) node[midway, font=\small, fill=white, inner xsep=5pt, inner ysep=1pt] {$t<T$};
\end{tikzpicture}
\caption{The sequential evaluation process.}
\label{fig:process}
\end{figure}

In round $t$, after the outcome is observed, we update the posterior
according to
\vspace{-4pt}
\begin{equation*}
    p_t(\boldsymbol{\theta})
    \propto
    p_{t-1}(\boldsymbol{\theta})
    \Phi\!\FitParens{X_t\beta_{\jmath_t}(\theta_{i_t}-\theta_{j_t})}
    \propto
    p_0(\boldsymbol{\theta})
    \prod_{s=1}^{t}
    \Phi\!\FitParens{X_s\beta_{\jmath_s}(\theta_{i_s}-\theta_{j_s})},
\end{equation*}
\vspace{-2pt}
where $p_0$ is the prior of $\boldsymbol{\theta}$ at $t=0$. The posterior after any round $t$ is denoted by $p_{t}$ and hence its prior is the posterior of the previous round, $p_{t-1}$. \Cref{fig:process} illustrates the sequential process.
\vspace{-4pt}
\vspace{-16pt} 
Since the Gaussian prior is not conjugate to the probit likelihood\footnote{Exact conjugacy is available through the unified skew-normal family, but computing its moments involves multivariate Gaussian CDFs whose dimension grows with the data \citep{durante2019conjugate}.}, the posterior after each round is generally non-Gaussian. One option is to carry out posterior inference by Markov Chain Monte Carlo (MCMC). However, MCMC yields posterior samples rather than a posterior with a functional form, and to formulate the round-$t$ evaluation allocation problem we rely on a closed-form approximation of $p_{t-1}$. It is therefore natural to use an approximate inference method such as expectation propagation \citep{Minka2001EP,herbrich2007trueskill}, Laplace approximation \citep{tierney1986accurate}, or variational inference (VI) \citep{jordan1999introduction} with a multivariate Gaussian variational family.
\begin{approximation}\label{approx:gaussian}
After each round $t$, approximate $p_t \approx N(\boldsymbol{\mu}_t,\Sigma_t)$ using one of the approximate inference methods above. We refer to this Gaussian approximation as the posterior.
\end{approximation}
We now define the terminal objective, expected opportunity cost (EOC), which we adopt for reasons discussed in the introduction. To write EOC it is convenient to let $S_k^*(\boldsymbol{\vartheta})$ denote the set of indices of the $k$ largest components of any vector $\boldsymbol{\vartheta}\in\mathbb R^I$. After $T$ rounds of evaluations, the decision-maker selects a set of items $S_k \subset \{1, \ldots, I \}$ with $|S_k|=k$ and the EOC is
\begin{equation}\label{eq:EOC}
    \mathrm{EOC}(S_k;p_T)
:=
\E_{\boldsymbol{\theta}}
\!\FitDelims{[}{]}{
\sum_{i\in S_k^*(\boldsymbol{\theta})} \theta_i-\sum_{i\in S_k}\theta_i
\,\FitMid{|}\, p_T
}
=
\E_{\boldsymbol{\theta}}
\!\FitDelims{[}{]}{
\sum_{i\in S_k^*(\boldsymbol{\theta})} \theta_i
\,\FitMid{|}\, p_T
}
-
\sum_{i\in S_k}\mu_{T,i}.
\end{equation}
Throughout the paper, conditioning on $p_t$ means that the expectation (or probability) is taken with respect to $\boldsymbol{\theta}\sim p_t$. The first term in \eqref{eq:EOC} does not depend on the choice of $S_k$, so minimizing EOC is equivalent to maximizing the sum of the selected set's posterior means (the second term). It follows that the optimal set to select given $\boldsymbol{\mu}_T$ is the set of $k$ items with the largest posterior means $S_k^*(\boldsymbol{\mu}_T)$.

With the terminal objective established, we next formulate the sequential problem of minimizing it as a dynamic program. At any round $t$ the state is described by $(p_{t-1},R_t)$. First consider a state where $\min(r_0,r_1)\le R_t<2\min(r_0,r_1)$, which is a sufficient condition for $t= T$. The decision-maker chooses an evaluation $(i,j)\in\mathcal I$ with $r_\jmath\le R_t$, an outcome $X_t=x$ is observed, and the posterior EOC is
\begin{equation}\label{eq:exoppcostT}
    \E_{\boldsymbol{\theta}}\!\FitDelims{[}{]}{
    \sum_{\ell\in S_k^*(\boldsymbol{\theta})}\theta_\ell
    \,\FitMid{|}\,X_t=x,(i,j),p_{t-1}
    }
    -
    \sum_{\ell\in S_k^*(\boldsymbol{\mu}_t(x\mid(i,j)))}
    \mu_{t,\ell}(x\mid(i,j)),
\end{equation}
where $\boldsymbol{\mu}_t(x\mid(i,j))=\E[\boldsymbol{\theta}\mid X_t=x,(i,j),p_{t-1}]$. Therefore, the problem is to choose the evaluation $(i,j)$ that minimizes the expectation of \eqref{eq:exoppcostT} over its outcome $X_t$. By the law of total expectation, the expectation of the first term equals $\E_{\boldsymbol{\theta}}[\sum_{\ell\in S_k^*(\boldsymbol{\theta})}\theta_\ell\mid p_{t-1}]$, which does not depend on the choice of $(i,j)$. Thus, the optimal evaluation is
\begin{equation*}
    (i_t,j_t)\in
    \arg\max_{(i,j)\in\mathcal I:\,r_\jmath\le R_t}
    \sum_{x\in\{-1,1\}}
    \Pp(X_{ij}=x\mid p_{t-1})
    \sum_{\ell\in S_k^*(\boldsymbol{\mu}_t(x\mid(i,j)))}
    \mu_{t,\ell}(x\mid(i,j)).
\end{equation*}
To consider the full sequential process, let $V^*(p_{t-1},R_t)$ denote the largest expected sum of terminal selected posterior means from state $(p_{t-1},R_t)$. If no evaluation is affordable, the decision-maker selects $S_k^*(\boldsymbol\mu_{t-1})$ now, giving
\begin{equation*}
V^*(p_{t-1},R_t)
=\sum_{\ell\in S_k^*(\boldsymbol\mu_{t-1})}\mu_{t-1,\ell},
\qquad R_t<\min(r_0,r_1).
\end{equation*}
Otherwise, the optimal value function satisfies the Bellman equation
\begin{equation}\label{eq:dpform}
V^*(p_{t-1},R_t)
=\max_{(i,j)\in\mathcal I:\ r_\jmath\le R_t}
\sum_{x\in\{-1,1\}}
\Pp\!\FitParens{X_{ij}=x\,\FitMid{|}\,p_{t-1}}
\times
V^*\!\FitParens{p_t(\cdot\mid x,(i,j)),R_t-r_\jmath},
\end{equation}
where $p_t(\cdot\mid x,(i,j))$ denotes the posterior obtained after starting from prior $p_{t-1}$, choosing evaluation $(i,j)$, and observing outcome $x$.

\subsection{Shortcomings of the knowledge gradient policy}\label{subsec:kgdiff}
\label{sec:model:kg}
Addressing the curse of dimensionality of the dynamic program in \eqref{eq:dpform}, the KG policy adopts the myopic version of it, given by
\begin{equation}\label{eq:vmyopic}
    V_{\text{myopic}}((i,j)\mid p_{t-1})
:=
\sum_{x\in\{-1,1\}}
\Pp\!\FitParens{X_{ij}=x\,\FitMid{|}\,p_{t-1}}
\FitParens{\sum_{\ell\in S_k^*(\boldsymbol{\mu}_t(x\mid(i,j)))}
\mu_{t,\ell}(x\mid(i,j))
-
\sum_{\ell\in S_k^*(\boldsymbol{\mu}_{t-1})}
\mu_{t-1,\ell}}.
\end{equation}
The KG policy then selects an evaluation according to
\begin{equation*}
\arg\max_{(i,j)\in\mathcal I:\ r_\jmath\le R_t}\frac{
V_{\text{myopic}}((i,j)\mid p_{t-1})}{r_\jmath}.
\end{equation*}
However, binary data can cause the KG policy to get ``stuck'' \citep{LiYucesan2024}. If the estimated top-$k$ set is sufficiently separated from the rest, no evaluation outcome can directly change the posterior top-$k$ set because the information a single binary outcome can carry is bounded, and with it how far the posterior means can move. This can occur frequently even when the top-$k$ set is far from certain. When this happens, KG has no basis to choose among evaluations, and a fallback rule, such as random allocation, is needed. This is not an issue with unbounded observations, such as a Gaussian likelihood, where there is always a possibility that a single evaluation can provide overwhelmingly positive/negative evidence that changes the posterior top-$k$ set.

\section{Surrogate objective}\label{sec:surrogate}
Developing the surrogate objective combines two ideas. The first is to establish the expected mismatch loss and connect it to EOC. The second is how it drives evaluation choices. We treat the expected mismatch loss as a static objective fixed at state $p_{t-1}$, using it to identify where uncertainty reduction is most valuable holding the current means fixed. The remainder of this section proceeds as follows. To provide insight into how the mismatch loss was developed, \cref{subsec:mismatch-intuition} presents our initial, simpler, intuition-based version, which then leads to the formal development in \cref{sec:miscost} that rigorously connects the surrogate to EOC and yields an improved form. Finally, \cref{sec:surrogate:program} develops the resulting surrogate loss.

\subsection{Mismatch loss: from intuition}\label{subsec:mismatch-intuition}
Recall from the introduction that mismatch loss is defined as the expected magnitude of the inclusion errors of $S_k^*(\boldsymbol{\mu}_{t-1})$ and the exclusion errors of $\{1,\ldots,I\} \setminus S_k^*(\boldsymbol{\mu}_{t-1})$. Measuring the magnitude of each error requires an anchor, a top-$k$ threshold against which they are measured. However, given the prior, the number of items above a given threshold is a random variable.  A natural approach is therefore to choose a threshold such that, in expectation, the number of items above it is $k$.

More explicitly, at each evaluation round $t$, fix a threshold $q$ on the utility scale such that, in expectation, $k$ items have utility above $q$ under the prior $p_{t-1}$. Then for each item $i\in S_k^*(\boldsymbol{\mu}_{t-1})$ the loss is $\E[(q-\theta_i)^+]$, the expected amount it falls short of the top-$k$ threshold, and similarly for each item $i\notin S_k^*(\boldsymbol{\mu}_{t-1})$ the loss is $\E[(\theta_i-q)^+]$, the expected amount it exceeds the top-$k$ threshold. We refer to the sum of these losses as the (informal) \emph{expected mismatch loss}
\begin{equation}\label{eq:missfixed}
   \sum_{i\in S_k^*(\boldsymbol{\mu}_{t-1})}\E\!\FitDelims{[}{]}{(q-\theta_i)^+\,\FitMid{|}\,p_{t-1}}
    +\sum_{i\notin S_k^*(\boldsymbol{\mu}_{t-1})}\E\!\FitDelims{[}{]}{(\theta_i-q)^+\,\FitMid{|}\,p_{t-1}}.
\end{equation}
The policy then chooses the evaluation whose resulting uncertainty reduction on $\boldsymbol{\theta}$ reduces expected mismatch loss the most per unit of resource. Two things make this conceptually attractive. First, it naturally directs evaluations toward items near the current expected top-$k$ cutoff ($q$) and toward items with substantial prior uncertainty, since those are the items that, if misclassified, will incur a high expected mismatch loss. Hence, this surrogate preserves the key feature of EOC: mistakes are penalized by their magnitude. Second, an evaluation can still be valuable even when its outcome cannot change this round's posterior top-$k$ membership if it tightens the posterior of items near $q$. This is in part because the evaluation informs where to direct later evaluations. Attempting a sports analogy: we don't only give credit for scoring goals but also for assists.

However, adaptively combining ratings and pairwise comparisons using this mismatch loss has a less than obvious nuance to it. Which items belong in the top-$k$ depends only on the relative utilities, not on the average utility $\overline{\theta}$. Thus, information about a change in $\overline{\theta}$ that shifts all utilities equally is \emph{decision-irrelevant}, and the surrogate should catch this. However, \eqref{eq:missfixed} may reward learning such information, since reducing uncertainty about such a common shift reduces the expectations $\E[(q-\theta_i)^+],\E[(\theta_i-q)^+]$. Ratings can provide information about $\overline{\theta}$ since they contain information about an item's absolute quality, while comparisons cannot since they only contain relative information. A surrogate based on this mismatch loss can therefore favor ratings over comparisons because it rewards decision-irrelevant uncertainty reduction.

Initially we addressed this by replacing $\boldsymbol{\theta}$ with its sum-to-zero counterpart $\widetilde{\boldsymbol{\theta}} = \boldsymbol{\theta} - \overline{\theta}\,\boldsymbol{1}$, where $\boldsymbol{1}$ is the all-ones vector, and computing the mismatch loss based on the differences $\widetilde{\theta}_i - q$ and $q - \widetilde{\theta}_i$, so that learning $\overline{\theta}$ does not affect the loss. We will next present a formal and more general treatment of mismatch loss, including its link to EOC. As a motivation for this, notice that $\widetilde{\theta}_i - q = \theta_i - (q+\overline{\theta})$, so the mismatch loss with the fixed threshold $q$ applied to $\widetilde{\boldsymbol{\theta}}$ can be thought of as our original mismatch loss with a random $\boldsymbol{\theta}$-dependent threshold $q+\overline{\theta}$. This observation invites the question of what is the ``best'' such \emph{random} top-$k$ threshold that allows for tractable analysis.

\subsection{Mismatch loss: formal derivation}\label{sec:miscost}
We begin by formally defining mismatch loss and showing how it relates to EOC. For any utility vector $\boldsymbol{\theta}$ and any threshold function $a:\R^I\to\R$, define the mismatch loss
\begin{equation}\label{eq:mismatchdef}
    \mathcal{M}(\boldsymbol{\theta};S_k,a)
    :=
    \sum_{i\in S_k}\big(a(\boldsymbol{\theta})-\theta_i\big)^+
    +\sum_{i\notin S_k}\big(\theta_i-a(\boldsymbol{\theta})\big)^+
    =
    k\,a(\boldsymbol{\theta})-\sum_{i\in S_k}\theta_i+\sum_{i=1}^I\big(\theta_i-a(\boldsymbol{\theta})\big)^+,
\end{equation}
where the second equality uses the identity $x^+=x+(-x)^+$. \cref{lem:mismatch} says that expected mismatch loss is an upper bound on EOC and that for specific threshold functions it equals EOC.

\begin{lemma}\label{lem:mismatch} Let $\theta^{(k)},\theta^{(k+1)}$ denote the $k$th and $(k\!+\!1)$th order statistics of $\boldsymbol{\theta}$ respectively.
\begin{enumerate}[label=(\roman*)]
\item For any threshold function $ a, \; \mathrm{EOC}(S_k;p_{t-1}) \le \E\!\FitDelims{[}{]}{\mathcal{M}(\boldsymbol{\theta};S_k,a)\,\FitMid{|}\,p_{t-1}}$.
    \item If $a(\boldsymbol{\theta})\in[\theta^{(k+1)},\theta^{(k)}]$ almost surely, then $\mathrm{EOC}(S_k;p_{t-1}) = \E\!\FitDelims{[}{]}{\mathcal{M}(\boldsymbol{\theta};S_k,a)\,\FitMid{|}\,p_{t-1}}$.
\end{enumerate}
\end{lemma}
\begin{proof}{Proof.}
All proofs are provided in the Appendix. \noHalmos
\end{proof}
\cref{lem:mismatch} applies to an arbitrary state $p_{t-1}$ but does not involve future evaluations. Thus, its usefulness is in ensuring that, over the course of the evaluation rounds, if we conduct evaluations that make mismatch loss small, then the resulting EOC will also be small. The bound holds with equality, for example, for the threshold $a(\boldsymbol{\theta})=\tfrac12\!\FitParens{\theta^{(k)}+\theta^{(k+1)}}$. However, taking the expectation of mismatch loss is not tractable when the threshold is a function of such order statistics. We therefore restrict $a$ to a family of functions that allows for tractable analysis and find, within that family, the threshold function that gives the tightest bound on EOC.

The expected mismatch loss $\E[\mathcal{M}(\boldsymbol{\theta};S_k,a)\mid p_{t-1}]$ has terms of the form $\E[(\theta_i-a(\boldsymbol{\theta}))^+\mid p_{t-1}]$. Since $\boldsymbol{\theta}\mid p_{t-1}$ is Gaussian, the difference $\theta_i-a(\boldsymbol{\theta})$ is Gaussian whenever the threshold is affine in $\boldsymbol{\theta}$, and the terms $\E[(\theta_i-a(\boldsymbol{\theta}))^+|p_{t-1}]$ then have expressions amenable to analysis. This motivates the affine family of thresholds
\begin{equation*}
    a_{q,\boldsymbol{b}}(\boldsymbol{\theta})
    :=
    q + \boldsymbol{b}^\top(\boldsymbol{\theta}-\boldsymbol{\mu}_{t-1}),
    \qquad q\in\R,\ \boldsymbol{b}\in\R^I,
\end{equation*}
where $q$ is the expected threshold level, and $\boldsymbol{b}$ governs how the threshold moves with the utilities. The difference $\theta_i-a_{q,\boldsymbol{b}}(\boldsymbol{\theta})$ is then $(\mu_{t-1,i}-q) + (\boldsymbol{e}_i-\boldsymbol{b})^\top(\boldsymbol{\theta}-\boldsymbol{\mu}_{t-1}),$ a constant mean term $\mu_{t-1,i}-q$ plus a mean-zero Gaussian term. By \cref{lem:mismatch} any choice of $(q,\boldsymbol{b})$ gives an upper bound for EOC, and a pair $(q_t^*,\boldsymbol{b}_t^*)$ that minimizes the expected mismatch loss gives the tightest upper bound for EOC within the affine threshold family. Before giving the optimality conditions for $(q_t^*,\boldsymbol{b}_t^*)$ in \cref{prop:affine}, it is convenient to define the following. Let ${\kappa}_{q,\boldsymbol{b}}(\boldsymbol{\theta}):=\bigl|\{i:\theta_i>a_{q,\boldsymbol{b}}(\boldsymbol{\theta})\}\bigr| - k$ denote the (random) deviation between the number of items above the threshold and its target $k$. That is, $k + {\kappa}_{q,\boldsymbol{b}}(\boldsymbol{\theta})$ is the number of items that exceed the threshold.
\begin{theorem}\label{prop:affine}
The expected mismatch loss $\E[\mathcal{M}(\boldsymbol{\theta};S_k,a_{q,\boldsymbol{b}})\mid p_{t-1}]$ is convex in $(q,\boldsymbol{b})$ and attains a minimum. A pair $(q_t^*,\boldsymbol{b}_t^*)$ is a minimizer only if either
\begin{enumerate}[label=(\roman*)]
    \item $(q_t^*,\boldsymbol{b}_t^*)=(\mu_{t-1,j},\boldsymbol{e}_j)$ for some $j=1,\ldots,I$, or
    \item $(q_t^*,\boldsymbol{b}_t^*)\neq(\mu_{t-1,j},\boldsymbol{e}_j)$ for every $j=1,\ldots,I$, and \vspace{-0.5\baselineskip}
    \begin{equation*}
        \E\!\FitDelims{[}{]}{\kappa_{q_t^*,\boldsymbol{b}_t^*}(\boldsymbol{\theta})\,\FitMid{|}\,p_{t-1}}=0
        \qquad\text{and}\qquad
        \Cov\!\FitParens{\theta_\ell,\kappa_{q_t^*,\boldsymbol{b}_t^*}(\boldsymbol{\theta})\,\FitMid{|}\,p_{t-1}}=0
        \quad\text{for every $\ell=1,\ldots,I$.}
    \end{equation*}
\end{enumerate}
\vspace{-0.5\baselineskip}
Conversely, any pair satisfying \textup{(ii)} is a minimizer.
\end{theorem}
Case \textup{(i)} arises when $\boldsymbol{b}=\boldsymbol{e}_j$, so $\theta_j-a_{q,\boldsymbol{b}}(\boldsymbol{\theta})=\mu_{t-1,j}-q$ and $\E[(\theta_j-a_{q,\boldsymbol{b}}(\boldsymbol{\theta}))^+\mid p_{t-1}]=(\mu_{t-1,j}-q)^+$ is not differentiable at $q=\mu_{t-1,j}$. In the Appendix, we derive the necessary and sufficient condition for when such an exceptional point would be optimal. Otherwise, any optimum is characterized by \textup{(ii)}, which will capture most practical cases. We henceforth refer to any minimizer satisfying \textup{(ii)} as an optimal affine threshold and focus our discussion and interpretation on this case.

\cref{prop:affine} says that the optimal affine threshold satisfies the following two properties. First, in expectation, $k$ items will have their utility above it. Second, the covariance between each item's utility and the deviation ${\kappa}_{q,\boldsymbol{b}}(\boldsymbol{\theta})$ is $0$. If $\kappa_{q,\boldsymbol{b}}$ were positively (negatively) correlated with $\theta_\ell$, then when item $\ell$'s utility is larger than expected, in expectation more (fewer) than $k$ items would be above the threshold, meaning that the threshold is not moving up enough (is moving up too much) with item $\ell$. At the optimum, no such \emph{linear} dependence remains: the threshold moves with $\boldsymbol{\theta}$ in a way such that knowing that any particular item is high or low gives no first-order information about whether the threshold has overshot or undershot. Next, \cref{cor:affine-gauss} gives an explicit fixed-point characterization of $(q_t^*,\boldsymbol{b}_t^*)$ in our Gaussian setting, which uses the following notation.
\begin{equation}\label{eq:varforms}
\begin{aligned}
    \sigma_{t-1,i}^2 &:= \Var\FitParens{\theta_i-a_{q,\boldsymbol{b}}(\boldsymbol{\theta})\FitMid{|}p_{t-1}} = (\boldsymbol{e}_i-\boldsymbol{b})^\top\Sigma_{t-1}(\boldsymbol{e}_i-\boldsymbol{b}), \\
    \sigma_{t-1,ij}^2 &:= \Var\FitParens{\theta_i-\theta_j\FitMid{|}p_{t-1}} = (\boldsymbol{e}_i-\boldsymbol{e}_j)^\top\Sigma_{t-1}(\boldsymbol{e}_i-\boldsymbol{e}_j),
\end{aligned}
\end{equation}
where $\boldsymbol{e}_0=\boldsymbol{0}$, so $\sigma_{t-1,i0}^2=\Sigma_{t-1,ii}$ is the variance of item $i$. We use this notation because, going forward, we will care more about the variance of items relative to the threshold $\sigma_{t-1,i}^2$ than about an item's variance $\sigma_{t-1,i0}^2$.

\begin{corollary}\label{cor:affine-gauss}
Define $z_{t-1,i}:=\frac{\mu_{t-1,i}-q}{\sigma_{t-1,i}}$ and $w_{t-1,i}:=\frac{\phi\!\FitParens{z_{t-1,i}}}{\sigma_{t-1,i}}$.
\begin{enumerate}[label=(\roman*)]
    \item The optimality conditions of \cref{prop:affine}(ii) are satisfied by $(q_t^*,\boldsymbol{b}_t^*)$ such that
\begin{equation*}
    \sum_{i=1}^I\Phi\!\FitParens{z_{t-1,i}}=k,
    \qquad
    {b}_{t,i}^*=\frac{w_{t-1,i}}{\sum_{\ell=1}^I w_{t-1,\ell}} \quad \text{for} \ i = 1,\ldots, I
    .
\end{equation*}
\item If the threshold is restricted to be constant (i.e., $\boldsymbol{b}=\boldsymbol{0}$), so that $a_{q,\boldsymbol{0}}(\boldsymbol{\theta})=q$, the optimal threshold $q^{\text{con}}_t$ satisfies
\begin{equation*}
    \sum_{i=1}^I\Phi\!\FitParens{\frac{\mu_{t-1,i}-q^{\text{con}}_t}{\sigma_{t-1,i0}}}=k.
\end{equation*}
\end{enumerate}
\end{corollary}
To lighten notation, we write $(q_t,\boldsymbol{b}_t)$ for the optimal pair $(q_t^*,\boldsymbol{b}_t^*)$. Each $b_{t,i}$ describes how the threshold responds to movements in item $i$'s utility. Since $b_{t,i}\propto w_{t-1,i}=\phi(z_{t-1,i})/\sigma_{t-1,i}$, this response is concentrated on items likely near the top-$k$ boundary. In this sense, the optimal threshold can be viewed as a first-order proxy for the random top-$k$ threshold: it moves linearly with the items that are most informative about where the threshold lies. Moreover, because $\sum_{i=1}^I b_{t,i}=1$, the difference $\theta_i-a_{q_t,\boldsymbol{b}_t}(\boldsymbol{\theta})$ depends only on the relative utilities, not on $\overline{\theta}$. Thus, mismatch loss does not reward reducing decision-irrelevant uncertainty about $\overline{\theta}$, putting ratings and comparisons on the same footing in line with the intuition of \cref{subsec:mismatch-intuition}. Note that the special case constructed in \cref{subsec:mismatch-intuition} is in the set of affine thresholds: subtracting $\overline{\theta}$ corresponds to $\boldsymbol{b}=\boldsymbol{1}/I$.

\subsection{The surrogate loss}\label{sec:surrogate:program}
As discussed above, we treat the expected mismatch loss $\E[\mathcal{M}(\boldsymbol{\theta};S_k^*(\boldsymbol{\mu}_{t-1}),a_{q_t,\boldsymbol{b}_t})\mid p_{t-1}]$ as a static objective fixed at state $p_{t-1}$ and use it to estimate how valuable each potential evaluation will be. We fix $\boldsymbol{\mu}_{t-1}$, $S_k^*(\boldsymbol{\mu}_{t-1})$, and $(q_t,\boldsymbol{b}_t)$ and value an evaluation $(i,j)$ through the uncertainty it is expected to reduce. Accordingly, define its expected posterior covariance after evaluation $(i,j)$
\begin{equation*}
\overline{\Sigma}_t((i,j),p_{t-1})
:=
\sum_{x\in\{-1,1\}}
\Pp(X_{ij}=x\mid p_{t-1})\,
\Cov\!\FitParens{\boldsymbol{\theta}\mid X_{ij}=x,p_{t-1}},
\end{equation*}
and $\overline{p}_t(i,j) := N\!\FitParens{\boldsymbol{\mu}_{t-1},\overline{\Sigma}_t((i,j),p_{t-1})}$. We call $\overline{p}_t(i,j)$ the \emph{expected posterior}, since both of its moments are the expected posterior moments: the expected covariance, and the mean by the martingale property of Bayesian updating. Then we define the surrogate loss associated with evaluation $(i,j)$ as the expected mismatch loss, with expectation taken with respect to $\overline{p}_t(i,j)$,
\begin{equation}\label{eq:surrogate-mismatch}
    \E_{\boldsymbol{\theta} \sim \overline{p}_t(i,j)}\!\FitDelims{[}{]}{
        \mathcal{M}\!\FitParens{
            \boldsymbol{\theta};
            S_k^*(\boldsymbol{\mu}_{t-1}),
            a_{q_t,\boldsymbol{b}_t}
        }
    }.
\end{equation}
In \eqref{eq:surrogate-mismatch}, the current prior and the threshold determine which items' uncertainty is consequential, while $\overline{\Sigma}_t((i,j),p_{t-1})$ determines how much of that uncertainty is expected to remain after evaluation $(i,j)$. \cref{lem:expected-cov} gives an explicit expression for $\overline{\Sigma}_t((i,j),p_{t-1})$.

\begin{lemma}\label{lem:expected-cov}
Starting with prior $N(\boldsymbol{\mu}_{t-1},\Sigma_{t-1})$, let $\mu_{t-1,0}:=\theta_0$ and, for evaluation $(i,j)\in\mathcal I$, define
\begin{equation*}
\alpha_{t,ij}:=
\frac{\beta_\jmath(\mu_{t-1,i}-\mu_{t-1,j})}
{\sqrt{1+\beta_\jmath^2\sigma_{t-1,ij}^2}} \qquad \text{and} \qquad \delta_{t,ij}:=
\frac{\beta_\jmath^2\sigma_{t-1,ij}^2}
{1+\beta_\jmath^2\sigma_{t-1,ij}^2}
\frac{\phi(\alpha_{t,ij})^2}
{\Phi(\alpha_{t,ij})\Phi(-\alpha_{t,ij})}.
\end{equation*}
Then the expected posterior covariance is
\begin{equation}\label{eq:expected-cov}
\overline{\Sigma}_t((i,j),p_{t-1})
=\Sigma_{t-1}
-\frac{\delta_{t,ij}}{\sigma_{t-1,ij}^2}
\Sigma_{t-1}(\boldsymbol e_i-\boldsymbol e_j)
(\boldsymbol e_i-\boldsymbol e_j)^\top\Sigma_{t-1}.
\end{equation}
\end{lemma}
Here $\delta_{t,ij}$ is the expected proportion of the variance of $\theta_i-\theta_j$ reduced by the evaluation. This depends on how uncertain the outcome is a priori, the variance of $\theta_i-\theta_j$, and the evaluation accuracy. The covariance terms in \eqref{eq:expected-cov} determine how the information learned about $(i,j)$ indirectly affects our posterior about other items through correlation. Plugging in the expected covariance in \eqref{eq:expected-cov}, the surrogate loss in \eqref{eq:surrogate-mismatch} is
\begin{equation*}
    \E_{\boldsymbol{\theta} \sim \overline p_t(i,j)}\!\FitDelims{[}{]}{\mathcal{M}\!\FitParens{\boldsymbol{\theta};S_k^*(\boldsymbol{\mu}_{t-1}),a_{q_t,\boldsymbol{b}_t}}}
    \;=\;
    k\,q_t
    - \sum_{\ell\in S_k^*(\boldsymbol{\mu}_{t-1})}\mu_{t-1,\ell}
    \;+\;\sum_{\ell=1}^I
    \E_{\boldsymbol{\theta} \sim \overline p_t(i,j)}\!\FitDelims{[}{]}{(\theta_\ell-a_{q_t,\boldsymbol{b}_t}(\boldsymbol{\theta}))^+}.
\end{equation*}
On the right-hand side, only the last term depends on $(i,j)$, which can be thought of as the controllable part of the surrogate loss. Let $\sigma'_{t,\ell}(i,j)$ and $z'_{t,\ell}(i,j)$ denote the analogues of $\sigma_{t-1,\ell}$ and $z_{t-1,\ell}$ from \eqref{eq:varforms} and \cref{cor:affine-gauss} computed under the expected posterior. The controllable surrogate loss can be written as
\begin{align}
    \mathcal{L}((i,j)\mid p_{t-1})
    &:=\sum_{\ell=1}^{I}
    \E_{\boldsymbol{\theta} \sim \overline p_t(i,j)}
    \FitDelims{[}{]}{(\theta_\ell-a_{q_t,\boldsymbol{b}_t}(\boldsymbol{\theta}))^+}
    \notag\\
    &=\sum_{\ell=1}^{I}\FitDelims{[}{]}{
    (\mu_{t-1,\ell}-q_t)\,\Phi(z'_{t,\ell}(i,j))
    +\sigma'_{t,\ell}(i,j)\,\phi(z'_{t,\ell}(i,j))
    }.
    \label{eq:surrogate-loss}
\end{align}
For later use, let $\mathcal L_\ell((i,j)\mid p_{t-1})$ denote the $\ell$th term of \eqref{eq:surrogate-loss}. The allocation rule in \cref{sec:greedy} chooses, in each round, the evaluation whose resulting decrease in surrogate loss per unit of resource is largest.

\section{Sequential evaluation allocation}\label{sec:greedy}
We next use the surrogate loss to construct the sequential evaluation policy. \Cref{sec:surrogate:greedy} derives the marginal decrease in the surrogate per evaluation and the allocation rule based on it, \cref{sec:surrogate:greedy_alg} gives the sequential procedure, and \cref{sec:surrogate:consistency} establishes that the policy is asymptotically consistent, meaning EOC $\to 0$ as $R\to\infty$.

\subsection{Marginal scores}
\label{sec:surrogate:greedy}
Consider an arbitrary round $t$ for which we write $\mathcal L((i,j))$ and $\mathcal L_\ell((i,j))$ for the losses in \eqref{eq:surrogate-loss}, leaving the dependence on $p_{t-1}$ implicit. Let $\mathcal L(0)=\sum_{\ell=1}^I\mathcal L_\ell(0)$ be the loss before any evaluation, obtained by replacing $\sigma'_{t,\ell}(i,j)$ and $z'_{t,\ell}(i,j)$ with $\sigma_{t-1,\ell}$ and $z_{t-1,\ell}$. To choose what evaluation to conduct we compute the decrease in surrogate loss per unit of resource
\begin{equation*}
\psi_{t,ij} :=\frac{\mathcal L(0)-\mathcal L((i,j))}{r_\jmath}
\end{equation*}
for each evaluation, then evaluate $\arg\max_{(i,j)\in\mathcal I:\,r_\jmath\leq R_t}\psi_{t,ij}$. This procedure, Surrogate Allocation (\textsc{SA}), is given in \cref{alg:surrogate}.

\begin{algorithm}[!htb]
\caption{Surrogate Allocation (\textsc{SA})}\label{alg:surrogate}
\makeatletter
\def\theHALG@line{surrogate.\arabic{ALG@line}}
\makeatother
\begin{algorithmic}[1]
\Require $\boldsymbol{\mu}_{t-1}$, $\Sigma_{t-1}$, $R_t\geq\min(r_0,r_1)$
\State Compute a minimizing threshold $(q_t,\boldsymbol b_t)$ via \cref{prop:affine,cor:affine-gauss}
\State Compute $\mathcal L(0)$
\For{$(i,j)\in\mathcal I$ with $r_\jmath\leq R_t$}
    \State Compute $\overline{\Sigma}_t((i,j),p_{t-1})$ by \eqref{eq:expected-cov}
    \State Compute $\mathcal L((i,j))$ by \eqref{eq:surrogate-loss}
    \State $\psi_{t,ij}\leftarrow[\mathcal L(0)-\mathcal L((i,j))]/r_\jmath$
\EndFor
\State \Return an evaluation in $\arg\max_{(i,j)\in\mathcal I:\,r_\jmath\leq R_t}\psi_{t,ij}$
\end{algorithmic}
\end{algorithm}
\noindent To get insight into what drives the evaluation choice, we write the surrogate marginal decrease explicitly.
\begin{equation}
    \psi_{t,ij}
    =\frac{\mathcal L(0)-\mathcal L((i,j))}{r_\jmath}
    =\frac{1}{r_\jmath}\sum_{\ell=1}^I
    \int_{(\sigma'_{t,\ell}(i,j))^2}^{\sigma_{t-1,\ell}^2}
    \frac{\partial\mathcal L_\ell}{\partial v}\,dv
    =\frac{1}{r_\jmath}\sum_{\ell=1}^I
    \int_0^{1-\frac{(\sigma'_{t,\ell}(i,j))^2}{\sigma_{t-1,\ell}^2}}
    \frac{\sigma_{t-1,\ell}}{2\sqrt{1-s}}
    \phi\!\FitParens{\frac{z_{t-1,\ell}}{\sqrt{1-s}}}\,ds.
    \label{eq:score}
\end{equation}
The upper limit in the integral corresponds to the proportion of item $\ell$'s uncertainty (relative to the threshold) reduced by conducting evaluation $(i,j)$. To see this more clearly, by \cref{lem:expected-cov} we have
\begin{equation*}
    1-\frac{(\sigma'_{t,\ell}(i,j))^2}{\sigma_{t-1,\ell}^2}
    =\delta_{t,ij}\,
    \mathrm{Corr}\!\FitParens{\theta_i-\theta_j,\,
    \theta_\ell-a_{q_t,\boldsymbol b_t}(\boldsymbol\theta)}^2,
\end{equation*}
where $\delta_{t,ij}$ is the expected proportion of the variance of $\theta_i-\theta_j$ reduced by the evaluation. The squared term measures how much of the variance reduction of $(i,j)$ transfers to item $\ell$ through correlation.

The integrand measures how valuable it is to reduce item $\ell$'s uncertainty for our objective. It is the instantaneous decrease in expected mismatch loss per additional proportion of its variance removed. The $\phi(z_{t-1,\ell})$ term is large when we are highly uncertain about which side of the top-$k$ threshold item $\ell$ lies on, which occurs when either the mean $\mu_{t-1,\ell}$ is close to the expected threshold level $q_t$ or when $\sigma_{t-1,\ell}$ is large. The additional standard deviation term $\sigma_{t-1,\ell}$ scales this uncertainty by how far the item's utility may be from the threshold, reflecting that selection mistakes are penalized by their magnitude.

Put together, \eqref{eq:score} scores an evaluation by how much its uncertainty reduction impacts expected mismatch loss across all items. The integrand captures how valuable reducing item $\ell$'s uncertainty (relative to the threshold) is, and the upper limit of the integral captures how much of this item's uncertainty is reduced by evaluating $(i,j)$, either directly when $\ell\in\{i,j\}$ or indirectly through correlation when $\ell\notin\{i,j\}$.

\subsection{The sequential procedure}\label{sec:surrogate:greedy_alg}
Now that we have an algorithm to choose the next evaluation, we lay out the full sequential evaluation procedure in \cref{alg:runner}. It takes a family of policies $\{\textsc{EvaluationRule}^{R_t}\}$ that map the state and budget to an affordable evaluation. For now, the evaluation rule can be thought of as \textsc{SA}, but we write it in general form as we will introduce a different evaluation rule for the endgame in \cref{sec:endgame}. It also assumes the posterior update comes from a \textsc{BeliefUpdate} procedure implementing an approximate inference method from \cref{approx:gaussian}.
\vspace{-8pt}
\begin{algorithm}[!htb]
\caption{Sequential procedure}\label{alg:runner}
\begin{algorithmic}[1]
\Require prior $(\boldsymbol{\mu}_0,\Sigma_0)$, $R$, $\{\textsc{EvaluationRule}^{R_t}\}$
\State $t\leftarrow1$, \quad $R_1\leftarrow R$
\While{$R_t\geq\min(r_0,r_1)$}
    \State $(i_t,j_t)\leftarrow\textsc{EvaluationRule}^{R_t}(\boldsymbol{\mu}_{t-1},\Sigma_{t-1})$
    \State Observe outcome $X_t$
    \State $(\boldsymbol{\mu}_t,\Sigma_t)\leftarrow
    \textsc{BeliefUpdate}(\boldsymbol{\mu}_{t-1},\Sigma_{t-1},(i_t,j_t),X_t)$
    \State $R_{t+1}\leftarrow R_t-r_{\jmath_t}$
    \State $t\leftarrow t+1$
\EndWhile
\State $T\leftarrow t-1$
\State \Return $S_k^*(\boldsymbol{\mu}_T)$
\end{algorithmic}
\end{algorithm}
\vspace{-15pt}
\subsection{Asymptotic consistency}\label{sec:surrogate:consistency}
A natural concern with a policy based on a static objective is that it may reach a state from which, even with an infinite remaining budget, some items outside the final top-$k$ choice retain a positive probability of belonging to the true top-$k$. We show that this can never be the case for \cref{alg:runner} with \textsc{EvaluationRule}$^{R_t}$ = \textsc{SA}: in particular, the algorithm is asymptotically consistent, meaning EOC $\to 0$ as $R\to\infty$, under two mild assumptions.

\begin{assumption}\label{ass:link}
The true link function $g:\R\to(0,1)$ is continuous, symmetric ($g(-x)=1-g(x)$), strictly increasing, and log-concave.
\end{assumption}
\cref{ass:link} is satisfied by the probit link we use ($g = \Phi$) and allows for more general links.

\begin{assumption}\label{ass:update}
Let $(\boldsymbol{\mu}_t^*,\Sigma_t^*)$ denote the exact first two moments of the posterior (which are generally unavailable in closed form). The approximate posterior moments $(\boldsymbol{\mu}_t,\Sigma_t)$ satisfy: \textup{(i)}~there exist constants $0<\chi_-\le \chi_+<\infty$ such that, almost surely, for all sufficiently large $t$, $\chi_-\,\Sigma_t^*\preceq\Sigma_t\preceq \chi_+\,\Sigma_t^*$, \textup{(ii)}~$\sup_t\|\boldsymbol{\mu}_t\|_\infty<\infty$ almost surely, and \textup{(iii)}~$\E[\|\boldsymbol{\mu}_T-\boldsymbol{\mu}_T^*\|_\infty]\to 0$ as $R\to\infty$, where $\|\cdot\|_\infty$ is the L-infinity norm.
\end{assumption}
\cref{ass:update} says that whatever method we use in the \textsc{BeliefUpdate} step eventually yields a ``good'' estimate of the first two moments of the true posterior given the collected data.

\begin{theorem}\label{thm:consistency}
Under \cref{ass:link,ass:update}, \cref{alg:runner} with \textnormal{\textsc{EvaluationRule}}$^{R_t}$ $=$ \textnormal{\textsc{SA}} satisfies
\begin{equation*}
\E\!\FitDelims{[}{]}{
    \sum_{i\in S_k^*(\boldsymbol{\theta})}\theta_i
    -
    \sum_{i\in S_k^*(\boldsymbol{\mu}_T)}\theta_i
}
\to0
\qquad\text{as}\qquad R\to\infty.
\end{equation*}
\end{theorem}
\noindent\textbf{Proof idea.}
We outline the two main components of the proof. The first and more technically challenging part is to show that the set of evaluations the algorithm samples infinitely many times spans $\boldsymbol{1}^\perp := \{\boldsymbol{x}\in\R^I : \boldsymbol{1}^\top\boldsymbol{x} = 0\}$, the decision-relevant subspace, which we prove by contradiction. For an evaluation $(i,j)$, we refer to $\boldsymbol{e}_i - \boldsymbol{e}_j$ as its ``direction''. Let $E_\infty$ be the set of directions sampled infinitely often. Our goal is then to prove that $\boldsymbol{1}^\perp \subseteq \mathrm{span}(E_\infty)$ almost surely. Equivalently, for any fixed set of directions $\mathcal E$ whose span does not contain $\boldsymbol{1}^\perp$, the event $\Omega_{\mathcal E} := \{E_\infty = \mathcal E\}$ has probability $0$. Suppose for contradiction that $\Omega_{\mathcal E}$ has positive probability. On $\Omega_{\mathcal E}$, after some finite random time, every subsequently sampled direction is in $\mathcal E$. This reduces the proof to showing two results: (i) the marginal scores of directions in $\mathcal E$ tend to $0$, and (ii) at least one direction not in $\mathcal E$ has a marginal score bounded away from $0$. Together these imply that in some round \textsc{SA} chooses a direction not in $\mathcal E$, a contradiction. The second component then demonstrates that if $\boldsymbol{1}^\perp \subseteq \mathrm{span}(E_\infty)$, the posterior variance on $\boldsymbol{1}^\perp$ goes to $0$, and thus EOC decays to $0$ as well.

The significance of this result lies not in the asymptotic guarantee itself, but in what it reveals about the surrogate's structure.  Despite being short-sighted and having no designated randomization or ``exploration'' phase, the policy explores sufficiently.

\section{The endgame}\label{sec:endgame}

Conceptually, the allocation rule in \textsc{SA} is compelling, particularly for early and middle rounds: even if it is not forward-looking in a dynamic sense, it values evaluations that are useful beyond their potential influence on this round's posterior top-$k$ set. But near the end of the horizon, the value of such exploration is small, and in the last round it is zero. Hence, since the surrogate objective is not horizon-aware, this strength of \textsc{SA} can turn negative in \emph{the endgame}. To ``zoom in'' on the items that are actual top-$k$ candidates in the endgame is precisely the nature of KG-type policies. We develop the KG policy for our problem in \cref{sec:kg:mval} and a rule for when to switch to it in \cref{sec:switching-rule}.

\subsection{The sequential KG policy}\label{sec:kg:mval}

Recall from \cref{subsec:kgdiff} that the KG policy chooses an evaluation
\begin{equation*}
(i_t,j_t)\in
\arg\max_{(i,j)\in\mathcal I:\,r_\jmath\le R_t}
\psi^{\mathrm{KG}}_{t,ij} \quad \text{where} \quad
\psi^{\mathrm{KG}}_{t,ij}
:=
\frac{V_{\text{myopic}}((i,j)\mid p_{t-1})}{r_\jmath}.
\end{equation*}
To compute $V_{\text{myopic}}((i,j)\mid p_{t-1})$, whose expression is given in \eqref{eq:vmyopic}, \Cref{lem:kg-update} gives explicit expressions for the probability of each outcome of an evaluation and the resulting posterior means given the outcome.

\begin{lemma}\label{lem:kg-update}
Starting with prior $N(\boldsymbol{\mu}_{t-1},\Sigma_{t-1})$, for evaluation $(i,j)\in\mathcal I$ and outcome $x\in\{-1,1\}$,
\begin{equation*}
\Pp(X_{ij}=x\mid p_{t-1})=\Phi(x\alpha_{t,ij}) \quad \text{and}\quad \displaystyle
    \boldsymbol{\mu}_t(x\mid(i,j))
    =\boldsymbol{\mu}_{t-1}
    +\frac{x\beta_\jmath\phi(x\alpha_{t,ij})}
    {\Phi(x\alpha_{t,ij})\sqrt{1+\beta_\jmath^2\sigma_{t-1,ij}^2}}
    \Sigma_{t-1}(\boldsymbol e_i-\boldsymbol e_j),
\end{equation*}
where $\alpha_{t,ij}$ is defined in \cref{lem:expected-cov}.
\end{lemma}
As discussed in \cref{subsec:kgdiff}, when no outcome of any affordable evaluation can change the posterior top-$k$ set, all of the myopic gains $\psi^{\mathrm{KG}}_{t,ij}$ are zero. The KG policy then requires a fallback rule to determine which evaluation to conduct. \Cref{alg:kg} gives the resulting KG procedure.

\begin{algorithm}[!htb]
\caption{Knowledge gradient (\textsc{KG})}\label{alg:kg}
\makeatletter
\def\theHALG@line{kg.\arabic{ALG@line}}
\makeatother
\begin{algorithmic}[1]
\Require $\boldsymbol{\mu}_{t-1}$, $\Sigma_{t-1}$, $R_t\geq\min(r_0,r_1)$
\State Compute $\psi^{\mathrm{KG}}_{t,ij}$ for all $(i,j)\in\mathcal I$ with $r_\jmath\leq R_t$
\State $(i,j)\leftarrow\arg\max_{(i,j)\in\mathcal I:\,r_\jmath\leq R_t}\psi^{\mathrm{KG}}_{t,ij}$
\If{$\psi^{\mathrm{KG}}_{t,ij}=0$}
    \State Choose an affordable $(i,j)$ using the fallback rule
\EndIf
\State \Return $(i,j)$
\end{algorithmic}
\end{algorithm}

\subsection{Endgame when?}\label{sec:switching-rule}
Up to this point, \cref{alg:runner} provides the general sequential procedure, and we have given two algorithms for its \textsc{EvaluationRule}: \textsc{SA} (\cref{alg:surrogate}) and \textsc{KG} (\cref{alg:kg}). What remains is a rule for when to use which. \textsc{SA} was designed for the bulk of the evaluation process and \textsc{KG} for the endgame, so we consider the following class of policies: run \textsc{SA} until a predetermined share $\nu$ of the budget has been spent, then switch to \textsc{KG}.\footnote{In practice, the switching point can also be chosen state-dependently, that is, during the run as a function of $p_{t-1}$ rather than fixed in advance. We restrict attention to a predetermined switching point for simplicity.} We refer to $\nu$ as the switching point. Within this class, we choose the switching point via Monte Carlo simulation. For $\nu\in[0,1]$, let $\pi_\nu$ denote the policy obtained from \cref{alg:runner} using $\textsc{EvaluationRule}^{R_t}=\textsc{SA}$ while $R-R_t<\nu R$ and $\textsc{EvaluationRule}^{R_t}=\textsc{KG}$ once $R-R_t\geq\nu R$, so that $\pi_1$ is \textsc{SA} throughout and $\pi_0$ is \textsc{KG} throughout. Given a problem specification $(I,k,R,r_0,r_1,p_0,\beta_0,\beta_1,\theta_0)$, let
\begin{equation*}
\mathrm{EOC}\FitParens{S_k^*(\boldsymbol{\mu}_T);\pi_\nu}
:=
\E\!\FitDelims{[}{]}{
\sum_{i\in S_k^*(\boldsymbol{\theta})} \theta_i-\sum_{i\in S_k^*(\boldsymbol{\mu}_T)}\theta_i
\,\FitMid{|}\ \pi_\nu},
\end{equation*}
where the expectation is over $\boldsymbol{\theta}$ with prior distribution $p_0$ and over the outcomes $X_1,\ldots,X_T$ drawn from the likelihood under policy $\pi_\nu$.

We choose the switching point by a marginal analysis over a fixed set of candidate switching points $0\leq\nu_1<\nu_2<\cdots<\nu_J\leq1$. Let $\widehat{\mathrm{EOC}}\FitParens{S_k^*(\boldsymbol{\mu}_T);\pi_\nu}$ denote the Monte Carlo estimate of $\mathrm{EOC}\FitParens{S_k^*(\boldsymbol{\mu}_T);\pi_\nu}$, obtained by averaging the realized opportunity cost $\sum_{i\in S_k^*(\boldsymbol{\theta})} \theta_i-\sum_{i\in S_k^*(\boldsymbol{\mu}_T)}\theta_i$ over $M$ independent simulated runs of $\pi_\nu$, each drawing $\boldsymbol{\theta}\sim p_0$ and the outcomes from the specified likelihood. Starting from $\nu_J$ ($\nu_J=1$ meaning \textsc{SA} all the way through), we move the switching point back one candidate at a time, from $\nu_h$ to $\nu_{h-1}$ (peeling back the last stretch of \textsc{SA} into \textsc{KG}), as long as the marginal reduction $\widehat{\mathrm{EOC}}\FitParens{S_k^*(\boldsymbol{\mu}_T);\pi_{\nu_h}}-\widehat{\mathrm{EOC}}\FitParens{S_k^*(\boldsymbol{\mu}_T);\pi_{\nu_{h-1}}}$ remains positive. We stop at the first $\nu^*$ at which peeling back further fails to reduce $\widehat{\mathrm{EOC}}$, and deploy the policy $\pi_{\nu^*}$. We will refer to this as the switching procedure, and to our method, the policy $\pi_{\nu^*}$, as SE (Surrogate to Endgame).

\section{Batching extension}\label{sec:batching}
To conclude the analysis, we consider settings where evaluations are assigned in batches whose outcomes are observed at once before the next batch is chosen. While we have presented the problem sequentially, the model and analysis for the surrogate were developed with batching in mind, and extend naturally.

For the batching setting, let $T$ denote the \emph{fixed} number of rounds and let $B_t\geq \min\{r_0,r_1\}$ denote the budget for round $t$, with $\sum_{t=1}^T B_t=R$. Let $n(t)$ denote the batch chosen in this round, where $n_{ij}(t)$ is the number of evaluations of $(i,j)$.

\subsection{Surrogate batch construction}\label{sec:batching:surrogate}
Choosing the batch that minimizes the surrogate loss in each round is an integer program that quickly becomes intractable as the batch budget grows. We therefore build a batch by committing to one evaluation at a time. The first evaluation is chosen as in the fully sequential problem, by maximizing the marginal decrease in surrogate loss per unit of resource from \cref{sec:surrogate:greedy}. However, unlike the sequential problem, we must choose subsequent evaluations before observing the outcomes of the evaluation already committed to the batch. If we continued to use the original marginal scores, we would value each evaluation as if it were the only one in the batch, allocating multiple evaluations to reduce the same uncertainty.

Consider the choice of the second evaluation. Suppose the first evaluation committed to the batch is $(i,j)$. Although we have not yet observed its outcome, we can account for the information we expect to learn from it through its expected posterior covariance. By \cref{lem:expected-cov},
\begin{equation*}
\overline{\Sigma}_t((i,j),p_{t-1})
=
\Sigma_{t-1}
-
\frac{\delta_{t,ij}}{\sigma_{t-1,ij}^2}
\Sigma_{t-1}(\boldsymbol e_i-\boldsymbol e_j)
(\boldsymbol e_i-\boldsymbol e_j)^\top\Sigma_{t-1}.
\end{equation*}
This covariance reflects the uncertainty expected to remain \emph{after} the first evaluation. To choose the second evaluation, we use this covariance in place of $\Sigma_{t-1}$ and recompute the marginal scores from \cref{sec:surrogate:greedy}, keeping $\boldsymbol{\mu}_{t-1}$ and $(q_t,\boldsymbol b_t)$ fixed. Each candidate evaluation is scored by its \emph{additional} reduction in surrogate loss, accounting for the information we already expect to learn from the first. We choose the evaluation with the highest marginal score per unit of resource.

We next extend this approach to constructing an entire batch. The complication is that computing the expected posterior covariance after several committed evaluations requires taking the weighted average over their (many) possible joint outcomes. Instead, we approximate it by repeatedly applying the single-evaluation expected covariance update, using the updated expected covariance from the evaluations already committed.

Formally, let $n$ denote the matrix recording the evaluations committed so far in the batch, where $n_{ij}$ corresponds to the number of evaluations of $(i,j)$. Define $F^{ij}$ as a matrix of the same size as $n$, with $F^{ij}_{ij}=1$ and zero elsewhere. Let $\Sigma'_t(n)$ denote our approximation of the expected posterior covariance, with $\Sigma'_t(0)=\Sigma_{t-1}$. After adding evaluation $(i,j)$ to the batch, it updates according to
\begin{equation}\label{eq:batch-cov-update}
    \Sigma'_t(n+F^{ij})
    =
    \Sigma'_t(n)
    -
    \frac{\delta'_{t,ij}(n)}{(\sigma'_{t,ij}(n))^2}
    \Sigma'_t(n)(\boldsymbol e_i-\boldsymbol e_j)
    (\boldsymbol e_i-\boldsymbol e_j)^\top\Sigma'_t(n),
\end{equation}
where $(\sigma'_{t,ij}(n))^2$ and $\delta'_{t,ij}(n)$ are computed as in \cref{lem:expected-cov}, using $\Sigma'_t(n)$ in place of $\Sigma_{t-1}$.

Let $\mathcal L(0;\Sigma'_t(n))$ be the surrogate loss before adding another evaluation and $\mathcal L((i,j);\Sigma'_t(n))$ be the loss after adding $(i,j)$. The marginal score of $(i,j)$ is
\begin{equation}\label{eq:batch-score}
    \psi_{t,ij}(n)
    =
    \frac{
        \mathcal L(0;\Sigma'_t(n))
        -
        \mathcal L((i,j);\Sigma'_t(n))
    }{r_\jmath}.
\end{equation}
Using these scores, the batch-building procedure is given in \cref{alg:batch-surrogate}. We consider elastic round budgets, meaning that as long as some budget remains in the current round, the algorithm may choose any evaluation provided it satisfies the total budget constraint. If its cost exceeds the remaining round budget, the excess is deducted from the next round's budget. This ensures that the knapsack nature of the round budget does not bias the choice of evaluation type.

\begin{algorithm}[!htb]
\caption{Surrogate batch allocation}\label{alg:batch-surrogate}
\makeatletter
\def\theHALG@line{batchsurrogate.\arabic{ALG@line}}
\makeatother
\begin{algorithmic}[1]
\Require $\boldsymbol{\mu}_{t-1}$, $\Sigma_{t-1}$, round budget $B_t$,
remaining total budget $R_{\mathrm{rem}}$ (initially $R$),
and $\text{excess}$ from the previous round (initially zero)
\State $B_{\mathrm{rem}}\leftarrow B_t-\text{excess}$
\State $n\leftarrow 0$, \quad $\Sigma'_t(0)\leftarrow\Sigma_{t-1}$
\State Compute $(q_t,\boldsymbol b_t)$
\While{$B_{\mathrm{rem}}>0$ \textbf{and}
       $R_{\mathrm{rem}}\geq\min(r_0,r_1)$}
    \State Compute $\psi_{t,ij}(n)$ by \eqref{eq:batch-score}
    for all $(i,j)\in\mathcal I$ with $r_\jmath\leq R_{\mathrm{rem}}$
    \State Choose $(i,j)\in
    \arg\max_{(i,j)\in\mathcal I:\,r_\jmath\leq R_{\mathrm{rem}}}
    \psi_{t,ij}(n)$
    \State Update the covariance from $\Sigma'_t(n)$ to
    $\Sigma'_t(n+F^{ij})$ by \eqref{eq:batch-cov-update}
    \State $n\leftarrow n+F^{ij}$
    \State $B_{\mathrm{rem}}\leftarrow B_{\mathrm{rem}}-r_\jmath$,
    \quad $R_{\mathrm{rem}}\leftarrow R_{\mathrm{rem}}-r_\jmath$
\EndWhile
\State $\text{excess}\leftarrow\max\{0,-B_{\mathrm{rem}}\}$
\State \Return $n(t)=n$, $R_{\mathrm{rem}}$, and $\text{excess}$
\end{algorithmic}
\end{algorithm}

\subsection{The batched endgame}\label{sec:batching:kg}
Extending the KG algorithm to batches introduces two computational difficulties. First, it optimizes over possible batch allocations, which we address through sequential batch construction, as in \cref{sec:batching:surrogate}. Second, computing the myopic value of a given batch requires the outcome probabilities and posterior means for every possible combination of batch outcomes. We derive explicit expressions for these quantities given a batch outcome (analogous to \cref{lem:kg-update}), but they involve multivariate normal CDFs whose dimension grows with the batch size, making them expensive to evaluate numerically \citep{GenzBretz2009}. Moreover, a batch of $U$ evaluations has $2^U$ possible outcomes, each requiring these calculations.

The computational difficulty comes from considering all the ways the evaluations in a batch can jointly move the posterior means. We instead approximate the myopic value of a batch by decomposing it into the myopic values of sub-batches of at most $w$ evaluations. This retains interaction effects among up to $w$ evaluations while keeping the computation tractable: the value of each sub-batch requires normal CDFs of dimension at most $w$ and a sum over at most $2^w$ outcomes. Because the details are technical and not the focus of this paper, we leave the full decomposition and batch construction procedure, along with additional details on surrogate batch allocation, to the Appendix.

\section{Numerical experiments}\label{sec:numerics}
The numerical experiments serve two purposes: to illustrate what the algorithm does, which supplements the insights from the analysis, and to investigate how well it performs. \Cref{sec:num:alloc} serves the first, showing how evaluations are allocated as we learn more, how the mix of ratings and comparisons shifts over the budget, and how these are influenced by the endgame. \Cref{sec:num:internal} serves both, comparing our method to versions of itself restricted to one evaluation type or to a fixed schedule of the two, which tests the value of combining the two evaluation types and how well the method balances between them. To further investigate our method's performance, \Cref{sec:num:external} compares it to external (and internal) benchmarks, \cref{sec:num:stress} stress-tests the Gaussian approximation and the method under misspecification, and \cref{sec:num:batch} measures the impact of batching.

\subsection{Setup}\label{sec:num:setup}

\paragraph{Configurations.} We consider $I \in \{10, 20, 40, 80, 160\}$ and selection sizes $k \in \{1,\ \sqrt{0.1\,I},\ 0.1\,I\}$ (rounded). The prior is exchangeable with $\boldsymbol{\mu}_0 = \boldsymbol{0}$, $\Sigma_0 = 10\cdot\operatorname{Id}_I$, and the rating threshold is $\theta_0 = 0$ (in expectation half of the items get thumbs-up ratings). Evaluator accuracies are $75\%$ for ratings and $85\%$ for comparisons, calibrated through \cref{rem:calibration} to $\beta_0 = 0.316$ and $\beta_1 = 0.439$. We fix the cost of a rating at $r_0 = 1$ and consider comparison costs $r_1 \in \{2, 3\}$. The total budget is $R = 4I$. These are the base configurations. Later subsections also test modifications of them to demonstrate that the results are not brittle to these choices.

\paragraph{Endgame.} In each configuration, we run the switching procedure of \cref{sec:switching-rule} with $M = \num{1000}$ to choose the switching point from \textsc{SA} to \textsc{KG}. The candidate switching points are $\nu = 0, 0.05, \ldots,0.95, 1$.

\paragraph{Policies.} Alongside SE, we run modified versions of it to gain insight into the algorithm's behavior as discussed above. To keep track of the modifications of SE, we specify each aspect changed by an optional argument. The first optional argument is the switching point. $\text{SE}(\text{sw}=\nu)$ switches from \textsc{SA} to \textsc{KG} once the share $\nu$ of the budget has been spent, so for example, $\text{SE}(\text{sw}=1)$ is \textsc{SA}. The second optional argument is which evaluation type is allowed at each point of the budget: rc (rate then compare). SE itself may mix the two types throughout, whereas $\text{SE}(\text{rc}=u)$ is restricted to only ratings until a proportion $u$ of the budget is spent and only comparisons after, so for example, $\text{SE}(\text{rc}=1)$ is ratings only for the entire budget.

To measure performance, we compare against three benchmarks: a slightly modified \textsc{RateRank} of \citet{LiZhangLi2018RatingRankingTopk}, which allocates based on myopic information gain about the top-$k$ set per unit cost, the KG policy with random fallback when it gets ``stuck'', and \textsc{KG} with \textsc{SA} as the fallback, KG(\textsc{SA}), which corresponds to $\text{SE}(\text{sw}=0)$.

\paragraph{Simulation protocol.} For each configuration, we run \num{1000} independent replicates where $\boldsymbol{\theta}$ is drawn from the prior and for each $\boldsymbol{\theta}$ the same stream of evaluation outcomes is drawn across policies. We report the average realized opportunity cost, which we refer to as the EOC (technically this is an estimator of the EOC). Posterior updates use variational inference with a multivariate Gaussian variational family. \textsc{RateRank} is run for $I \le 40$ only, due to its computational cost. Figures that demonstrate the algorithm's behavior are presented on the $I=40$, $k=2$, $r_1=3$ configuration. The corresponding
figures for all other configurations are provided in the Appendix.

\subsection{What the policy does}\label{sec:num:alloc}

\begin{figure}[!htb]
\centering
\includegraphics[width=0.9\textwidth]{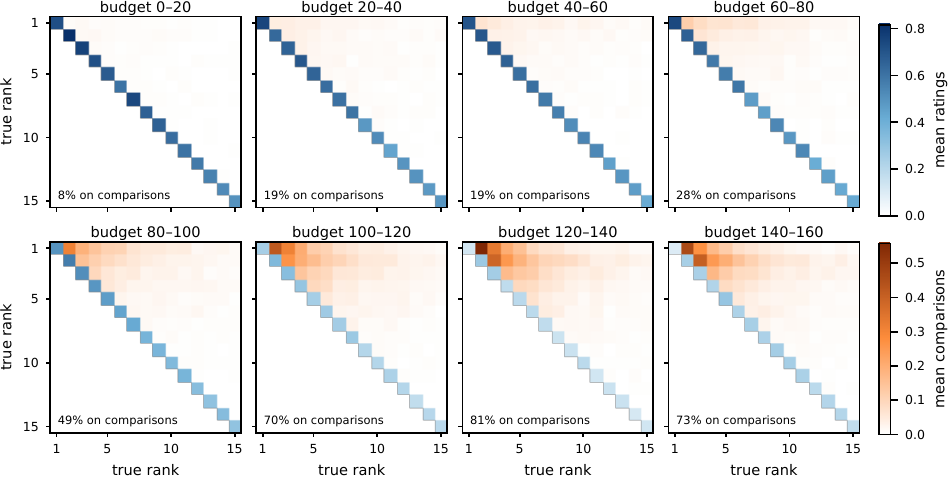}
\caption{Mean \textsc{SA} allocation by true rank in each eighth of the budget
($I=40$, $k=2$, $r_1=3$), with ratings on diagonal and comparisons off diagonal.
The lower triangle reports the percent of budget spent on comparisons.}
\label{fig:alloc}
\end{figure}

\begin{figure}[!htb]
\centering
\includegraphics[height=0.3\textwidth]{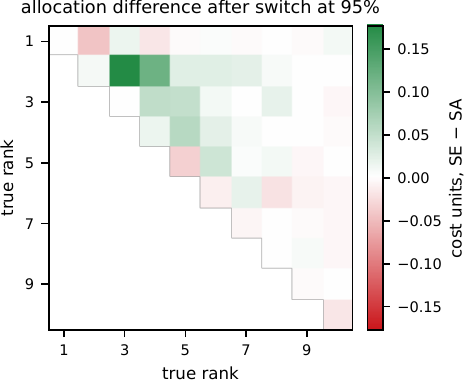} \hspace{30pt}
\includegraphics[height=0.3\textwidth]{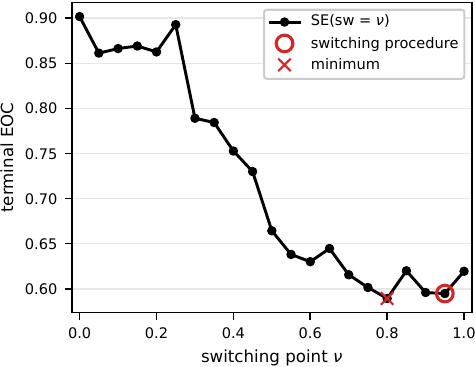}
\caption{(a) Average difference of the last $5\%$ of the budget allocation between SE and SA ($I=40$, $k=2$,
$r_1=3$). Green denotes SE spends more, red it
spends less. (b) Terminal EOC of $\text{SE}(\text{sw}=\nu)$ against the switching point.}
\label{fig:endgame}
\vspace{-2pt}
\end{figure}

To get insight into the \textsc{SA} allocations, \Cref{fig:alloc} shows the average allocation of \textsc{SA} at each stage of the horizon. The rating-then-comparison pattern emerges. In the first half of the budget, mostly ratings are allocated, nearly uniformly across the items and then concentrating towards the top items. In the second half, comparisons gradually take over, concentrating on items near the top-$k$ boundary.
\vspace{-5pt}
\vspace{-5pt}
To see how the endgame of SE differs from \textsc{SA}, we consider the allocation difference between SE and \textsc{SA} \emph{after} SE switches, which on this instance is at $95\%$ of the budget. \Cref{fig:endgame}a shows the difference in where the budget is spent. Both focus on the top-$k$ boundary, but \textsc{KG} concentrates much more on comparing items $2$ and $3$, while \textsc{SA} spreads more across nearby items as it is still looking for long-term uncertainty reduction around the top-$k$ threshold. This is the value of switching to \textsc{KG}: at the end of the horizon we want to narrow in on evaluations with potential for immediate improvement.

\Cref{fig:endgame}b shows the terminal EOC of $\text{SE}(\text{sw}=\nu)$ which, for illustrative purposes, we compute for all $\nu$. The marginal analysis stops at $\nu = 0.95$, while the minimum is at $\nu = 0.8$, showing that it need not give the optimal switching point. However, the two differ by less than $1\%$ of the EOC, well within one standard error of the estimated EOC. In general, as $\nu$ moves back the EOC first dives to the optimal point and then increases, and the same pattern appears in the other configurations. In the majority of the configurations, the switching procedure identifies the minimum.

\subsection{The value of combining the two evaluation types}\label{sec:num:internal}

\begin{figure}[!htb]
\centering
\includegraphics[width=0.70\textwidth]{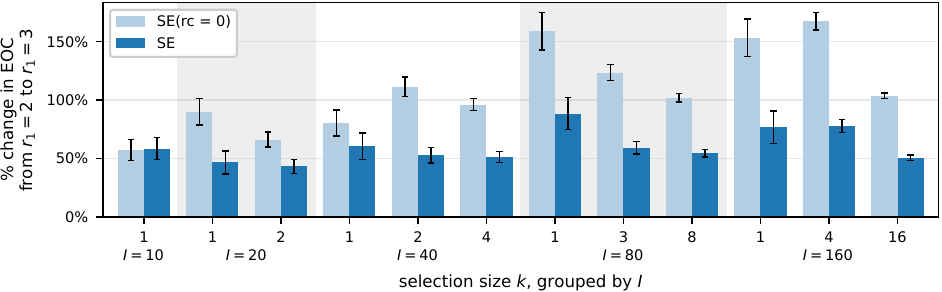}
\caption{Percent change in terminal EOC of SE and $\text{SE}(\text{rc}=0)$ when the
comparison cost goes from $r_1=2$ to $r_1=3$, each policy measured
against itself at $r_1=2$. Error bars are one standard error.}
\label{fig:compensation}
\end{figure}

\begin{figure}[!htb]
\centering
\includegraphics[width=0.70\textwidth]{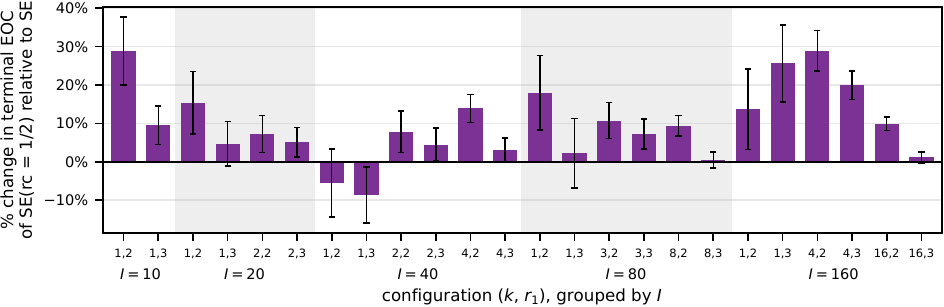}
\caption{Percent difference in terminal EOC of $\text{SE}(\text{rc}=1/2)$ relative to SE, all configurations. Above zero SE is better,
  below zero $\text{SE}(\text{rc}=1/2)$ is. Error bars are one standard error.}
\label{fig:block}
\end{figure}

We next focus on the value of combining the two evaluation types and test how well SE balances between them by comparing SE to the restricted versions of itself defined in \cref{sec:num:setup}. We find that rating only ($\text{SE}(\text{rc}=1)$) is not competitive, with terminal EOC between $1.34$ and $15.9$ times that of SE across the configurations. As discussed in the introduction, ratings are not effective in separating items whose utilities are close together, which is key for narrowing in on the terminal selection. Conversely, when $r_1 = 2$, SE and $\text{SE}(\text{rc}=0)$ are quite comparable, that is, ratings do not add much value. Thus, to see the value in combining the two input types, one of the main premises of this paper, we look at the change in EOC of SE and $\text{SE}(\text{rc}=0)$ as the cost of a comparison goes from $r_1 = 2$ to $r_1 = 3$, shown in \Cref{fig:compensation}. SE can adapt to this, while $\text{SE}(\text{rc}=0)$ must still rely on comparisons alone. We see that having the ability to rate can be a major improvement. The EOC of $\text{SE}(\text{rc}=0)$ increases by between $57\%$ and $167\%$, whereas that of SE increases by only between $43\%$ and $88\%$, since it can substitute the now more expensive comparisons with ratings.
\vspace{-2pt}
\vspace{-10pt}
\noindent Next, we investigate the rate-then-compare intuition, consistent with the allocation patterns in \Cref{fig:alloc}. To do this, we compare SE to $\text{SE}(\text{rc}=1/2)$, which still allocates by our method but simplifies the problem by not having to choose between the two evaluation types at each round. \Cref{fig:block} shows the percent difference in its EOC relative to SE. As expected, SE comes out ahead in $22$ of the $24$ configurations, by up to $29\%$, and loses in two, by $5.5\%$ at $I=40$, $k=1$, $r_1=2$ and by $8.6\%$ at $I=40$, $k=1$, $r_1=3$, which are both well within two standard errors of zero. This shows that SE's flexibility to allocate dynamically and balance between the two adds value, but in some cases a simple heuristic performs similarly. However, one can choose settings based on the budget, $r_0$, $r_1$, $\beta_0$, $\beta_1$, and $\theta_0$, such that a fixed heuristic would do much worse, but SE will still do well by adapting to the configuration.
\vspace{-4pt}
\vspace{-16pt}
\subsection{Comparison to external benchmarks}\label{sec:num:external}
\begin{figure}[!htb]
\centering
\includegraphics[width=0.9\textwidth]{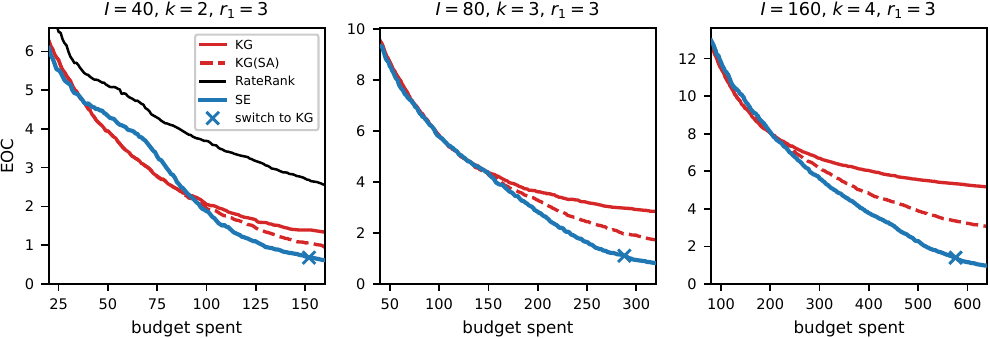}
\vspace{-2pt}
\caption{EOC against budget spent.}
\label{fig:kgtraj}
\end{figure}

We next compare the performance of our method to \textsc{RateRank} and the two KG variants. \Cref{fig:kgtraj} plots the EOC against the budget spent on three configurations. In addition to the terminal EOCs, this figure highlights two interesting differences in how the alternative approaches get there. First, we can see the ``forward-looking'' property of SE. During a large part of the early budget spend, its EOC is higher than that of KG(\textsc{SA}): SE chooses evaluations that do not maximize immediate EOC improvement, but enable it to surpass the KG policies later in the horizon. Towards the middle of the horizon, this early investment starts to pay off, and SE pulls ahead. The second takeaway from the figure is the separation between \textsc{KG} and KG(\textsc{SA}), which allows us to clearly see where, on average, in the budget \textsc{KG} gets stuck and how much it suffers from it.

\begin{figure}[!htb]
\centering
\includegraphics[width=0.9\textwidth]{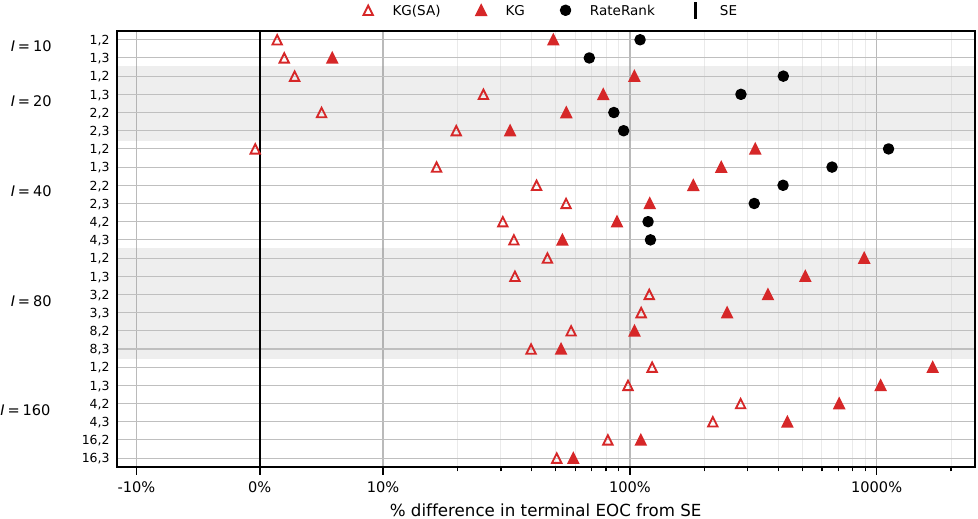}
\caption{Percent difference in terminal EOC from SE. Rows are labeled $k, r_1$ and grouped by $I$.}
\label{fig:external}
\end{figure}

\begin{figure}[!htb]
\centering
\includegraphics[width=0.9\textwidth]{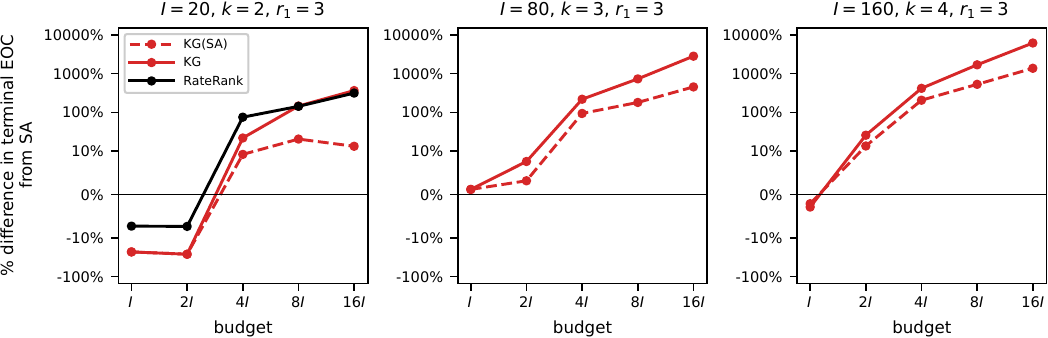}
\caption{Percent difference in terminal EOC from \textsc{SA} against the total budget.}
\label{fig:longrun}
\vspace{-4pt}
\end{figure}

\vspace{-5pt}
\noindent \Cref{fig:external} shows the terminal EOC of the competing policies across all configurations, as the percent difference from SE. Remarkably, SE tends to substantially outperform KG(\textsc{SA}): KG(\textsc{SA}) has a higher terminal EOC in $23$ of the $24$ configurations, by more than $30\%$ in $16$ of them, and up to $281\%$ at $I=160$, $k=4$, $r_1=2$. We say this is remarkable since KG(\textsc{SA}) is a strong benchmark for our problem: the budgets are small, a regime where \textsc{KG} performs well, and we equip it with the best available fallback rule. Furthermore, the fallback is invoked precisely in states where \textsc{SA}'s exploratory nature is most valuable to \textsc{KG}, where no single evaluation can directly change the current top-$k$ set. This shows that while the most obvious drawback of KG being myopic is that it can get stuck, that alone is not the entire problem. The eagerness to directly improve the selected $k$ throughout the budget results in evaluation choices with limited information contribution to the ultimate selection. We also note that KG(\textsc{SA}) outperforms \textsc{KG} in every configuration, by $4\%$ to $701\%$, and for the configurations where it is computationally practical, \textsc{RateRank} does considerably worse than our method (and KG), with $1.7$ to $12$ times higher terminal EOC.
\vspace{-4pt}
\vspace{-8pt}
\noindent While a large motivating factor in this work is small budgets, there are settings of crowdsourcing from human evaluators where budgets far beyond $4I$ are realistic. Thus we compare the benchmarks to our method in this regime, with no switching (\textsc{SA} throughout), since as the budget grows the impact of the endgame and the value of switching decrease. \Cref{fig:longrun} shows the percent difference between the benchmarks and \textsc{SA} as a function of the budget spent, with the budget extended to $16I$. As expected, at the extremely small budget $R = I$, \textsc{KG} is as good as \textsc{SA} or better, but as the budget grows the advantage of \textsc{SA} grows substantially with it, which is to be expected, as this is the regime where the myopia of \textsc{KG} hurts more.

\subsection{Robustness to the Gaussian approximation and to misspecification}\label{sec:num:stress}

\begin{figure}[!htb]
\centering
\includegraphics[width=0.9\textwidth]{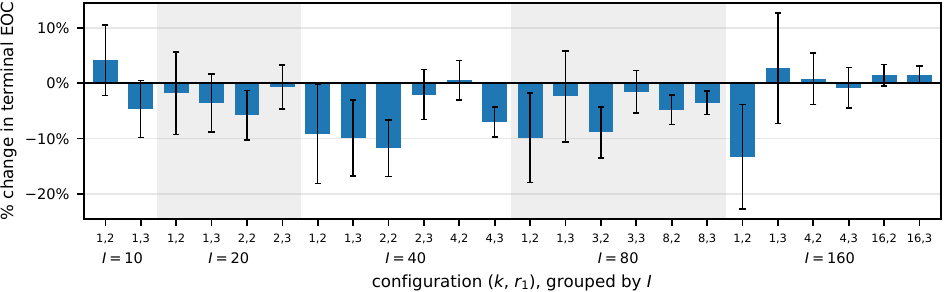}
\caption{\% difference in terminal EOC between the Gaussian and exact posterior.}
\label{fig:exact}
\end{figure}

We next stress-test the effect of \cref{approx:gaussian} on SE. In the absence of a conjugate prior for the probit likelihood that is amenable to analysis, this approximation replaces the posterior with a multivariate Gaussian fitted by variational inference, and the expected mismatch loss is computed with respect to it. To check whether this significantly alters the performance of the method, we run SE with the exact posterior in place of the Gaussian one. The posterior is represented by \num{2000} draws of a Gibbs sampler in each round, and everything SE depends on, the outcome probabilities, the expected posterior covariance, the threshold, and the expected mismatch loss, is computed as a Monte Carlo average over those draws, so that the Gaussian approximation does not enter anywhere. \Cref{fig:exact} plots the percent change in the terminal EOC of SE when the Gaussian approximation is replaced by the exact posterior, in all configurations. It shows that the approximation negatively affects the terminal EOC by $4\%$ on average across all configurations and by at most $13\%$, which is quite mild compared to SE's advantage over the baselines.

\vspace{-10pt}
\noindent SE assumes perfect knowledge of the parameters and that evaluators are homogeneous (in the base configurations, accuracies of $75\%$ and $85\%$ and $\theta_0 = 0$). We next test the method under misspecification when the assumed parameter values are wrong and when evaluators are heterogeneous. We compare the resulting performance with unbiased SE, which runs with the correct assumptions. We consider the following misspecifications:

\begin{itemize}[leftmargin=*]
    \item[--] \textbf{Evaluator heterogeneity.} Each evaluation comes from a different evaluator. We consider heterogeneity in accuracy and in rating threshold separately, and in each, an evaluator is one of two types with equal probability.
      \begin{enumerate}[leftmargin=*]
          \item \textbf{Accuracies.} The rating accuracy is $70\%$ or $80\%$ and the comparison accuracy is $80\%$ or $90\%$.
          \item \textbf{Threshold.} The threshold is $\theta_0 = -1.58$ or $\theta_0 = 1.58$, half a prior standard deviation below or above the assumed threshold.
      \end{enumerate}
    \item[--] \textbf{Evaluator accuracy.} The true rating and comparison accuracies are $65\%$ and $75\%$, or $85\%$ and $90\%$.
    \item[--] \textbf{Rating threshold.} The true threshold is $\theta_0 = -1.58$ or $\theta_0 = 1.58$.
\end{itemize}
\Cref{tab:misspec} reports the percent increase in terminal EOC of SE relative to unbiased SE. In $16$ of the $18$ cases it is between $-9\%$ and $27\%$. While this decrease in performance (on average) is not negligible, it is small relative to SE's advantage even over KG(\textsc{SA}) in the corresponding configurations ($20\%$, $111\%$, and $217\%$).
\vspace{-5pt}
\begin{table}[!htb]
\centering
\caption{\% increase in the EOC of SE relative to unbiased SE (standard error), $r_1=3$.}
\label{tab:misspec}
\input{figs/table1_robust.tex}
\end{table}
\vspace{-10pt}
\noindent The setting we consider where misspecification has a substantial impact is when the true accuracies are $85\%$ and $90\%$. There, the terminal EOC of SE is $41\%$ and $74\%$ higher than unbiased SE at $I = 80$ and $160$. We note that this misspecification is extreme: $\beta_0$ is twice what SE assumes ($0.621$ instead of $0.316$), and $\beta_1$ is $1.6$ times ($0.688$ instead of $0.439$). To understand what drives the difference in performance, \Cref{tab:alloc_misspec} shows for the $I = 80$ configuration how SE and unbiased SE split spending between ratings and comparisons over the course of the budget. Unbiased SE shows the rating-then-comparison pattern of \Cref{fig:alloc} even more sharply than in the base configurations: it spends the first half of the budget on ratings, and then turns almost entirely to comparisons, making good use of the more accurate evaluations (ratings in particular). In contrast, SE mixes in comparisons early, and its shift to more comparisons comes later and is less pronounced. The mechanism driving this is the following. Consider a good item, one prior standard deviation above the mean. With the more accurate evaluators, a rating results in a thumbs-up $97.5\%$ of the time, while SE assumes it is $84\%$. SE sees multiple good items receive several thumbs-up, quickly becomes overconfident that they are good, and starts comparing them early. As a result, it does not rate enough of the other items in the first half of the budget, and later goes back to rating them.
\vspace{-5pt}
\begin{table}[!htb]
\centering
\caption{\% of each eighth of the budget spent on comparisons. Accuracies: $85\%$ and $90\%$ ($I=80$, $k=3$, $r_1=3$).}
\label{tab:alloc_misspec}
\input{figs/table2_allocation_85_90.tex}
\end{table}
\vspace{-15pt}
\noindent Overall, misspecification generally has a negative impact on SE's performance, which can be substantial when the misspecification is extreme, but the loss is small relative to SE's advantage over the benchmarks.

\subsection{Impact of batching}\label{sec:num:batch}

\begin{figure}[!htb]
\centering
\includegraphics[width=0.9\textwidth]{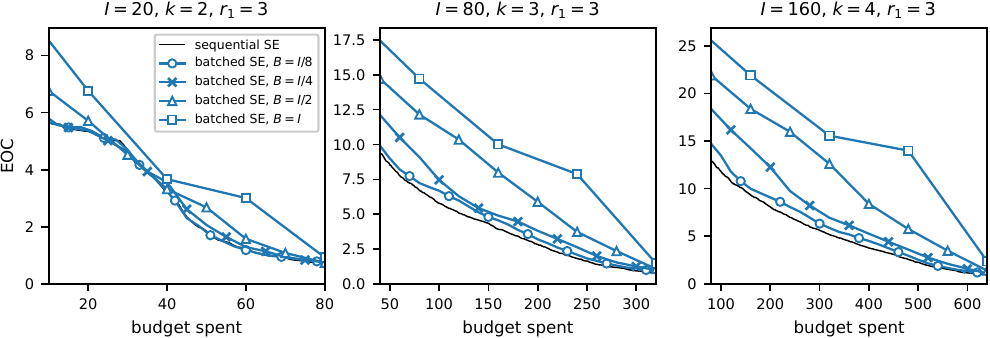}
\caption{EOC against the budget spent.}
\label{fig:batch}
\end{figure}

Finally, we test the batched version of the method with the batch endgame parameter $w = 3$ and the switching round chosen by the procedure of \cref{sec:switching-rule} applied to rounds. We run it on three configurations with the total budget of $4I$ split into round budgets of $B = I/8$, $I/4$, $I/2$, and $I$ (rounded up), that is, from $32$ rounds down to $4$. \Cref{fig:batch} shows how EOC improves as the budget is spent. The impact of batching is small at $B = I/8$ with a $5\%$ to $8\%$ increase in terminal EOC relative to SE, and grows with the round budget, with $31\%$ to $145\%$ at $B = I$. Even then, with only four total rounds, batched SE has a lower terminal EOC than sequential KG(\textsc{SA}) on the $I=80$ and $I=160$ configurations.
\vspace{-5pt}
\vspace{-15pt}
\section{Concluding remarks}\label{sec:conc}
For adaptive top-$k$ selection, we consider a unifying generative model for ratings (thumbs-up/down) and pairwise comparisons and allow the two input types to differ in both accuracy and cost. We replace the resulting intractable dynamic program with a novel surrogate-based allocation method. Despite consisting of a sequence of state-based static optimization problems, it is effectively forward-looking by rewarding decision-relevant uncertainty reduction. Based on the surrogate, we construct an allocation policy whose decision rule decomposes into interpretable parts that give insight into exactly why a particular evaluation was chosen. For the endgame regime, we switch to the knowledge gradient policy, with a criterion for when to do so, and extend the procedure to the batched evaluation setting. Finally, we show that our method performs well in simulations, stress-test the approximation made in the analysis, and test the method under model misspecification.

There are a number of exciting possible extensions for future work. One can consider heterogeneous evaluators whose $\beta$'s and $\theta_0$ are unknown and learned from the evaluations. Additionally, one may consider richer evaluation types. Our model and analysis extend to ratings with any number of categories using an ordered probit model. However, considering partial rankings of a subset larger than two items would require additional work. We also hope that our surrogate method can be further improved upon and applied to additional Bayesian sequential learning settings. For example, in \citet{GoldbergerRudi2026simple} we develop a variant of it for the classical setting of top-$k$ selection with untransformed evaluations. There, it has strong theoretical properties and performance guarantees, and performs very well in simulation against a strong set of benchmarks from the well-developed ranking and selection and best-arm identification literatures.

\ACKNOWLEDGMENT{We thank Enver Y\"ucesan for his helpful comments that improved the paper.}

\bibliography{refs}

\begin{APPENDICES}
\clearpage

\etocdepthtag.toc{appendix} \etocsettagdepth{main}{none}\etocsettagdepth{appendix}{subsection} \etocsettocstyle{\noindent{\AppendixHeadSize\textbf{Appendix contents}}\par\vskip4pt}{\vskip8pt} \tableofcontents

\section{Batching extension}\label{app:batching}
In this section we present the additional technical details regarding the batching extension of \cref{sec:batching}.

\subsection{Surrogate batch construction}\label{app:batching:surrogate}

Throughout the construction of a batch, we hold $\boldsymbol{\mu}_{t-1}$, $S_k^*(\boldsymbol{\mu}_{t-1})$, and $(q_t,\boldsymbol b_t)$ fixed, as in \cref{sec:surrogate:program}. For a candidate batch $n$, its expected posterior covariance is
\begin{equation*}
\overline{\Sigma}_t(n,p_{t-1})
:=
\E\!\FitDelims{[}{]}{
\Cov(\boldsymbol\theta\mid p_t)
\,\FitMid{|}\,n,p_{t-1}}.
\end{equation*}
Here $p_t$ denotes the posterior after observing the outcomes of $n$ and the expectation is over all possible outcomes of the batch. For a round budget $B_t$, substituting this covariance into the surrogate gives the integer program
\begin{equation}\label{eq:batch-surrogate-program}
\begin{aligned}
\min_n\quad &
\sum_{\ell=1}^I
\E_{\boldsymbol\theta\sim N(\boldsymbol\mu_{t-1},\overline\Sigma_t(n,p_{t-1}))}
\!\FitDelims{[}{]}{
\bigl(\theta_\ell-a_{q_t,\boldsymbol b_t}(\boldsymbol\theta)\bigr)^+
}\\
\text{s.t.}\quad &
\sum_{(i,j)\in\mathcal I}r_\jmath n_{ij}\le B_t,
\qquad n_{ij}\in\mathbb Z_{\ge0}\quad\text{for all $(i,j)\in\mathcal I$.}
\end{aligned}
\end{equation}
As in \eqref{eq:surrogate-loss}, we have dropped the terms that do not depend on $n$. We approximately solve \eqref{eq:batch-surrogate-program} by constructing the batch one evaluation at a time, as in \cref{alg:batch-surrogate}. To choose each additional evaluation, we approximate the expected posterior covariance in the objective by $\Sigma'_t(n)$ using \eqref{eq:batch-cov-update}.

Consider adding an evaluation $(i,j)$ to $n$. If the outcomes of $n$ were observed, the resulting posterior $p_t$ would give expected covariance $\overline\Sigma_t((i,j),p_t)$ after the additional evaluation. Averaging over the outcomes of $n$ gives
\begin{equation*}
\overline\Sigma_t(n+F^{ij},p_{t-1})
=
\E\!\FitDelims{[}{]}{
\overline\Sigma_t((i,j),p_t)
\,\FitMid{|}\,n,p_{t-1}}.
\end{equation*}
Here $\overline\Sigma_t((i,j),p_t)$ averages over the two outcomes of $(i,j)$ given $p_t$, while the outer expectation averages over the outcomes of the already committed batch $n$.

We approximate this expectation by fixing the covariance after $n$ at $\Sigma'_t(n)$. We then compute the expected covariance after adding $(i,j)$ by averaging over its two possible outcomes under $N(\boldsymbol\mu_{t-1},\Sigma'_t(n))$, giving
\begin{equation*}
\overline\Sigma_t(n+F^{ij},p_{t-1})
\approx
\overline\Sigma_t\!\FitParens{(i,j),N(\boldsymbol\mu_{t-1},\Sigma'_t(n))}
=
\Sigma'_t(n+F^{ij}),
\end{equation*}
where the equality follows from \cref{lem:expected-cov} and gives \eqref{eq:batch-cov-update}. The first update is exact because $n=0$ and $\Sigma'_t(0)=\Sigma_{t-1}$. Note that for the same batch matrix $n$, the approximation $\Sigma'_t(n)$ can differ depending on the order in which evaluations are committed.

\subsection{The batched endgame}\label{app:batching:kg}

In this subsection, it is convenient to use an alternative notation for a batch. Above, a batch was represented by its count matrix $n(t)$, which records how many times each evaluation $(i,j)$ is conducted in round $t$. Here it will be useful to distinguish between repeats of the same evaluation, so we fix an ordering of the evaluations, writing a batch of $U$ evaluations as $(i_1,j_1),\ (i_2,j_2),\ \ldots,\ (i_U,j_U),$ where each $(i_u,j_u) \in \mathcal I$ and $\jmath_u:=\min(j_u,1)$. The associated count matrix is $n_{ij}(t) = \sum_{u=1}^{U} \1\{(i_u,j_u) = (i,j)\}$ for each $(i,j) \in \mathcal{I}$.

Consider a candidate feasible batch $(i_1,j_1),\ldots,(i_U,j_U)$ with count matrix $n$. Let $X_u\in\{-1,1\}$ denote the outcome of the $u$th evaluation and define $\boldsymbol{X}:=(X_1,\ldots,X_U)^\top.$ For an outcome vector $\boldsymbol x$, write $\boldsymbol\mu_t(\boldsymbol x\mid n,p_{t-1})=\E[\boldsymbol\theta\mid\boldsymbol X=\boldsymbol x,n,p_{t-1}]$.

\subsubsection{The myopic value of a batch}\label{app:kg:mval}

We begin by writing the myopic value function of \cref{sec:model:kg} using the notation introduced above. The myopic value function which the KG policy maximizes each round (reformulated from \cref{subsec:kgdiff}) is
\begin{equation}\label{eq:endgame-myopic-value}
V_{\text{myopic}}(n\mid p_{t-1})
=
\sum_{\boldsymbol{x}\in\{-1,1\}^U}
\Pp(\boldsymbol{X}=\boldsymbol{x}\mid n,p_{t-1})
\smashoperator{\sum_{i\in S_k^*(\boldsymbol{\mu}_t(\boldsymbol{x}\mid n,p_{t-1}))}}
\mu_{t,i}(\boldsymbol{x}\mid n,p_{t-1})
-\smashoperator[r]{\sum_{i\in S_k^*(\boldsymbol{\mu}_{t-1})}}\mu_{t-1,i},
\end{equation}
with $V_{\text{myopic}}(0\mid p_{t-1})=0.$ \Cref{prop:mu_plus}, which applies the results of \citet{Mullahy2017}, gives explicit expressions for the outcome-dependent quantities $\Pp(\boldsymbol{X}=\boldsymbol{x}\mid n,p_{t-1})$ and $\boldsymbol{\mu}_t(\boldsymbol{x}\mid n,p_{t-1})$.

\begin{proposition}\label{prop:mu_plus}
Define $C\in\mathbb R^{U\times U}$ by
\begin{equation*}
C_{uu'}
=
\beta_{\jmath_u}\beta_{\jmath_{u'}}
(\boldsymbol{e}_{i_u}-\boldsymbol{e}_{j_u})^\top
\Sigma_{t-1}
(\boldsymbol{e}_{i_{u'}}-\boldsymbol{e}_{j_{u'}})
+
\1\{u=u'\},
\qquad
u,u'=1,\ldots,U.
\end{equation*}
For any $\boldsymbol{x}\in\{-1,1\}^U$, let $D:=\operatorname{diag}\!\FitParens{\sqrt{C_{11}},\ldots,\sqrt{C_{UU}}}$, $D_{\boldsymbol{x}}:=\operatorname{diag}(x_1,\ldots,x_U)$, $\widetilde C_{\boldsymbol{x}}:=D_{\boldsymbol{x}}D^{-1}CD^{-1}D_{\boldsymbol{x}}$, and $\boldsymbol{\alpha}(\boldsymbol{x}):= \FitParens{\alpha_1(\boldsymbol{x}),\ldots,\alpha_U(\boldsymbol{x})}^\top$ where $\alpha_u(\boldsymbol{x}) = \frac{x_u\,\beta_{\jmath_u}\,(\mu_{t-1,i_u}-\mu_{t-1,j_u})}{\sqrt{C_{uu}}}$ for each $u$. Let $\Phi_U(\cdot;\widetilde C_{\boldsymbol{x}})$ denote the CDF of a mean-zero $U$-variate normal vector with covariance matrix $\widetilde C_{\boldsymbol{x}}$, let $\boldsymbol{\alpha}_{-u}(\boldsymbol{x})$ denote the vector $\boldsymbol{\alpha}(\boldsymbol{x})$ with its $u$th component removed, and let $(\widetilde C_{\boldsymbol{x}})_{-u\mid u} := (\widetilde C_{\boldsymbol{x}})_{-u,-u} - (\widetilde C_{\boldsymbol{x}})_{-u,u}(\widetilde C_{\boldsymbol{x}})_{u,-u}.$ Finally, define $\boldsymbol{\lambda}_{\boldsymbol{x}}\in\mathbb R^U$ by
\begin{equation*}
(\boldsymbol{\lambda}_{\boldsymbol{x}})_u
=
\frac{x_u}{\sqrt{C_{uu}}}
\cdot
\frac{
\phi(\alpha_u(\boldsymbol{x}))
\,
\Phi_{U-1}\!\Big(
\boldsymbol{\alpha}_{-u}(\boldsymbol{x})-(\widetilde C_{\boldsymbol{x}})_{-u,u}\alpha_u(\boldsymbol{x});
(\widetilde C_{\boldsymbol{x}})_{-u\mid u}
\Big)
}{
\Phi_U(\boldsymbol{\alpha}(\boldsymbol{x});\widetilde C_{\boldsymbol{x}})
},
\qquad
u=1,\ldots,U.
\end{equation*}
For $U=1$, the $\Phi_{U-1}$ factor is $1$. Then
\begin{align}
\text{(i)}\quad &
\Pp(\boldsymbol{X}=\boldsymbol{x}\mid n,p_{t-1})
=\Phi_U(\boldsymbol{\alpha}(\boldsymbol{x});\widetilde C_{\boldsymbol{x}}),
\label{eq:outcomeprob}\\
\text{(ii)}\quad &
\boldsymbol{\mu}_t(\boldsymbol{x}\mid n,p_{t-1})
=
\boldsymbol{\mu}_{t-1}
+
\sum_{u=1}^U
\beta_{\jmath_u}
(\boldsymbol{\lambda}_{\boldsymbol{x}})_u
\Sigma_{t-1}(\boldsymbol{e}_{i_u}-\boldsymbol{e}_{j_u}).
\label{eq:mu-after-batch}
\end{align}
\end{proposition}

Substituting \eqref{eq:outcomeprob} and \eqref{eq:mu-after-batch} into \eqref{eq:endgame-myopic-value} gives an explicit expression for $V_{\text{myopic}}(\cdot\mid p_{t-1})$. Since the general formula is not easy to read in its full $U$-dimensional form, we first work out the cases $U=1$ and $U=2$. These examples also point to the intractability for larger batches. Exact evaluation of \eqref{eq:endgame-myopic-value} requires a sum over $2^U$ outcome vectors, with a $U$-dimensional multivariate normal CDF in each term. These CDFs must be evaluated numerically, which becomes expensive as $U$ grows \citep{GenzBretz2009}. This motivates the approximate value function introduced in \cref{sec:kg:decw}.

\paragraph{Single evaluation ($U=1$).}
Consider a batch with a single evaluation $(i_1,j_1)=(i,j)$. Then $C_{11}=1+\beta_{\jmath}^2\sigma_{t-1,ij}^2$ and $\alpha_1(x)=x\,\beta_{\jmath}(\mu_{t-1,i}-\mu_{t-1,j})/\sqrt{C_{11}}$ for $x\in\{-1,1\}$, so
\begin{equation*}
\begin{aligned}
\Pp(X=x\mid n,p_{t-1})&=\Phi(\alpha_1(x)),
\quad
\lambda_x=\frac{x\,\phi(\alpha_1(x))}{\Phi(\alpha_1(x))\sqrt{C_{11}}},\\
\boldsymbol{\mu}_t(x\mid n,p_{t-1})
&=
\boldsymbol{\mu}_{t-1}
+
\frac{x\,\beta_{\jmath}\,\phi(\alpha_1(x))}{\Phi(\alpha_1(x))\sqrt{C_{11}}}\,
\Sigma_{t-1}(\boldsymbol{e}_i-\boldsymbol{e}_j).
\end{aligned}
\end{equation*}

\paragraph{Two evaluations ($U=2$).}
Consider a batch with two evaluations $(i_u,j_u)$, $u=1,2$. Then $C_{uu}=1+\beta_{\jmath_u}^2\sigma_{t-1,i_uj_u}^2$, $C_{12}=\beta_{\jmath_1}\beta_{\jmath_2}(\boldsymbol{e}_{i_1}-\boldsymbol{e}_{j_1})^\top\Sigma_{t-1}(\boldsymbol{e}_{i_2}-\boldsymbol{e}_{j_2})$, and $\alpha_u(\boldsymbol{x})=x_u\,\beta_{\jmath_u}(\mu_{t-1,i_u}-\mu_{t-1,j_u})/\sqrt{C_{uu}}$, and the correlation matrix $\widetilde C_{\boldsymbol{x}}$ has off-diagonal entry $x_1x_2\rho_{12}$ with $\rho_{12}:=C_{12}/\sqrt{C_{11}C_{22}}$. Hence, for $\boldsymbol{x}=(x_1,x_2)\in\{-1,1\}^2$, $\Pp(\boldsymbol{X}=\boldsymbol{x}\mid n,p_{t-1})=\Phi_2\!\FitParens{\alpha_1(\boldsymbol{x}),\alpha_2(\boldsymbol{x});x_1x_2\rho_{12}}$ and, for $\{u,u'\}=\{1,2\}$,
\begin{equation*}
(\boldsymbol{\lambda}_{\boldsymbol{x}})_u
=
\frac{x_u}{\sqrt{C_{uu}}}
\cdot
\frac{
\phi(\alpha_u(\boldsymbol{x}))
\,
\Phi\!\FitParens{
\frac{\alpha_{u'}(\boldsymbol{x})-x_1x_2\rho_{12}\alpha_u(\boldsymbol{x})}
{\sqrt{1-\rho_{12}^2}}
}
}{
\Phi_2\!\FitParens{\alpha_1(\boldsymbol{x}),\alpha_2(\boldsymbol{x});x_1x_2\rho_{12}}
},
\quad
\boldsymbol{\mu}_t(\boldsymbol{x}\mid n,p_{t-1})
=
\boldsymbol{\mu}_{t-1}
+
\sum_{u=1}^2
\beta_{\jmath_u}
(\boldsymbol{\lambda}_{\boldsymbol{x}})_u
\Sigma_{t-1}(\boldsymbol{e}_{i_u}-\boldsymbol{e}_{j_u}).
\end{equation*}

\subsubsection{Decomposing the myopic batch value}\label{sec:kg:decw}

To obtain a tractable approximation of $V_{\text{myopic}}(\cdot\mid p_{t-1})$ when $U$ is large, we first decompose the myopic value of the batch into terms, one for each subset of the batch, and refer to the size of the subset (how many evaluations are in it) as the order of its term. We then approximate $V_{\text{myopic}}(n\mid p_{t-1})$ by truncating this decomposition at some order $w$, keeping all terms of order at most $w$ and discarding the rest.

Fix a batch with count matrix $n$ and ordered evaluations $(i_1,j_1),\ldots,(i_U,j_U)$. For any subset $G\subseteq\{1,\ldots,U\}$, let $n_G$ denote the sub-batch consisting of the evaluations whose positions are in $G$, and for any subset $H\subseteq\{1,\ldots,U\}$, define its \emph{interaction term}
\begin{equation*}
\Gamma_H
:=
\sum_{G\subseteq H}
(-1)^{|H|-|G|}
V_{\text{myopic}}(n_G\mid p_{t-1}).
\end{equation*}
To illustrate what the interaction terms represent, we write them explicitly for three examples. The order-zero interaction term is the value with no further evaluation, $\Gamma_\varnothing = V_{\text{myopic}}(0\mid p_{t-1}) =0.$ For a single evaluation $u \in \{1,\ldots,U\}$, $\Gamma_{\{u\}} = V_{\text{myopic}}(n_{\{u\}}\mid p_{t-1})-V_{\text{myopic}}(0\mid p_{t-1})$ is its standalone value of information. For a pair of distinct evaluations $\{u,u'\}$,
\begin{equation*}
\Gamma_{\{u,u'\}}
=
V_{\text{myopic}}(n_{\{u,u'\}}\mid p_{t-1})
-
V_{\text{myopic}}(n_{\{u\}}\mid p_{t-1})
-
V_{\text{myopic}}(n_{\{u'\}}\mid p_{t-1})
+
V_{\text{myopic}}(0\mid p_{t-1})
\end{equation*}
is the net value of observing the two evaluations together beyond the sum of their individual effects. Applying the Boolean M\"obius inversion to the set function $G \mapsto V_{\text{myopic}}(n_G\mid p_{t-1})$ gives \cref{lem:moebius}, which says we can write $V_{\text{myopic}}(n\mid p_{t-1})$ as the sum of the $\Gamma_H$ terms.

\begin{lemma}[Boolean M\"obius inversion]\label{lem:moebius}
For a batch with count matrix $n$ and positions $\{1,\ldots,U\}$,
\begin{equation*}
V_{\text{myopic}}(n\mid p_{t-1})
=
\sum_{H\subseteq \{1,\ldots,U\}}
\Gamma_H.
\end{equation*}
\end{lemma}
We next define the truncated approximation. For any integer $w\ge 1$, define the order-$w$ truncation $\widehat V_{\text{myopic},w}(n\mid p_{t-1}) := \sum_{H\subseteq \{1,\ldots,U\},\,|H|\le w} \Gamma_H$.
\begin{approximation}\label{approx:trunc}
We approximate $V_{\text{myopic}}(n\mid p_{t-1}) \approx \widehat V_{\text{myopic},w}(n\mid p_{t-1})$ with error
\begin{equation*}
V_{\text{myopic}}(n\mid p_{t-1})-\widehat V_{\text{myopic},w}(n\mid p_{t-1})
=
\sum_{\substack{H\subseteq \{1,\ldots,U\}, |H|\ge w+1}}
\Gamma_H.
\end{equation*}
\end{approximation}
If there are no interaction terms of higher order than $w$, i.e., if the batch size is at most $w$, then the approximation is exact. This truncation makes the value function tractable, while still allowing us to capture some of the interaction effects between evaluations in a batch. Every retained interaction term involves computing $V_{\text{myopic}}$ on sub-batches of size at most $w$, which requires sums over at most $2^w$ outcome vectors with multivariate normal CDFs of dimension at most $w$, regardless of the batch size $U$.

\subsubsection{Batch construction}\label{sec:kg:batchcon}

Using \cref{approx:trunc} for some fixed $w$, we build batches in a similar fashion to \cref{alg:batch-surrogate}. Consider a partially committed batch $(i_1,j_1),\ldots,(i_U,j_U)$ with count matrix $n$. For a (feasible) candidate evaluation $(i,j)\in\mathcal I$, appended to the batch as $(i_{U+1},j_{U+1})$, define its marginal gain by
\begin{equation}\label{eq:delta-U-w}
\Delta_n^{(w)}((i,j))
:=
\widehat V_{\text{myopic},w}(n+F^{ij}\mid p_{t-1})-\widehat V_{\text{myopic},w}(n\mid p_{t-1})
\; =
\sum_{\substack{H\subseteq \{1,\ldots,U\}, |H|\le w-1}}
\Gamma_{H\cup\{U+1\}}.
\end{equation}
In words, the marginal gain of $(i,j)$ under \cref{approx:trunc} is the sum, over all subsets of at most $w-1$ already committed evaluations, of the interaction term of that subset together with the new candidate evaluation.

Starting from the empty batch, we repeatedly add the feasible candidate with the largest per-unit-resource marginal gain. As discussed in \cref{subsec:kgdiff}, all marginal gains may be zero, in which case we fall back to the evaluation that \cref{alg:batch-surrogate} would choose. We use the elastic round budgets of \cref{sec:batching:surrogate}. The procedure terminates when the round budget is spent or no evaluation is affordable, and the output is the batch's count matrix $n$. The endgame batch-building procedure, which we call \textsc{AKGw}, is summarized in \cref{alg:akgw}.

\begin{algorithm}[!htb]
\caption{Approximate Knowledge Gradient with order-$w$ truncation (\textsc{AKGw})}
\label{alg:akgw}
\makeatletter
\def\theHALG@line{akgw.\arabic{ALG@line}}
\makeatother
\begin{algorithmic}[1]
\Require $\boldsymbol{\mu}_{t-1}$, $\Sigma_{t-1}$, round budget $B_t$,
remaining total budget $R_{\mathrm{rem}}$ (initially $R$),
$\text{excess}$ from the previous round (initially zero), and $w$
\State $B_{\mathrm{rem}}\leftarrow B_t-\text{excess}$
\State $n\leftarrow 0$, \quad $U\leftarrow 0$, \quad $\Sigma'_t(0)\leftarrow\Sigma_{t-1}$
\State Compute $(q_t,\boldsymbol b_t)$ via \cref{cor:affine-gauss}
\While{$B_{\mathrm{rem}}>0$ \textbf{and}
       $R_{\mathrm{rem}}\geq\min(r_0,r_1)$}
    \State $(i,j) \leftarrow \arg\max\limits_{(i,j)\in\mathcal I:\, r_\jmath \leq R_{\mathrm{rem}}}
    \dfrac{\Delta_n^{(w)}((i,j))}{r_\jmath}$
    \Comment{\eqref{eq:delta-U-w}}
    \If{$\Delta_n^{(w)}((i,j))=0$}
        \State $(i,j)\leftarrow
        \arg\max_{(i,j) \in \mathcal I : r_\jmath \leq R_{\mathrm{rem}}} \psi_{t,ij}(n)$
        \Comment{\eqref{eq:batch-score}}
    \EndIf
    \State Update the covariance from $\Sigma'_t(n)$ to
    $\Sigma'_t(n+F^{ij})$ by \eqref{eq:batch-cov-update}
    \State $(i_{U+1},j_{U+1}) \leftarrow (i,j)$,\quad $U \leftarrow U+1$,\quad $n \leftarrow n + F^{ij}$
    \State $B_{\mathrm{rem}}\leftarrow B_{\mathrm{rem}}-r_\jmath$,
    \quad $R_{\mathrm{rem}}\leftarrow R_{\mathrm{rem}}-r_\jmath$
\EndWhile
\State $\text{excess}\leftarrow\max\{0,-B_{\mathrm{rem}}\}$
\State \Return $n(t)=n$, $R_{\mathrm{rem}}$, and $\text{excess}$
\end{algorithmic}
\end{algorithm}

The choice of $w$ depends on how large the problem is, and on the time and computational resources available. With a current batch of size $U \geq w -1$, the computation of \eqref{eq:delta-U-w} for all candidate evaluations is dominated by computing $|\mathcal I|\binom{U}{w-1}2^w$ $w$-dimensional multivariate normal CDFs. The method is therefore tractable either when the batch size $U$ is small (for example, with a small $B_t$) or when $w$ is sufficiently small (in our simulations, we use $w=3$).

\subsubsection{Proof of \texorpdfstring{\cref{prop:mu_plus}}{Proposition~\ref{prop:mu_plus}}}
\label{app:mu_plus}

\begin{proof}{Proof.}
Fix a round $t$, a candidate batch $(i_1,j_1),\ldots,(i_U,j_U),$ with count matrix $n$, and an outcome vector $\boldsymbol{x}=(x_1,\ldots,x_U)^\top\in\{-1,1\}^U.$ Recall that $\jmath_u:=\min(1,j_u), \boldsymbol{e}_0:=\boldsymbol{0}.$ For each $u=1,\ldots,U$, introduce the latent Gaussian variable
\begin{equation*}
    Z_u
    =
    \beta_{\jmath_u}
    \FitDelims{[}{]}{
        (\boldsymbol{e}_{i_u}-\boldsymbol{e}_{j_u})^\top
        \boldsymbol{\theta}
        -
        \theta_0\1\{j_u=0\}
    }
    +
    \varepsilon_u,
\end{equation*}
where $\boldsymbol{\varepsilon} = (\varepsilon_1,\ldots,\varepsilon_U)^\top \sim N(\boldsymbol{0},I_U)$ is independent of $\boldsymbol{\theta}$. Then
\begin{equation*}
    X_u=1
    \quad\Longleftrightarrow\quad
    Z_u\ge 0,
    \qquad
    X_u=-1
    \quad\Longleftrightarrow\quad
    Z_u<0.
\end{equation*}
Under the prior $p_{t-1}(\cdot)=N(\boldsymbol{\mu}_{t-1},\Sigma_{t-1}),$ the vector $\boldsymbol{Z}=(Z_1,\ldots,Z_U)^\top$ is multivariate normal. Its mean vector is $\boldsymbol{m}=(m_1,\ldots,m_U)^\top$, with
\begin{equation*}
    m_u
    =
    \E[Z_u]
    =
    \beta_{\jmath_u}(\mu_{t-1,i_u}-\mu_{t-1,j_u}).
\end{equation*}
Its covariance matrix is $C\in\mathbb R^{U\times U}$ with
\begin{equation*}
    C_{uu'}
    =
    \Cov(Z_u,Z_{u'})
    =
    \beta_{\jmath_u}\beta_{\jmath_{u'}}
    (\boldsymbol{e}_{i_u}-\boldsymbol{e}_{j_u})^\top
    \Sigma_{t-1}
    (\boldsymbol{e}_{i_{u'}}-\boldsymbol{e}_{j_{u'}})
    +
    \1\{u=u'\}.
\end{equation*}
Define
\begin{equation*}
    D
    =
    \diag\FitParens{\sqrt{C_{11}},\ldots,\sqrt{C_{UU}}},
    \qquad
    D_{\boldsymbol{x}}
    =
    \diag(x_1,\ldots,x_U),
    \qquad
    \widetilde C_{\boldsymbol{x}}
    =
    D_{\boldsymbol{x}}D^{-1}CD^{-1}D_{\boldsymbol{x}}.
\end{equation*}
The matrix $D^{-1}CD^{-1}$ is the correlation matrix of $\boldsymbol{Z}$, and $\widetilde C_{\boldsymbol{x}}$ is the corresponding sign-flipped correlation matrix. This is exactly the construction of \citet[Appendix~B]{Mullahy2017}. In Mullahy's notation, for a binary outcome he defines signs $s_{jp}=2k_{jp}-1$, forms the diagonal sign matrix $T_p=\diag(s_{jp})$, and replaces the original correlation matrix $R$ by $Q_p=T_pRT_p.$ Here, $D_{\boldsymbol{x}}$ plays the role of $T_p$, and $\widetilde C_{\boldsymbol{x}}$ plays the role of $Q_p$. Now define
\begin{equation*}
\begin{gathered}
    \boldsymbol{\alpha}(\boldsymbol{x})
    =
    \FitParens{
        \alpha_1(\boldsymbol{x}),\ldots,\alpha_U(\boldsymbol{x})
    }^\top\\
\text{where}
    \qquad
    \alpha_u(\boldsymbol{x})
    =
    \frac{x_um_u}{\sqrt{C_{uu}}}
    =
    \frac{
        x_u\beta_{\jmath_u}(\mu_{t-1,i_u}-\mu_{t-1,j_u})
    }{
        \sqrt{C_{uu}}
    },
    \quad
    u=1,\ldots,U.
\end{gathered}
\end{equation*}
Let $\boldsymbol{W} = -D_{\boldsymbol{x}}D^{-1}(\boldsymbol{Z}-\boldsymbol{m})$ so $\boldsymbol{W}\sim N(\boldsymbol{0},\widetilde C_{\boldsymbol{x}}).$ For each coordinate $u$,
\begin{equation*}
\begin{aligned}
    W_u\le \alpha_u(\boldsymbol{x})
    \quad\Longleftrightarrow\quad
    -\frac{x_u(Z_u-m_u)}{\sqrt{C_{uu}}}
    \le
    \frac{x_um_u}{\sqrt{C_{uu}}}
    \Longleftrightarrow\quad
    x_uZ_u\ge 0.
\end{aligned}
\end{equation*}
Therefore $\{\boldsymbol{X}=\boldsymbol{x}\} = \{\boldsymbol{W}\le \boldsymbol{\alpha}(\boldsymbol{x})\}$ almost surely. Hence
\begin{equation*}
    \Pp(\boldsymbol{X}=\boldsymbol{x}\mid n,p_{t-1})
    =
    \Pp\FitParens{
        \boldsymbol{W}\le \boldsymbol{\alpha}(\boldsymbol{x})
    }
    =
    \Phi_U\FitParens{
        \boldsymbol{\alpha}(\boldsymbol{x});
        \widetilde C_{\boldsymbol{x}}
    }.
\end{equation*}
This proves the outcome-probability formula.

We next derive the formula for $\boldsymbol{\lambda}_{\boldsymbol{x}}$. Viewing the outcome probability above as a function of the latent mean vector $\boldsymbol{m}$, write $p_{\boldsymbol{x}}(\boldsymbol{m}) := \Phi_U\!\FitParens{ \boldsymbol{\alpha}(\boldsymbol{x}); \widetilde C_{\boldsymbol{x}} }.$ Since $C$, and hence $D$ and $\widetilde C_{\boldsymbol{x}}$, does not depend on $\boldsymbol{m}$, the dependence on $m_u$ is only through $\alpha_u(\boldsymbol{x})=x_um_u/\sqrt{C_{uu}}$. Thus,
\begin{equation*}
\frac{\partial p_{\boldsymbol{x}}(\boldsymbol{m})}{\partial m_u}
=
\frac{\partial}{\partial \alpha_u}
\Phi_U\!\FitParens{
\boldsymbol{\alpha}(\boldsymbol{x});
\widetilde C_{\boldsymbol{x}}
}
\frac{x_u}{\sqrt{C_{uu}}}.
\end{equation*}
We next compute this derivative. For $\boldsymbol{W}\sim N(\boldsymbol{0},\widetilde C_{\boldsymbol{x}})$,
\begin{equation*}
\frac{\partial}{\partial \alpha_u}
\Phi_U(\boldsymbol{\alpha};\widetilde C_{\boldsymbol{x}})
=
\phi(\alpha_u)
\Pp\!\FitParens{
\boldsymbol{W}_{-u}\le\boldsymbol{\alpha}_{-u}
\,\FitMid{|}\,
W_u=\alpha_u
}
=
\phi(\alpha_u)
\Phi_{U-1}\!\FitParens{
\boldsymbol{\alpha}_{-u}
-
(\widetilde C_{\boldsymbol{x}})_{-u,u}\alpha_u;\,
(\widetilde C_{\boldsymbol{x}})_{-u\mid u}
},
\end{equation*}
where the second equality follows from the conditional multivariate normal formula \citetext{\citealp[Eq.~(8a.2.11)]{Rao1973}, \citealp[Appendix~B]{Mullahy2017}}. Substituting this into the derivative of $p_{\boldsymbol{x}}(\boldsymbol{m})$ gives
\begin{equation*}
\frac{\partial p_{\boldsymbol{x}}(\boldsymbol{m})}{\partial m_u}
=
\frac{x_u}{\sqrt{C_{uu}}}
\phi\!\FitParens{\alpha_u(\boldsymbol{x})}
\Phi_{U-1}\!\FitParens{
\boldsymbol{\alpha}_{-u}(\boldsymbol{x})
-
(\widetilde C_{\boldsymbol{x}})_{-u,u}
\alpha_u(\boldsymbol{x});\,
(\widetilde C_{\boldsymbol{x}})_{-u\mid u}
}.
\end{equation*}
Therefore,
\begin{align*}
\frac{\partial}{\partial m_u}
\log p_{\boldsymbol{x}}(\boldsymbol{m})
&=
\frac{1}{p_{\boldsymbol{x}}(\boldsymbol{m})}
\frac{\partial p_{\boldsymbol{x}}(\boldsymbol{m})}{\partial m_u} \\
&=
\frac{x_u}{\sqrt{C_{uu}}}
\frac{
\phi\!\FitParens{\alpha_u(\boldsymbol{x})}
\Phi_{U-1}\!\FitParens{
\boldsymbol{\alpha}_{-u}(\boldsymbol{x})
-
(\widetilde C_{\boldsymbol{x}})_{-u,u}
\alpha_u(\boldsymbol{x});\,
(\widetilde C_{\boldsymbol{x}})_{-u\mid u}
}
}{
\Phi_U\!\FitParens{
\boldsymbol{\alpha}(\boldsymbol{x});
\widetilde C_{\boldsymbol{x}}
}
}
=
(\boldsymbol{\lambda}_{\boldsymbol{x}})_u.
\end{align*}
Hence $\boldsymbol{\lambda}_{\boldsymbol{x}} = \nabla_{\boldsymbol{m}} \log p_{\boldsymbol{x}}(\boldsymbol{m}).$

It remains to connect $\boldsymbol{\lambda}_{\boldsymbol{x}}$ to the posterior mean of $\boldsymbol{\theta}$. The pair $(\boldsymbol{\theta},\boldsymbol{Z})$ is jointly Gaussian. For each $u=1,\ldots,U$,
\begin{equation*}
    \Cov(\boldsymbol{\theta},Z_u)
    =
    \beta_{\jmath_u}
    \Sigma_{t-1}
    (\boldsymbol{e}_{i_u}-\boldsymbol{e}_{j_u}).
\end{equation*}
By the conditional multivariate normal formula \citep[Eq.~(8a.2.11)]{Rao1973},
\begin{equation*}
    \E[\boldsymbol{\theta}\mid \boldsymbol{Z}]
    =
    \boldsymbol{\mu}_{t-1}
    +
    \Cov(\boldsymbol{\theta},\boldsymbol{Z})
    C^{-1}
    (\boldsymbol{Z}-\boldsymbol{m}).
\end{equation*}
Since $\boldsymbol{X}$ is determined by $\boldsymbol{Z}$, the tower property gives
\begin{equation*}
    \E[
        \boldsymbol{\theta}
        \mid
        \boldsymbol{X}=\boldsymbol{x},n,p_{t-1}
    ]
    =
    \boldsymbol{\mu}_{t-1}
    +
    \Cov(\boldsymbol{\theta},\boldsymbol{Z})
    C^{-1}
    \FitParens{
        \E[
            \boldsymbol{Z}
            \mid
            \boldsymbol{X}=\boldsymbol{x},n,p_{t-1}
        ]
        -
        \boldsymbol{m}
    }.
\end{equation*}
We will show $C^{-1}\FitParens{\E[\boldsymbol{Z}\mid\boldsymbol{X}=\boldsymbol{x},n,p_{t-1}]-\boldsymbol{m}}=\boldsymbol{\lambda}_{\boldsymbol{x}}$. Define $A_{\boldsymbol{x}} := \{\boldsymbol{z}\in\mathbb R^U:x_uz_u\ge 0,\ u=1,\ldots,U\}.$ Since $\{\boldsymbol{X}=\boldsymbol{x}\}=\{\boldsymbol{Z}\in A_{\boldsymbol{x}}\}$ almost surely, we have
\begin{equation*}
    p_{\boldsymbol{x}}(\boldsymbol{m})
    =
    \int_{A_{\boldsymbol{x}}}
    \phi_C(\boldsymbol{z}-\boldsymbol{m})\,d\boldsymbol{z},
\end{equation*}
where $\phi_C$ is the $N(\boldsymbol{0},C)$ density. Differentiating gives
\begin{equation*}
    \nabla_{\boldsymbol{m}}
    \phi_C(\boldsymbol{z}-\boldsymbol{m})
    =
    C^{-1}(\boldsymbol{z}-\boldsymbol{m})
    \phi_C(\boldsymbol{z}-\boldsymbol{m})
\end{equation*}
which is locally dominated by some integrable function of $\boldsymbol{z}$. Therefore, by the dominated convergence theorem,
\begin{equation*}
\begin{aligned}
    \nabla_{\boldsymbol{m}}p_{\boldsymbol{x}}(\boldsymbol{m})
    &=
    \int_{A_{\boldsymbol{x}}}
    C^{-1}(\boldsymbol{z}-\boldsymbol{m})
    \phi_C(\boldsymbol{z}-\boldsymbol{m})\,d\boldsymbol{z} \\
    &=
    C^{-1}
    \FitDelims{[}{]}{
        \int_{A_{\boldsymbol{x}}}
        \boldsymbol{z}\,
        \phi_C(\boldsymbol{z}-\boldsymbol{m})
        \,d\boldsymbol{z}
        -
        \boldsymbol{m}
        \int_{A_{\boldsymbol{x}}}
        \phi_C(\boldsymbol{z}-\boldsymbol{m})
        \,d\boldsymbol{z}
    } \\
    &=
    C^{-1}
    \FitDelims{[}{]}{
        p_{\boldsymbol{x}}(\boldsymbol{m})
        \E[
            \boldsymbol{Z}
            \mid
            \boldsymbol{X}=\boldsymbol{x},n,p_{t-1}
        ]
        -
        p_{\boldsymbol{x}}(\boldsymbol{m})\boldsymbol{m}
    } \\
    &=
    p_{\boldsymbol{x}}(\boldsymbol{m})
    C^{-1}
    \FitParens{
        \E[
            \boldsymbol{Z}
            \mid
            \boldsymbol{X}=\boldsymbol{x},n,p_{t-1}
        ]
        -
        \boldsymbol{m}
    }.
\end{aligned}
\end{equation*}
Therefore,
\begin{equation*}
C^{-1}
\FitParens{
\E[
\boldsymbol{Z}
\mid
\boldsymbol{X}=\boldsymbol{x},n,p_{t-1}
]
-
\boldsymbol{m}
}
=
\nabla_{\boldsymbol{m}}
\log p_{\boldsymbol{x}}(\boldsymbol{m})
=
\boldsymbol{\lambda}_{\boldsymbol{x}}.
\end{equation*}
Substituting this into the expression above gives $\boldsymbol{\mu}_t(\boldsymbol{x}\mid n,p_{t-1}) = \boldsymbol{\mu}_{t-1} + \Cov(\boldsymbol{\theta},\boldsymbol{Z}) \boldsymbol{\lambda}_{\boldsymbol{x}}$, and
\begin{equation*}
\Cov(\boldsymbol{\theta},\boldsymbol{Z})
\boldsymbol{\lambda}_{\boldsymbol{x}}
=
\sum_{u=1}^U
\beta_{\jmath_u}
(\boldsymbol{\lambda}_{\boldsymbol{x}})_u
\Sigma_{t-1}
(\boldsymbol{e}_{i_u}-\boldsymbol{e}_{j_u}).
\end{equation*}
Hence
\begin{equation*}
\boldsymbol{\mu}_t(\boldsymbol{x}\mid n,p_{t-1})
=
\boldsymbol{\mu}_{t-1}
+
\sum_{u=1}^U
\beta_{\jmath_u}
(\boldsymbol{\lambda}_{\boldsymbol{x}})_u
\Sigma_{t-1}
(\boldsymbol{e}_{i_u}-\boldsymbol{e}_{j_u}).
\end{equation*}
This also proves \cref{lem:kg-update} which is the special case $U=1$.
\end{proof}

\section{Supplements to \texorpdfstring{\cref{sec:model}}{Section 2}}\label{app:model}
\subsection{Derivations of calibration formulas}
\label{app:calform}
\paragraph{Calibrating the pairwise accuracy factor $\beta_1$.}
We calibrate $\beta_1$ from the probability that a comparison of two randomly chosen items agrees with their true ordering. Let
\begin{equation*}
P_P:=\Pp(\text{a comparison of two randomly chosen items agrees with the true ordering}).
\end{equation*}
Our goal is to express $P_P$ as a function of $\beta_1$. Under the prior, the utility difference $\Delta:=\theta_i-\theta_j$ satisfies $\Delta\sim N(0,2s)$. Writing $E\sim N(0,1)$ independently and $Z:=\beta_1\Delta+E$, the comparison outcome is determined by the sign of $Z$, while the true ordering is determined by the sign of $\Delta$. Hence
\begin{equation*}
P_P=\Pp\!\FitParens{\operatorname{sign}(Z)=\operatorname{sign}(\Delta)}.
\end{equation*}
We first compute the correlation of $(\Delta,Z)$.
\begin{equation*}
\Var(\Delta)=2s,\qquad
\Var(Z)=\beta_1^2\Var(\Delta)+\Var(E)=1+2s\beta_1^2, \qquad
\Cov(\Delta,Z) =\Cov(\Delta,\beta_1\Delta+E)=2s\beta_1,
\end{equation*}
thus their correlation is
\begin{equation*}
\rho_P
=
\frac{\Cov(\Delta,Z)}{\sqrt{\Var(\Delta)\Var(Z)}}
=
\frac{\beta_1\sqrt{2s}}{\sqrt{1+2s\beta_1^2}}.
\end{equation*}
For a mean-zero bivariate normal pair with correlation $\rho_P$, the probability that the two variables have the same sign is $\frac12+\frac{1}{\pi}\arcsin(\rho_P)$. Therefore,
\begin{equation*}
P_P
=
\frac12+\frac{1}{\pi}
\arcsin\!\FitParens{
\frac{\beta_1\sqrt{2s}}{\sqrt{1+2s\beta_1^2}}
},
\end{equation*}
which gives the calibration equation for $\beta_1$.

\paragraph{Calibrating the rating accuracy factor $\beta_0$ and the threshold $\theta_0$.}
We calibrate $\beta_0$ and $\theta_0$ jointly from the probability that a rating is correct and the expected thumbs-up rate under the prior. Let
\begin{equation*}
P_R:=\Pp(\text{a rating of a randomly chosen item is correct}),
\qquad
\overline P:=\Pp(X_{i0}=1).
\end{equation*}

\medskip\noindent \emph{Step a: Calibrate the thumbs-up rate.} Writing $E\sim N(0,1)$ independently, a thumbs-up occurs when $\beta_0(\theta_i-\theta_0)+E>0$. Since $\theta_i\sim N(0,s)$, we have $\beta_0\theta_i+E\sim N(0,1+s\beta_0^2)$, and therefore
\begin{equation*}
\overline P
=
\Pp\!\FitParens{\beta_0\theta_i+E>\beta_0\theta_0}
=
\Phi\!\FitParens{
-\frac{\beta_0\theta_0}{\sqrt{1+s\beta_0^2}}
}.
\end{equation*}

\medskip\noindent \emph{Step b: Calibrate the probability that a rating is correct.} Let $\Delta_i:=\theta_i-\theta_0$ and $Z:=\beta_0\Delta_i+E$. The true rating of item $i$ is determined by the sign of $\Delta_i$, while the estimated rating is determined by the sign of $Z$. Hence
\begin{equation*}
P_R
=
\Pp(\Delta_i>0,Z>0)
+
\Pp(\Delta_i<0,Z<0).
\end{equation*}
To compute these probabilities, note that $(\Delta_i,Z)$ is bivariate normal with
\begin{equation*}
\Delta_i\sim N(-\theta_0,s), \qquad
Z\sim N(-\beta_0\theta_0,1+s\beta_0^2), \qquad
\Cov(\Delta_i,Z)=\beta_0s,
\end{equation*}
so their correlation is
\begin{equation*}
\rho_0
=
\frac{\beta_0s}{\sqrt{s(1+s\beta_0^2)}}
=
\frac{\beta_0\sqrt{s}}{\sqrt{1+s\beta_0^2}}.
\end{equation*}
Standardizing with
\begin{equation*}
U:=\frac{\Delta_i+\theta_0}{\sqrt{s}},
\qquad
V:=\frac{Z+\beta_0\theta_0}{\sqrt{1+s\beta_0^2}},
\end{equation*}
gives a mean-zero bivariate normal pair $(U,V)$ with correlation $\rho_0$, and
\begin{equation*}
\Delta_i>0
\iff
U>\frac{\theta_0}{\sqrt{s}},
\qquad
Z>0
\iff
V>\frac{\beta_0\theta_0}{\sqrt{1+s\beta_0^2}}.
\end{equation*}
Writing $\Phi_2(\cdot,\cdot;\rho)$ for the standard bivariate normal CDF with correlation $\rho$, we therefore obtain
\begin{align*}
P_R
&=
\Pp\!\FitParens{
U>\frac{\theta_0}{\sqrt{s}},
V>\frac{\beta_0\theta_0}{\sqrt{1+s\beta_0^2}}
}
+
\Pp\!\FitParens{
U<\frac{\theta_0}{\sqrt{s}},
V<\frac{\beta_0\theta_0}{\sqrt{1+s\beta_0^2}}
}\\
&=
1-\Phi\!\FitParens{\frac{\theta_0}{\sqrt{s}}}
-\Phi\!\FitParens{\frac{\beta_0\theta_0}{\sqrt{1+s\beta_0^2}}}
+2\Phi_2\!\FitParens{
\frac{\theta_0}{\sqrt{s}},
\frac{\beta_0\theta_0}{\sqrt{1+s\beta_0^2}};
\rho_0
}.
\end{align*}
Together, the equations for $\overline P$ and $P_R$ give the calibration equations for $(\beta_0,\theta_0)$.

\section{Proofs of results in \texorpdfstring{\cref{sec:surrogate}}{Section 3}}\label{app:proofs}

\begin{proof}{Proof of \cref{lem:mismatch}.}
First consider a fixed utility vector $\boldsymbol{\vartheta}\in\R^I$, and let $S_k^*(\boldsymbol{\vartheta})$ denote the set of indices of the $k$ largest elements of $\boldsymbol{\vartheta}$. Define the opportunity cost $O(\boldsymbol{\vartheta};S_k):=\sum_{i\in S_k^*(\boldsymbol{\vartheta})}\vartheta_i-\sum_{i\in S_k}\vartheta_i$, and recall the mismatch loss $\mathcal{M}(\boldsymbol{\vartheta};S_k,a)$ from \eqref{eq:mismatchdef} evaluated at $\boldsymbol{\vartheta}$. Since $|S_k|=|S_k^*(\boldsymbol{\vartheta})|=k$, the sets $S_k\setminus S_k^*(\boldsymbol{\vartheta})$ and $S_k^*(\boldsymbol{\vartheta})\setminus S_k$ have the same cardinality. Then
\begin{align*}
    \mathcal{M}(\boldsymbol{\vartheta};S_k,a)
    &=
    \sum_{i\in S_k}\big(a(\boldsymbol{\vartheta})-\vartheta_i\big)^+
    +\sum_{i\notin S_k}\big(\vartheta_i-a(\boldsymbol{\vartheta})\big)^+ \\
    &\ge
    \sum_{i\in S_k\setminus S_k^*(\boldsymbol{\vartheta})}\big(a(\boldsymbol{\vartheta})-\vartheta_i\big)^+
    +\sum_{i\in S_k^*(\boldsymbol{\vartheta})\setminus S_k}\big(\vartheta_i-a(\boldsymbol{\vartheta})\big)^+ \\
    &\ge
    \sum_{i\in S_k\setminus S_k^*(\boldsymbol{\vartheta})}\big(a(\boldsymbol{\vartheta})-\vartheta_i\big)
    +\sum_{i\in S_k^*(\boldsymbol{\vartheta})\setminus S_k}\big(\vartheta_i-a(\boldsymbol{\vartheta})\big) \\
    &=
    \sum_{i\in S_k^*(\boldsymbol{\vartheta})\setminus S_k}\vartheta_i
    -\sum_{i\in S_k\setminus S_k^*(\boldsymbol{\vartheta})}\vartheta_i
    \;=\;
    O(\boldsymbol{\vartheta};S_k).
\end{align*}
The first inequality drops the nonnegative terms indexed by $i\in S_k\cap S_k^*(\boldsymbol{\vartheta})$ and $i\notin S_k\cup S_k^*(\boldsymbol{\vartheta})$, the second uses $x^+\ge x$, and the third line's $a$ terms cancel because the two index sets have equal cardinality. This is the upper bound. When $a(\boldsymbol{\vartheta})\in[\vartheta^{(k+1)},\vartheta^{(k)}]$ we have $\vartheta_i\ge a(\boldsymbol{\vartheta})$ for $i\in S_k^*(\boldsymbol{\vartheta})$ and $\vartheta_i\le a(\boldsymbol{\vartheta})$ for $i\notin S_k^*(\boldsymbol{\vartheta})$, so the dropped terms in the first inequality are zero, and each $x^+ = x$ in the second inequality. Hence both inequalities are then equalities, giving $O(\boldsymbol{\vartheta};S_k)=\mathcal{M}(\boldsymbol{\vartheta};S_k,a)$.

Applying this pointwise with $\boldsymbol{\vartheta}=\boldsymbol{\theta}$ and taking expectations over $\boldsymbol{\theta}\mid p_{t-1}\sim N(\boldsymbol{\mu}_{t-1},\Sigma_{t-1})$ gives
\begin{equation*}
\mathrm{EOC}(S_k;p_{t-1})=\E\!\FitDelims{[}{]}{O(\boldsymbol{\theta};S_k)\,\FitMid{|}\,p_{t-1}}\le\E\!\FitDelims{[}{]}{\mathcal{M}(\boldsymbol{\theta};S_k,a)\,\FitMid{|}\,p_{t-1}},
\end{equation*}
with equality whenever $a(\boldsymbol{\theta})\in[\theta^{(k+1)},\theta^{(k)}]$ almost surely.
\end{proof}

\begin{proof}{Proof of \cref{prop:affine}.}
$\mathcal{M}(\boldsymbol{\theta};S_k,a_{q,\boldsymbol{b}})=k\,a_{q,\boldsymbol{b}}(\boldsymbol{\theta})-\sum_{i\in S_k}\theta_i+\sum_{i=1}^I(\theta_i-a_{q,\boldsymbol{b}}(\boldsymbol{\theta}))^+$, so with $\E[a_{q,\boldsymbol{b}}(\boldsymbol{\theta})\mid p_{t-1}]=q$,
\begin{equation*}
\E[\mathcal{M}(\boldsymbol{\theta};S_k,a_{q,\boldsymbol{b}})\mid p_{t-1}]
=
kq-\sum_{i\in S_k}\mu_{t-1,i}
+\sum_{i=1}^I\E\!\FitDelims{[}{]}{(\theta_i-a_{q,\boldsymbol{b}}(\boldsymbol{\theta}))^+\,\FitMid{|}\,p_{t-1}}.
\end{equation*}
For each fixed $\boldsymbol{\theta}$, $(\theta_i-a_{q,\boldsymbol{b}}(\boldsymbol{\theta}))^+$ is the positive part of an affine function of $(q,\boldsymbol{b})$, hence convex. Summing, taking expectations, and adding the linear term $kq$ preserves convexity, so the expected mismatch loss is convex in $(q,\boldsymbol{b})$.

We next show that the minimum is attained. Using $(x-y)^+\ge x^+-|y|$ in the definition of mismatch loss,
\begin{equation*}
\mathcal{M}(\boldsymbol{\theta};S_k,a_{q,\boldsymbol{b}})
\ge
k\,a_{q,\boldsymbol{b}}(\boldsymbol{\theta})^+
+(I-k)\bigl(-a_{q,\boldsymbol{b}}(\boldsymbol{\theta})\bigr)^+
-\sum_{i=1}^I|\theta_i|
\ge
\min\{k,I-k\}\FitDelims{|}{|}{a_{q,\boldsymbol{b}}(\boldsymbol{\theta})}
-\sum_{i=1}^I|\theta_i|.
\end{equation*}
Moreover, since $a_{q,\boldsymbol{b}}(\boldsymbol{\theta})=q+\boldsymbol{b}^\top(\boldsymbol{\theta}-\boldsymbol{\mu}_{t-1})$,
\begin{equation*}
\E\!\FitDelims{[}{]}{\FitDelims{|}{|}{a_{q,\boldsymbol{b}}(\boldsymbol{\theta})}\,\FitMid{|}\,p_{t-1}}\ge |q|,
\qquad
2\E\!\FitDelims{[}{]}{\FitDelims{|}{|}{a_{q,\boldsymbol{b}}(\boldsymbol{\theta})}\,\FitMid{|}\,p_{t-1}}
\ge
\sqrt{\frac{2}{\pi}}\sqrt{\boldsymbol{b}^\top\Sigma_{t-1}\boldsymbol{b}}.
\end{equation*}
Since $\Sigma_{t-1}$ is positive definite, the expected mismatch loss therefore tends to infinity as $\|(q,\boldsymbol{b})\|_2\to\infty$. Since it is finite and convex, hence continuous, it attains a minimum.

It remains to characterize any minimizer not of the form $(\mu_{t-1,j},\boldsymbol{e}_j)$. For such a pair, for every $i$, either $\boldsymbol{b}\neq\boldsymbol{e}_i$, in which case $\theta_i-a_{q,\boldsymbol{b}}(\boldsymbol{\theta})$ is a nondegenerate Gaussian, or $\boldsymbol{b}=\boldsymbol{e}_i$ and $q\neq\mu_{t-1,i}$, in which case $\theta_i-a_{q,\boldsymbol{b}}(\boldsymbol{\theta})=\mu_{t-1,i}-q\neq0$. Thus,
\begin{equation*}
\Pp\!\FitParens{\theta_i=a_{q,\boldsymbol{b}}(\boldsymbol{\theta})\,\FitMid{|}\,p_{t-1}}=0
\qquad\text{for every $i$,}
\end{equation*}
so the expected mismatch loss is differentiable at $(q,\boldsymbol{b})$. Differentiating under the expectation gives
\begin{equation*}
\frac{\partial}{\partial q}(\theta_i-a_{q,\boldsymbol{b}}(\boldsymbol{\theta}))^+
=
-\1\{\theta_i>a_{q,\boldsymbol{b}}(\boldsymbol{\theta})\},
\qquad
\frac{\partial}{\partial\boldsymbol{b}}(\theta_i-a_{q,\boldsymbol{b}}(\boldsymbol{\theta}))^+
=
-\1\{\theta_i>a_{q,\boldsymbol{b}}(\boldsymbol{\theta})\}
(\boldsymbol{\theta}-\boldsymbol{\mu}_{t-1}).
\end{equation*}
Summing over $i$, the indicators add to $\kappa_{q,\boldsymbol{b}}(\boldsymbol{\theta})+k$, giving
\begin{align*}
\frac{\partial}{\partial q}
\E[\mathcal{M}(\boldsymbol{\theta};S_k,a_{q,\boldsymbol{b}})\mid p_{t-1}]
&=
k-\E\!\FitDelims{[}{]}{\kappa_{q,\boldsymbol{b}}(\boldsymbol{\theta})+k\,\FitMid{|}\,p_{t-1}}
=
-\E\!\FitDelims{[}{]}{\kappa_{q,\boldsymbol{b}}(\boldsymbol{\theta})\,\FitMid{|}\,p_{t-1}},\\
\frac{\partial}{\partial\boldsymbol{b}}
\E[\mathcal{M}(\boldsymbol{\theta};S_k,a_{q,\boldsymbol{b}})\mid p_{t-1}]
&=
-\E\!\FitDelims{[}{]}{(\boldsymbol{\theta}-\boldsymbol{\mu}_{t-1})
\{\kappa_{q,\boldsymbol{b}}(\boldsymbol{\theta})+k\}\,\FitMid{|}\,p_{t-1}}=
-\E\!\FitDelims{[}{]}{(\boldsymbol{\theta}-\boldsymbol{\mu}_{t-1})
\kappa_{q,\boldsymbol{b}}(\boldsymbol{\theta})\,\FitMid{|}\,p_{t-1}},
\end{align*}
where the last equality uses $\E[\boldsymbol{\theta}-\boldsymbol{\mu}_{t-1}\mid p_{t-1}]=0$. The $\ell$th component of the second derivative is $-\Cov(\theta_\ell,\kappa_{q,\boldsymbol{b}}(\boldsymbol{\theta})\mid p_{t-1})$. Thus, for any pair not of the form $(\mu_{t-1,j},\boldsymbol{e}_j)$, setting both derivatives to zero gives the conditions in \cref{prop:affine}(ii), which are necessary and sufficient for optimality by convexity.

We next characterize when an exceptional point in \cref{prop:affine}(i) is a minimizer. Fix $j$ and consider $(q,\boldsymbol{b})=(\mu_{t-1,j},\boldsymbol{e}_j)$. For any direction $(c,\boldsymbol{d})$, the directional derivative of the $j$th positive-part term is
\begin{align*}
&\lim_{\epsilon\downarrow0}
\frac{
\E\!\FitDelims{[}{]}{
\bigl(\theta_j-a_{\mu_{t-1,j}+\epsilon c,\boldsymbol{e}_j+\epsilon\boldsymbol{d}}(\boldsymbol{\theta})\bigr)^+
\,\FitMid{|}\,p_{t-1}
}
-
\E\!\FitDelims{[}{]}{
\bigl(\theta_j-a_{\mu_{t-1,j},\boldsymbol{e}_j}(\boldsymbol{\theta})\bigr)^+
\,\FitMid{|}\,p_{t-1}
}
}{\epsilon}\\&=
\lim_{\epsilon\downarrow0}
\frac{1}{\epsilon}
\E\!\FitDelims{[}{]}{
\FitParens{
\theta_j-\mu_{t-1,j}-\epsilon c
-(\boldsymbol{e}_j+\epsilon\boldsymbol{d})^\top
(\boldsymbol{\theta}-\boldsymbol{\mu}_{t-1})
}^+
\,\FitMid{|}\,p_{t-1}
}\\
&=
\lim_{\epsilon\downarrow0}
\frac{1}{\epsilon}
\E\!\FitDelims{[}{]}{
\FitParens{
-\epsilon c-\epsilon\boldsymbol{d}^\top
(\boldsymbol{\theta}-\boldsymbol{\mu}_{t-1})
}^+
\,\FitMid{|}\,p_{t-1}
}\\
&=
\E\!\FitDelims{[}{]}{
\FitParens{
-c-\boldsymbol{d}^\top
(\boldsymbol{\theta}-\boldsymbol{\mu}_{t-1})
}^+
\,\FitMid{|}\,p_{t-1}
}.
\end{align*}
Hence the directional derivative of the expected mismatch loss at $(\mu_{t-1,j},\boldsymbol{e}_j)$ in direction $(c,\boldsymbol{d})$ is
\begin{equation*}
c\FitDelims{[}{]}{
k-\sum_{i\neq j}
\Phi\!\FitParens{
\frac{\mu_{t-1,i}-\mu_{t-1,j}}
{\sigma_{t-1,ij}}
}
}
-
\boldsymbol{d}^\top\Sigma_{t-1}
\sum_{i\neq j}
\frac{
\phi\!\FitParens{
\frac{\mu_{t-1,i}-\mu_{t-1,j}}
{\sigma_{t-1,ij}}
}
}{
\sigma_{t-1,ij}
}
(\boldsymbol{e}_i-\boldsymbol{e}_j)
+
\E\!\FitDelims{[}{]}{
\bigl(-c-\boldsymbol{d}^\top(\boldsymbol{\theta}-\boldsymbol{\mu}_{t-1})\bigr)^+
\,\FitMid{|}\,p_{t-1}
}.
\end{equation*}
By convexity, $(\mu_{t-1,j},\boldsymbol{e}_j)$ is a minimizer if and only if this directional derivative is nonnegative for every $(c,\boldsymbol{d})$. At $\boldsymbol{d}=\boldsymbol{0}$ this requires and is equivalent to
\begin{equation}\label{eq:affine-exceptional-i}
0
\le
k-\sum_{i\neq j}
\Phi\!\FitParens{
\frac{\mu_{t-1,i}-\mu_{t-1,j}}
{\sigma_{t-1,ij}}
}
\le
1.
\end{equation}
Now, for any fixed $\boldsymbol{d}$, taking the infimum of the directional derivative over $c$ gives
\begin{equation*}
\phi\!\FitParens{
\Phi^{-1}\!\FitDelims{[}{]}{
k-\sum_{i\neq j}
\Phi\!\FitParens{
\frac{\mu_{t-1,i}-\mu_{t-1,j}}
{\sigma_{t-1,ij}}
}
}
}
\sqrt{\boldsymbol{d}^\top\Sigma_{t-1}\boldsymbol{d}}
-
\boldsymbol{d}^\top\Sigma_{t-1}
\sum_{i\neq j}
\frac{
\phi\!\FitParens{
\frac{\mu_{t-1,i}-\mu_{t-1,j}}
{\sigma_{t-1,ij}}
}
}{
\sigma_{t-1,ij}
}
(\boldsymbol{e}_i-\boldsymbol{e}_j).
\end{equation*}
Requiring this to be nonnegative for every $\boldsymbol{d}$ is equivalent to
\begin{equation}\label{eq:affine-exceptional-ii}
\FitParens{
\sum_{i\neq j}
\frac{
\phi\!\FitParens{
\frac{\mu_{t-1,i}-\mu_{t-1,j}}
{\sigma_{t-1,ij}}
}
}{
\sigma_{t-1,ij}
}
(\boldsymbol{e}_i-\boldsymbol{e}_j)
}^\top
\Sigma_{t-1}
\FitParens{
\sum_{i\neq j}
\frac{
\phi\!\FitParens{
\frac{\mu_{t-1,i}-\mu_{t-1,j}}
{\sigma_{t-1,ij}}
}
}{
\sigma_{t-1,ij}
}
(\boldsymbol{e}_i-\boldsymbol{e}_j)
}
\le
\phi\!\FitParens{
\Phi^{-1}\!\FitDelims{[}{]}{
k-\sum_{i\neq j}
\Phi\!\FitParens{
\frac{\mu_{t-1,i}-\mu_{t-1,j}}
{\sigma_{t-1,ij}}
}
}
}^2.
\end{equation}
Therefore, $(\mu_{t-1,j},\boldsymbol{e}_j)$ is a minimizer if and only if \eqref{eq:affine-exceptional-i} and \eqref{eq:affine-exceptional-ii} hold.\end{proof}

\begin{proof}{Proof of \cref{cor:affine-gauss}.}
For part (i), by \cref{prop:affine}(ii), it is enough to verify the two optimality conditions. We first consider the condition $\E[\kappa_{q_t,\boldsymbol{b}_t}(\boldsymbol{\theta})\mid p_{t-1}]=0$.
\begin{equation*}
\E\!\FitDelims{[}{]}{\kappa_{q_t,\boldsymbol{b}_t}(\boldsymbol{\theta})\,\FitMid{|}\,p_{t-1}}
=
\E\!\FitDelims{[}{]}{
\sum_{i=1}^I
\1\!\FitDelims{\{}{\}}{\theta_i>a_{q_t,\boldsymbol{b}_t}(\boldsymbol{\theta})}
-k
\,\FitMid{|}\,p_{t-1}
}=
\sum_{i=1}^I
\Pp\!\FitParens{
\theta_i>a_{q_t,\boldsymbol{b}_t}(\boldsymbol{\theta})
\,\FitMid{|}\,p_{t-1}
}-k.
\end{equation*}
We next compute these probabilities
\begin{equation*}
\Pp\!\FitParens{
\theta_i>a_{q_t,\boldsymbol{b}_t}(\boldsymbol{\theta})
\,\FitMid{|}\,p_{t-1}
}
=
\Pp\!\FitParens{
(\boldsymbol{e}_i-\boldsymbol{b}_t)^\top
(\boldsymbol{\theta}-\boldsymbol{\mu}_{t-1})
>
q_t-\mu_{t-1,i}
\,\FitMid{|}\,p_{t-1}
}=
1-\Phi\!\FitParens{
\frac{q_t-\mu_{t-1,i}}{\sigma_{t-1,i}}
}
=
\Phi\!\FitParens{z_{t-1,i}}.
\end{equation*}
Substituting this back gives
\begin{equation*}
\E\!\FitDelims{[}{]}{\kappa_{q_t,\boldsymbol{b}_t}(\boldsymbol{\theta})\,\FitMid{|}\,p_{t-1}}
=
\sum_{i=1}^I\Phi(z_{t-1,i})-k.
\end{equation*}
We next verify the covariance condition in \cref{prop:affine}(ii). Expanding $\kappa_{q_t,\boldsymbol{b}_t}$ and using $\E[\theta_\ell-\mu_{t-1,\ell}\mid p_{t-1}]=0$,
\begin{equation*}
\Cov\!\FitParens{
\theta_\ell,\kappa_{q_t,\boldsymbol{b}_t}(\boldsymbol{\theta})
\,\FitMid{|}\,p_{t-1}
}
=
\sum_{i=1}^I
\E\!\FitDelims{[}{]}{
(\theta_\ell-\mu_{t-1,\ell})
\1\!\FitDelims{\{}{\}}{
\theta_i>a_{q_t,\boldsymbol{b}_t}(\boldsymbol{\theta})
}
\,\FitMid{|}\,p_{t-1}
}.
\end{equation*}
For each $i$, the variables $\theta_\ell-\mu_{t-1,\ell}$ and $(\boldsymbol{e}_i-\boldsymbol{b}_t)^\top(\boldsymbol{\theta}-\boldsymbol{\mu}_{t-1})$ are jointly Gaussian. Thus,
\begin{align*}
&\E\!\FitDelims{[}{]}{
(\theta_\ell-\mu_{t-1,\ell})
\1\!\FitDelims{\{}{\}}{
\theta_i>a_{q_t,\boldsymbol{b}_t}(\boldsymbol{\theta})
}
\,\FitMid{|}\,p_{t-1}
}\\
&\qquad=
\frac{
\boldsymbol{e}_\ell^\top\Sigma_{t-1}(\boldsymbol{e}_i-\boldsymbol{b}_t)
}{
\sigma_{t-1,i}^2
}
\E\!\FitDelims{[}{]}{
(\boldsymbol{e}_i-\boldsymbol{b}_t)^\top
(\boldsymbol{\theta}-\boldsymbol{\mu}_{t-1})
\1\!\FitDelims{\{}{\}}{
(\boldsymbol{e}_i-\boldsymbol{b}_t)^\top
(\boldsymbol{\theta}-\boldsymbol{\mu}_{t-1})
>
q_t-\mu_{t-1,i}
}
\,\FitMid{|}\,p_{t-1}
}\\
&\qquad=
\frac{\phi(z_{t-1,i})}{\sigma_{t-1,i}}
\boldsymbol{e}_\ell^\top
\Sigma_{t-1}(\boldsymbol{e}_i-\boldsymbol{b}_t)
=
w_{t-1,i}\,
\boldsymbol{e}_\ell^\top
\Sigma_{t-1}(\boldsymbol{e}_i-\boldsymbol{b}_t).
\end{align*}
Therefore,
\begin{equation*}
\Cov\!\FitParens{
\theta_\ell,\kappa_{q_t,\boldsymbol{b}_t}(\boldsymbol{\theta})
\,\FitMid{|}\,p_{t-1}
}
=
\sum_{i=1}^I
w_{t-1,i}\,
\boldsymbol{e}_\ell^\top
\Sigma_{t-1}(\boldsymbol{e}_i-\boldsymbol{b}_t)=
\boldsymbol{e}_\ell^\top\Sigma_{t-1}
\FitParens{
\sum_{i=1}^I w_{t-1,i}\boldsymbol{e}_i
-
\FitParens{\sum_{i=1}^I w_{t-1,i}}\boldsymbol{b}_t
}.
\end{equation*}
Thus, taking
\begin{equation*}
\boldsymbol{b}_t
=
\frac{\sum_{i=1}^I w_{t-1,i}\boldsymbol{e}_i}
{\sum_{i=1}^I w_{t-1,i}},
\qquad\text{equivalently}\qquad
b_{t,i}
=
\frac{w_{t-1,i}}{\sum_{\ell=1}^I w_{t-1,\ell}},
\end{equation*}
makes the covariance zero for every $\ell=1,\ldots,I$. Hence both optimality conditions of \cref{prop:affine} are satisfied, proving part (i).

For part (ii), set $\boldsymbol{b}=\boldsymbol{0}$, so only the constant threshold $q$ is optimized. By convexity, the optimum is characterized by the first-order condition
\begin{equation*}
0=
\frac{\partial}{\partial q}
\E\!\FitDelims{[}{]}{
\mathcal{M}(\boldsymbol{\theta};S_k,a_{q,\boldsymbol{0}})
\,\FitMid{|}\,p_{t-1}
}=
k-\sum_{i=1}^I
\Pp(\theta_i>q\mid p_{t-1})=
k-\sum_{i=1}^I
\Phi\!\FitParens{
\frac{\mu_{t-1,i}-q}{\sigma_{t-1,i0}}
}.
\end{equation*}
Therefore the optimal constant threshold $q_t^{\mathrm{con}}$ satisfies
\begin{equation*}
\sum_{i=1}^I
\Phi\!\FitParens{
\frac{\mu_{t-1,i}-q_t^{\mathrm{con}}}{\sigma_{t-1,i0}}
}
=
k. \Halmoshere
\end{equation*}
\end{proof}

\begin{proof}{Proof of \cref{lem:expected-cov}.}
Fix an evaluation $(i,j)\in\mathcal I$. By the definition of expected posterior covariance and the law of total covariance,
\begin{equation*}
\overline{\Sigma}_t((i,j),p_{t-1})
=\Cov(\boldsymbol{\theta}\mid p_{t-1})
-\Cov\!\FitParens{
\E[\boldsymbol{\theta}\mid X_{ij},p_{t-1}]
\,\FitMid{|}\,p_{t-1}}
=\Sigma_{t-1}
-\Cov\!\FitParens{
\E[\boldsymbol{\theta}\mid X_{ij},p_{t-1}]
\,\FitMid{|}\,p_{t-1}}.
\end{equation*}
Thus, we need the covariance of the posterior mean across the two possible outcomes. We compute this from the probability of each outcome and its resulting posterior mean.

First consider the outcome probabilities. Let $Z\sim N(0,1)$ be independent of $\boldsymbol{\theta}$. Then
\begin{align*}
\Pp(X_{ij}=x\mid p_{t-1})
&=\E\!\FitDelims{[}{]}{
\Phi\!\FitParens{x\beta_\jmath(\theta_i-\theta_j)}
\,\FitMid{|}\,p_{t-1}}\\
&=\Pp\!\FitParens{
Z\le x\beta_\jmath(\theta_i-\theta_j)
\,\FitMid{|}\,p_{t-1}}\\
&=\Pp\!\FitParens{
\frac{Z-x\beta_\jmath(\theta_i-\theta_j-\mu_{t-1,i}+\mu_{t-1,j})}
{\sqrt{1+\beta_\jmath^2\sigma_{t-1,ij}^2}}
\le
\frac{x\beta_\jmath(\mu_{t-1,i}-\mu_{t-1,j})}
{\sqrt{1+\beta_\jmath^2\sigma_{t-1,ij}^2}}
\,\FitMid{|}\,p_{t-1}}\\
&=\Phi\!\FitParens{
\frac{x\beta_\jmath(\mu_{t-1,i}-\mu_{t-1,j})}
{\sqrt{1+\beta_\jmath^2\sigma_{t-1,ij}^2}}
}
=\Phi(x\alpha_{t,ij}).
\end{align*}

We next compute the posterior mean. By Bayes' rule,
\begin{equation*}
\E[\boldsymbol{\theta}\mid X_{ij}=x,p_{t-1}]
-\boldsymbol{\mu}_{t-1}
=\frac{
\E\!\FitDelims{[}{]}{
(\boldsymbol{\theta}-\boldsymbol{\mu}_{t-1})
\Pp(X_{ij}=x\mid\boldsymbol{\theta})
\,\FitMid{|}\,p_{t-1}}}
{\Pp(X_{ij}=x\mid p_{t-1})}=\frac{
\E\!\FitDelims{[}{]}{
(\boldsymbol{\theta}-\boldsymbol{\mu}_{t-1})
\Phi\!\FitParens{x\beta_\jmath(\theta_i-\theta_j)}
\,\FitMid{|}\,p_{t-1}}}
{\Phi(x\alpha_{t,ij})}.
\end{equation*}
It remains to evaluate the numerator.
\begin{align*}
&\E\!\FitDelims{[}{]}{
(\boldsymbol{\theta}-\boldsymbol{\mu}_{t-1})
\Phi\!\FitParens{x\beta_\jmath(\theta_i-\theta_j)}
\,\FitMid{|}\,p_{t-1}}\\
&\quad=
\E\!\FitDelims{[}{]}{
(\boldsymbol{\theta}-\boldsymbol{\mu}_{t-1})
\1\{Z\le x\beta_\jmath(\theta_i-\theta_j)\}
\,\FitMid{|}\,p_{t-1}}\\
&\quad=
\frac{x\beta_\jmath\Sigma_{t-1}(\boldsymbol e_i-\boldsymbol e_j)}
{1+\beta_\jmath^2\sigma_{t-1,ij}^2}
\E\!\FitDelims{[}{]}{
\bigl(x\beta_\jmath(\theta_i-\theta_j-\mu_{t-1,i}+\mu_{t-1,j})-Z\bigr)
\1\{Z\le x\beta_\jmath(\theta_i-\theta_j)\}
\,\FitMid{|}\,p_{t-1}}\\
&\quad=
\frac{x\beta_\jmath\Sigma_{t-1}(\boldsymbol e_i-\boldsymbol e_j)}
{\sqrt{1+\beta_\jmath^2\sigma_{t-1,ij}^2}}
\int_{-x\alpha_{t,ij}}^\infty u\phi(u)\,du\\
&\quad=
\frac{x\beta_\jmath\phi(\alpha_{t,ij})}
{\sqrt{1+\beta_\jmath^2\sigma_{t-1,ij}^2}}
\Sigma_{t-1}(\boldsymbol e_i-\boldsymbol e_j).
\end{align*}
The second equality uses the covariance $\Cov(\boldsymbol{\theta},x\beta_\jmath(\theta_i-\theta_j)-Z\mid p_{t-1}) =x\beta_\jmath\Sigma_{t-1}(\boldsymbol e_i-\boldsymbol e_j)$ and variance $1+\beta_\jmath^2\sigma_{t-1,ij}^2$. The third equality standardizes the Gaussian, and the last uses $\phi'(u)=-u\phi(u)$ and symmetry of $\phi$.

Substituting into the posterior mean gives
\begin{equation*}
\E[\boldsymbol{\theta}\mid X_{ij}=x,p_{t-1}]
-\boldsymbol{\mu}_{t-1}
=
\frac{x\beta_\jmath\phi(\alpha_{t,ij})}
{\Phi(x\alpha_{t,ij})\sqrt{1+\beta_\jmath^2\sigma_{t-1,ij}^2}}
\Sigma_{t-1}(\boldsymbol e_i-\boldsymbol e_j).
\end{equation*}
We can now compute the covariance term. By the law of total expectation, $\E[\E[\boldsymbol{\theta}\mid X_{ij},p_{t-1}]\mid p_{t-1}]=\boldsymbol{\mu}_{t-1}$, so
\begin{align*}
&\Cov\!\FitParens{
\E[\boldsymbol{\theta}\mid X_{ij},p_{t-1}]
\,\FitMid{|}\,p_{t-1}}\\
&\quad=
\sum_{x\in\{-1,1\}}
\Pp(X_{ij}=x\mid p_{t-1})
\FitParens{
\E[\boldsymbol{\theta}\mid X_{ij}=x,p_{t-1}]
-\boldsymbol{\mu}_{t-1}
}
\FitParens{
\E[\boldsymbol{\theta}\mid X_{ij}=x,p_{t-1}]
-\boldsymbol{\mu}_{t-1}
}^\top\\
&\quad=
\frac{\beta_\jmath^2\phi(\alpha_{t,ij})^2}
{1+\beta_\jmath^2\sigma_{t-1,ij}^2}
\FitParens{\sum_{x\in\{-1,1\}}
\Phi(x\alpha_{t,ij})
\FitDelims{[}{]}{\frac{x}{\Phi(x\alpha_{t,ij})}}^2}
\Sigma_{t-1}(\boldsymbol e_i-\boldsymbol e_j)
(\boldsymbol e_i-\boldsymbol e_j)^\top\Sigma_{t-1}\\
&\quad=
\frac{\beta_\jmath^2\phi(\alpha_{t,ij})^2}
{1+\beta_\jmath^2\sigma_{t-1,ij}^2}
\FitParens{
\frac{1}{\Phi(\alpha_{t,ij})}
+\frac{1}{\Phi(-\alpha_{t,ij})}
}
\Sigma_{t-1}(\boldsymbol e_i-\boldsymbol e_j)
(\boldsymbol e_i-\boldsymbol e_j)^\top\Sigma_{t-1}\\
&\quad=
\frac{\beta_\jmath^2\phi(\alpha_{t,ij})^2}
{(1+\beta_\jmath^2\sigma_{t-1,ij}^2)
\Phi(\alpha_{t,ij})\Phi(-\alpha_{t,ij})}
\Sigma_{t-1}(\boldsymbol e_i-\boldsymbol e_j)
(\boldsymbol e_i-\boldsymbol e_j)^\top\Sigma_{t-1}\\
&\quad=
\frac{\delta_{t,ij}}{\sigma_{t-1,ij}^2}
\Sigma_{t-1}(\boldsymbol e_i-\boldsymbol e_j)
(\boldsymbol e_i-\boldsymbol e_j)^\top\Sigma_{t-1}.
\end{align*}
Here we used $x^2=1$, $\Phi(\alpha_{t,ij})+\Phi(-\alpha_{t,ij})=1$, and the definition of $\delta_{t,ij}$. Substituting back into the expected posterior covariance yields
\begin{equation*}
\overline{\Sigma}_t((i,j),p_{t-1})
=\Sigma_{t-1}
-\frac{\delta_{t,ij}}{\sigma_{t-1,ij}^2}
\Sigma_{t-1}(\boldsymbol e_i-\boldsymbol e_j)
(\boldsymbol e_i-\boldsymbol e_j)^\top\Sigma_{t-1}. \Halmoshere
\end{equation*}
\end{proof}

\section{Proof of \cref{thm:consistency}}\label{app:consistency}
All probabilities and expectations below are with respect to the prior, the evaluation model, and the policy. Until the final convergence argument, consider \textsc{SA} run continuously without a budget constraint.

\subsection{Notation}\label{app:consistency:notation}
Let
\begin{equation*}
P:=\operatorname{Id}_I-\frac{1}{I}\boldsymbol{1}\boldsymbol{1}^\top,
\qquad
\boldsymbol{1}^\perp:=\{\boldsymbol{x}\in\R^I:\boldsymbol{1}^\top\boldsymbol{x}=0\}.
\end{equation*}

Let $\mathcal D$ denote the finite set of evaluation directions available to the policy. A rating of item $i$ has direction $\boldsymbol{e}_i$, and a pairwise comparison of items $i$ and $j$ has direction $\boldsymbol{e}_i-\boldsymbol{e}_j$. Thus,
\begin{equation*}
\mathcal D
=
\{\boldsymbol{e}_i-\boldsymbol{e}_j:(i,j)\in\mathcal I\},
\end{equation*}
where $\boldsymbol{e}_0=\boldsymbol{0}$, and $\boldsymbol{1}^\perp\subseteq\operatorname{span}(\mathcal D).$ For $L\in\mathcal D$, define
\begin{equation*}
\beta_L
:=
\begin{cases}
\beta_0, & L\text{ is a rating direction},\\
\beta_1, & L\text{ is a comparison direction},
\end{cases}
\qquad
\theta_0(L)
:=
\begin{cases}
\theta_0, & L\text{ is a rating direction},\\
0, & L\text{ is a comparison direction},
\end{cases}
\end{equation*}
and define $r_L$ analogously as $r_0$ or $r_1$. Conditional on $\boldsymbol{\theta}$ and the history before an evaluation, its success probability depends only on the chosen direction. For every $L\in\mathcal D$,
\begin{equation*}
\Pp(X_L=+1\mid\boldsymbol{\theta})
=
\E_{\epsilon_\jmath}\!\FitDelims{[}{]}{
g\!\FitParens{L^\top\boldsymbol{\theta}-\theta_0(L)+\epsilon_\jmath}
},
\end{equation*}
where $\jmath=0$ for a rating direction and $\jmath=1$ for a comparison direction. This probability is continuous and strictly increasing in $L^\top\boldsymbol{\theta}$.

Let $\mathcal H_t$ be the history available at the start of round $t$, i.e., the $\sigma$-algebra generated by all evaluation choices and outcomes through round $t-1$, and let
\begin{equation*}
\boldsymbol{\mu}_{t-1}^*:=\E[\boldsymbol{\theta}\mid\mathcal H_t],
\qquad
\Sigma_{t-1}^*:=\Cov(\boldsymbol{\theta}\mid\mathcal H_t)
\end{equation*}
be the exact posterior moments at the start of round $t$. \textsc{SA} uses the approximate moments $(\boldsymbol{\mu}_{t-1},\Sigma_{t-1})$.

For notational convenience, write
\begin{equation*}
\boldsymbol{g}_{t,i}:=\boldsymbol{e}_i-\boldsymbol{b}_t,
\qquad
\sigma_{t-1,i}^2:=\boldsymbol{g}_{t,i}^\top\Sigma_{t-1}\boldsymbol{g}_{t,i},
\qquad
z_{t-1,i}:=\frac{\mu_{t-1,i}-q_t}{\sigma_{t-1,i}},
\qquad
w_{t-1,i}:=\frac{\phi(z_{t-1,i})}{\sigma_{t-1,i}}.
\end{equation*}
Here $z_{t-1,i}$ and $w_{t-1,i}$ are defined only when $\sigma_{t-1,i}>0$. By \cref{prop:affine} and the proof of \cref{cor:affine-gauss}, either
\begin{enumerate}[label=(\roman*)]
\item $(q_t,\boldsymbol{b}_t) =(\mu_{t-1,j},\boldsymbol{e}_j)$ for some $j$. Then $\sigma_{t-1,j}=0$, while $\sigma_{t-1,i}>0$ for $i\ne j$. Since $a_{q_t,\boldsymbol{b}_t}(\boldsymbol{\theta})=\theta_j$, the $j$th loss term is zero before and after every evaluation, so its contribution to the marginal score is zero.

\item $\sigma_{t-1,i}>0$ for every $i$, and
\begin{equation*}
\sum_{i=1}^I\Phi(z_{t-1,i})=k,
\qquad
b_{t,i}=\frac{w_{t-1,i}}{\sum_{\ell=1}^I w_{t-1,\ell}},
\quad i=1,\ldots,I.
\end{equation*}
\end{enumerate}

Note that $\boldsymbol{1}^\top\boldsymbol{b}_t=1$, which gives the following
\begin{equation*}
\boldsymbol{g}_{t,i}\in\boldsymbol{1}^\perp,
\qquad
\boldsymbol{g}_{t,i}-\boldsymbol{g}_{t,j}=\boldsymbol{e}_i-\boldsymbol{e}_j, \qquad
\operatorname{span}\{\boldsymbol{g}_{t,1},\ldots,\boldsymbol{g}_{t,I}\}
=
\boldsymbol{1}^\perp
\quad\text{for every $t$,}
\end{equation*}
used repeatedly below.

For $L\in\mathcal D$, let $\alpha_t(L)$ and $\delta_t(L)$ denote the quantities $\alpha_{t,ij}$ and $\delta_{t,ij}$ in \cref{lem:expected-cov} for the corresponding evaluation. The marginal score of \textsc{SA} can be written in unified form as

\begin{equation}\label{eq:cons-score}
G_t(L)
:=
\frac{1}{r_L}
\sum_{i:\,\sigma_{t-1,i}>0}
\int_0^{
\frac{
\delta_t(L)
\FitParens{\boldsymbol{g}_{t,i}^\top\Sigma_{t-1}L}^2
}{
\sigma_{t-1,i}^2L^\top\Sigma_{t-1}L
}
}
\frac{\sigma_{t-1,i}}{2\sqrt{1-s}}
\phi\!\FitParens{\frac{z_{t-1,i}}{\sqrt{1-s}}}\,ds.
\end{equation}
This is \eqref{eq:score} for the corresponding evaluation, with any zero-variance summand omitted because its contribution is zero.

Define the realized opportunity cost after $T$ rounds by
\begin{equation*}
\mathrm{OC}_T
:=
\sum_{i\in S_k^*(\boldsymbol{\theta})}\theta_i
-
\sum_{i\in S_k^*(\boldsymbol{\mu}_T)}\theta_i.
\end{equation*}
Thus, the conclusion of \cref{thm:consistency} is $\E[\mathrm{OC}_T]\to0$ as $R\to\infty$.

\subsection{Proof roadmap}\label{app:consistency:roadmap}
There are two main components to the proof. The first, and far more technical, is to argue that the set of directions the algorithm samples infinitely many times contains the decision-relevant subspace $\boldsymbol{1}^\perp$ in its span. The second component then argues that, if this is the case, the full history identifies $P\boldsymbol{\theta}$, and thus the expected opportunity cost decays to $0$ as well.

To prove the first claim, we will argue by contradiction. Let $E_\infty$ be the set of directions sampled infinitely often, and let $M:=\operatorname{span}(E_\infty)$. We assume by contradiction that $\boldsymbol{1}^\perp\nsubseteq M$, and condition on an event $\Omega_{\mathcal E}:=\{E_\infty=\mathcal E\}$, where the set of recurrent directions is fixed. On this event, after some finite random time, every subsequently sampled direction lies in $\mathcal E$. This reduces the proof to showing two things: for the directions $L\in\mathcal E$, their marginal score $G_t(L)\to0$, while the total score over $\mathcal D$ stays bounded away from $0$. Thus, for all sufficiently large $t$, at least one direction outside $\mathcal E$ has a larger score than every direction in $\mathcal E$, contradicting the assumption that every subsequently sampled direction lies in $\mathcal E$.

The proof of \cref{thm:consistency} relies on eight lemmas, organized as follows. \Cref{lem:cons-learning} establishes that posterior variance goes to $0$ and the posterior mean converges along every direction in the span of recurrently sampled directions. \Cref{lem:cons-offspan} proves that posterior variance remains uniformly positive outside that span. \Cref{lem:cons-bl,lem:cons-approx} transfer these conclusions to the approximate posterior covariance and provide the bounds needed for the marginal scores. \Cref{lem:cons-threshold} bounds the affine threshold and the weights $w_{t-1,i}$. \Cref{lem:cons-recurrent-score} shows that the scores of recurrent directions decay, while \cref{lem:cons-positive-score} shows that the total score stays bounded away from $0$ whenever $\boldsymbol{1}^\perp$ is not contained in the recurrent span. Finally, \cref{lem:cons-projection} shows that ``projected'' mean convergence suffices for expected opportunity cost.

\subsection{Supporting lemmas}\label{app:consistency:lemmas}

\begin{lemma}\label{lem:cons-learning}
Fix a set $\mathcal E\subseteq\mathcal D$, and let $M:=\operatorname{span}(\mathcal E)$. Let $\mathcal H_\infty := \sigma\!\FitParens{\bigcup_{t\ge1}\mathcal H_t}$ denote the $\sigma$-algebra generated by the full infinite history. Let $\Omega\in\mathcal H_\infty$. Suppose that every $L\in\mathcal E$ is sampled infinitely often on $\Omega$. Then, for every $\boldsymbol{v}\in M$,
\begin{equation*}
\boldsymbol{v}^\top\boldsymbol{\mu}_{t-1}^*\to\boldsymbol{v}^\top\boldsymbol{\theta},
\qquad
\boldsymbol{v}^\top\Sigma_{t-1}^*\boldsymbol{v}\to0
\qquad\text{a.s. on $\Omega$.}
\end{equation*}
Moreover, $\1_\Omega\boldsymbol{v}^\top\boldsymbol{\theta}$ is $\mathcal H_\infty$-measurable.
\end{lemma}

\begin{proof}{Proof.}
For evaluation $t$, let $L_t:=\boldsymbol e_{i_t}-\boldsymbol e_{j_t}\in\mathcal D$ be its chosen direction and let $X_t\in\{-1,1\}$ be the observed outcome. Let $(\mathcal F_t)_{t\ge0}$ be the filtration defined by $\mathcal F_{t-1}:=\sigma(\boldsymbol{\theta},\mathcal H_t,L_t)$ for $t\ge1$. Then $\mathcal F_{t-1}$ contains all previously observed outcomes and the direction chosen before observing $X_t$. In particular, $L_t$ is $\mathcal F_{t-1}$-measurable.

Fix $L\in\mathcal E$, and define
\begin{equation*}
N_t(L):=\sum_{s=1}^t\1\{L_s=L\},
\qquad
\widehat X_t:=\1\{X_t=+1\}.
\end{equation*}
Set
\begin{equation*}
\xi_t(L)
:=
\1\{L_t=L\}
\FitParens{
\widehat X_t-
\Pp(X_L=+1\mid\boldsymbol{\theta})
}.
\end{equation*}
Since $\1\{L_t=L\}$ is $\mathcal F_{t-1}$-measurable, and since conditional on $\mathcal F_{t-1}$ the success probability on the event $\{L_t=L\}$ is $\Pp(X_L=+1\mid\boldsymbol{\theta})$, we have $\E[\xi_t(L)\mid\mathcal F_{t-1}]=0.$ Thus $(\xi_t(L))_{t\ge1}$ is a martingale difference sequence with respect to $(\mathcal F_t)_{t\ge0}$. Now define
\begin{equation*}
M_t(L):=\sum_{s=1}^t\frac{\xi_s(L)}{\max\{N_s(L),1\}}.
\end{equation*}
Then $(M_t(L))_{t\ge1}$ is a martingale, and
\begin{equation*}
\E\!\FitDelims{[}{]}{
\{M_t(L)-M_{t-1}(L)\}^2
\,\FitMid{|}\,
\mathcal F_{t-1}
}
\le
\frac{\1\{L_t=L\}}{\max\{N_t(L),1\}^2}.
\end{equation*}
Hence
\begin{equation*}
\sup_{t\ge1}\E[M_t(L)^2]
\le
\E\!\FitDelims{[}{]}{
\sum_{s=1}^\infty
\frac{\1\{L_s=L\}}{\max\{N_s(L),1\}^2}
}
\le
\sum_{j=1}^\infty\frac{1}{j^2}
<\infty.
\end{equation*}
By the square-integrable martingale convergence theorem, $M_t(L)$ converges almost surely. On $\Omega$, $N_t(L)\to\infty$, so Kronecker's lemma gives
\begin{equation*}
\frac{1}{\max\{N_t(L),1\}}
\sum_{s=1}^t\xi_s(L)
\to0
\qquad\text{a.s. on $\Omega$.}
\end{equation*}
Therefore, on $\Omega$,
\begin{equation*}
\frac{1}{N_t(L)}
\sum_{s=1}^t
\1\{L_s=L\}\widehat X_s
\to
\Pp(X_L=+1\mid\boldsymbol{\theta})
\qquad\text{a.s.}
\end{equation*}
The left-hand side is $\mathcal H_\infty$-measurable, so on $\Omega$ the limit $\Pp(X_L=+1\mid\boldsymbol{\theta})$ is $\mathcal H_\infty$-measurable. Since the link is strictly increasing, $L^\top\boldsymbol{\theta}$ is also $\mathcal H_\infty$-measurable on $\Omega$. Since this holds for every $L\in\mathcal E$, the same holds for every $\boldsymbol{v}\in M=\operatorname{span}(\mathcal E)$.

Fix $\boldsymbol{v}\in M$. Since $\boldsymbol{v}^\top\boldsymbol{\theta}\in L^2$, Doob's martingale convergence theorem gives
\begin{equation*}
\E[\boldsymbol{v}^\top\boldsymbol{\theta}\mid\mathcal H_t]
\to
\E[\boldsymbol{v}^\top\boldsymbol{\theta}\mid\mathcal H_\infty]
\qquad\text{a.s.},
\end{equation*}
and
\begin{equation*}
\E[(\boldsymbol{v}^\top\boldsymbol{\theta})^2\mid\mathcal H_t]
\to
\E[(\boldsymbol{v}^\top\boldsymbol{\theta})^2\mid\mathcal H_\infty]
\qquad\text{a.s.}
\end{equation*}
Since $\Omega\in\mathcal H_\infty$ and both $\boldsymbol{v}^\top\boldsymbol{\theta}$ and $(\boldsymbol{v}^\top\boldsymbol{\theta})^2$ are $\mathcal H_\infty$-measurable on $\Omega$, we have
\begin{equation*}
\E[\boldsymbol{v}^\top\boldsymbol{\theta}\mid\mathcal H_\infty]
=
\boldsymbol{v}^\top\boldsymbol{\theta},
\qquad
\E[(\boldsymbol{v}^\top\boldsymbol{\theta})^2\mid\mathcal H_\infty]
=
(\boldsymbol{v}^\top\boldsymbol{\theta})^2
\end{equation*}
almost surely on $\Omega$. Therefore,
\begin{equation*}
\boldsymbol{v}^\top\boldsymbol{\mu}_{t-1}^*
=
\E[\boldsymbol{v}^\top\boldsymbol{\theta}\mid\mathcal H_t]
\to
\boldsymbol{v}^\top\boldsymbol{\theta}
\qquad\text{a.s. on $\Omega$,}
\end{equation*}
and
\begin{equation*}
\boldsymbol{v}^\top\Sigma_{t-1}^*\boldsymbol{v}
=
\Var(\boldsymbol{v}^\top\boldsymbol{\theta}\mid\mathcal H_t)
\to0
\qquad\text{a.s. on $\Omega$.} \Halmoshere
\end{equation*}
\end{proof}
We next show that if sampling eventually remains in this recurrent span, uncertainty cannot go to zero in directions outside it.
\begin{lemma}\label{lem:cons-offspan}
Fix a deterministic set $\mathcal E\subseteq\mathcal D$, let $M:=\operatorname{span}(\mathcal E)$, and let $R_M$ and $Q$ be the orthogonal projectors onto $M$ and $M^\perp$. Suppose that on an event $\Omega$:
\begin{enumerate}[label=(\roman*)]
\item there exists an almost-surely finite round $T_0$ such that every sampled direction after round $T_0$ belongs to $\mathcal E$, and
\item $R_M\boldsymbol{\mu}_{t-1}^*\to R_M\boldsymbol{\theta}$ and $\|R_M\Sigma_{t-1}^*R_M\|_{\mathrm{op}}\to0$ almost surely on $\Omega$.
\end{enumerate}
Then, almost surely on $\Omega$, there exists $\eta>0$ such that, for all sufficiently large $t$,
\begin{equation*}
Q\Sigma_{t-1}^*Q\succeq\eta Q.
\end{equation*}
\end{lemma}

\begin{proof}{Proof.}
Work on $\Omega$. We can decompose $\R^I=M\oplus M^\perp$, so every $\boldsymbol{\theta}\in\R^I$ can be written uniquely as
\begin{equation*}
\boldsymbol{\theta}=\boldsymbol{u}+\boldsymbol{v},
\qquad
\boldsymbol{u}\in M,
\quad
\boldsymbol{v}\in M^\perp,
\end{equation*}
in particular $\boldsymbol{u}=R_M\boldsymbol{\theta}$ and $\boldsymbol{v}=Q\boldsymbol{\theta}$. Let $f_0(\boldsymbol{u},\boldsymbol{v})$ denote the posterior density for $\boldsymbol{\theta}$ after round $T_0$ evaluated at $\boldsymbol{\theta}=\boldsymbol{u}+\boldsymbol{v}$. Since $T_0<\infty$, this density is obtained from the Gaussian prior density by multiplying finitely many one-observation likelihood factors and normalizing. Thus $f_0$ is continuous and strictly positive. Since each likelihood factor belongs to $(0,1)$,
\begin{equation*}
f_0(\boldsymbol{u},\boldsymbol{v})
\le
Z_{T_0}^{-1}\pi_0(\boldsymbol{u},\boldsymbol{v}),
\end{equation*}
where $\pi_0$ is the Gaussian prior density and $Z_{T_0}>0$ is the normalizing constant.

Now fix $t>T_0$ and let $f_t$ denote the posterior density given $\mathcal H_t$. Every observation collected after round $T_0$ is taken along some direction $L\in\mathcal E\subseteq M$. The corresponding likelihood factor depends on $\boldsymbol{\theta}$ only through $L^\top\boldsymbol{\theta}$. If $\boldsymbol{\theta}=\boldsymbol{u}+\boldsymbol{v}$ with $\boldsymbol{u}\in M$ and $\boldsymbol{v}\in M^\perp$, then
\begin{equation*}
L^\top\boldsymbol{\theta}
=
L^\top(\boldsymbol{u}+\boldsymbol{v})
=
L^\top\boldsymbol{u},
\end{equation*}
because $L\in M$ and $\boldsymbol{v}\perp M$. It follows that
\begin{equation*}
f_t(\boldsymbol{u},\boldsymbol{v})
=
\frac{f_0(\boldsymbol{u},\boldsymbol{v})\Psi_t(\boldsymbol{u})}{Z_t}
\end{equation*}
for some nonnegative function $\Psi_t$ on $M$ and some normalizing constant $Z_t>0$. Therefore, for fixed $\boldsymbol{u}$, the conditional distribution of the $M^\perp$ component does not change after round $T_0$.

Define
\begin{equation*}
c(\boldsymbol{u})
:=
\int_{M^\perp}
f_0(\boldsymbol{u},\boldsymbol{v})\,d\boldsymbol{v},
\end{equation*}
and let $C(\boldsymbol{u})$ denote the covariance matrix under the conditional density $f_0(\boldsymbol{u},\cdot)/c(\boldsymbol{u})$ on $M^\perp$. Choose a bounded closed neighborhood $O\subseteq M$ of $R_M\boldsymbol{\theta}$. Since $f_0$ is continuous and is bounded above by a constant multiple of the Gaussian prior density, dominated convergence implies that $C(\boldsymbol{u})$ is continuous on $O$.

For every $\boldsymbol{u}\in O$, the conditional distribution of $\boldsymbol{v}$ has a strictly positive density on all of $M^\perp$. Hence, its covariance matrix $C(\boldsymbol{u})$ is positive definite on $M^\perp$. Since $O$ is compact and $C(\boldsymbol{u})$ is continuous, there exists $\eta_0>0$ such that
\begin{equation*}
C(\boldsymbol{u})\succeq\eta_0\operatorname{Id}_{M^\perp}
\qquad(\boldsymbol{u}\in O).
\end{equation*}

It remains to show that, under the round-$t$ posterior, the $\boldsymbol{u}$-coordinate belongs to $O$ with probability that approaches $1$. Choose $\epsilon>0$ such that the closed ball of radius $2\epsilon$ around $R_M\boldsymbol{\theta}$ is contained in $O$. By condition (ii), for all sufficiently large $t$,
\begin{equation*}
\|R_M\boldsymbol{\mu}_{t-1}^*-R_M\boldsymbol{\theta}\|_2<\epsilon.
\end{equation*}
For such $t$, if $\boldsymbol{u}\notin O$, then
\begin{equation*}
\|\boldsymbol{u}-R_M\boldsymbol{\mu}_{t-1}^*\|_2\ge\epsilon.
\end{equation*}
Thus, by Markov's inequality,
\begin{equation*}
\Pp(\boldsymbol{u}\notin O\mid\mathcal H_t)
\le
\frac{\operatorname{tr}(R_M\Sigma_{t-1}^*R_M)}{\epsilon^2}
\to0.
\end{equation*}
Finally, the law of total covariance gives
\begin{equation*}
Q\Sigma_{t-1}^*Q
\succeq
\E\!\FitDelims{[}{]}{C(\boldsymbol{u})\mid\mathcal H_t}
\succeq
\eta_0\Pp(\boldsymbol{u}\in O\mid\mathcal H_t)Q.
\end{equation*}
The probability on the right converges to $1$, so the claim holds eventually with $\eta=\eta_0/2$.
\end{proof}
Next we get a uniform upper bound on the posterior covariance.
\begin{lemma}[Brascamp--Lieb]\label{lem:cons-bl}
For every $t$,
\begin{equation*}
\Sigma_{t-1}^*\preceq\Sigma_0.
\end{equation*}
Consequently, under \cref{ass:update}(i), eventually almost surely, $\Sigma_{t-1}\preceq \chi_+\Sigma_0.$ In particular, there exists $\bar\sigma<\infty$ such that
\begin{equation*}
\sigma_{t-1,i}\le \bar\sigma,
\qquad i=1,\ldots,I,
\end{equation*}
eventually almost surely.
\end{lemma}

\begin{proof}{Proof.}
Fix $t$. The likelihood factors are log-concave by \cref{ass:link}, and remain log-concave after averaging over the Gaussian evaluator noise. The posterior is the Gaussian prior multiplied by the product of these factors. By the Brascamp--Lieb covariance inequality \citep{BrascampLieb1975}, $\Sigma_{t-1}^*\preceq\Sigma_0$. By \cref{ass:update}(i), eventually almost surely $\Sigma_{t-1}\preceq \chi_+\Sigma_{t-1}^*\preceq \chi_+\Sigma_0$.

Since $\boldsymbol{b}_t$ has nonnegative entries summing to $1$,
\begin{equation*}
\|\boldsymbol{g}_{t,i}\|_2
\le
\|\boldsymbol{e}_i\|_2+\|\boldsymbol{b}_t\|_2
\le2.
\end{equation*}
Therefore,
\begin{equation*}
\sigma_{t-1,i}^2
=
\boldsymbol{g}_{t,i}^\top\Sigma_{t-1}\boldsymbol{g}_{t,i}
\le
4\chi_+\|\Sigma_0\|_{\mathrm{op}}. \Halmoshere
\end{equation*}

\end{proof}

We next transfer the exact-posterior variance conclusions above to the approximate covariance used by \textsc{SA}.
\begin{lemma}\label{lem:cons-approx}
Fix a deterministic set $\mathcal E\subseteq\mathcal D$, let $M:=\operatorname{span}(\mathcal E)$, and let $R_M$ and $Q$ be the orthogonal projectors onto $M$ and $M^\perp$. Suppose that on an event $\Omega$,
\begin{equation*}
\|R_M\Sigma_{t-1}^*R_M\|_{\mathrm{op}}\to0
\end{equation*}
and, for some $\eta>0$ and all sufficiently large $t$,
\begin{equation*}
Q\Sigma_{t-1}^*Q\succeq\eta Q.
\end{equation*}
Then, under \cref{ass:update}(i) and \cref{lem:cons-bl}, almost surely on $\Omega$,
\begin{equation*}
\|R_M\Sigma_{t-1}R_M\|_{\mathrm{op}}\to0,
\qquad
\|Q\Sigma_{t-1}R_M\|_{\mathrm{op}}+\|R_M\Sigma_{t-1}Q\|_{\mathrm{op}}\to0,
\end{equation*}
and, for all sufficiently large $t$,
\begin{equation*}
Q\Sigma_{t-1}Q\succeq \chi_-\eta Q.
\end{equation*}
\end{lemma}

\begin{proof}{Proof.}
From \cref{ass:update}(i), $0\preceq R_M\Sigma_{t-1}R_M\preceq \chi_+R_M\Sigma_{t-1}^*R_M,$ so $\|R_M\Sigma_{t-1}R_M\|_{\mathrm{op}}\to0$. For the other terms, let $\boldsymbol{x}\in M^\perp$ and $\boldsymbol{y}\in M$ be unit vectors. Since $\Sigma_{t-1}$ is positive semidefinite,
\begin{equation*}
|\boldsymbol{x}^\top\Sigma_{t-1}\boldsymbol{y}|^2
\le
(\boldsymbol{x}^\top\Sigma_{t-1}\boldsymbol{x})
(\boldsymbol{y}^\top\Sigma_{t-1}\boldsymbol{y})
\le
\|Q\Sigma_{t-1}Q\|_{\mathrm{op}}
\|R_M\Sigma_{t-1}R_M\|_{\mathrm{op}}.
\end{equation*}
By \cref{lem:cons-bl}, the first factor is eventually bounded, while the second converges to $0$. Taking the supremum over unit $\boldsymbol{x}$ and $\boldsymbol{y}$ gives
\begin{equation*}
\|Q\Sigma_{t-1}R_M\|_{\mathrm{op}}
+
\|R_M\Sigma_{t-1}Q\|_{\mathrm{op}}
\to0.
\end{equation*}
Finally, \cref{ass:update}(i) gives
\begin{equation*}
Q\Sigma_{t-1}Q
\succeq
\chi_-Q\Sigma_{t-1}^*Q
\succeq
\chi_-\eta Q. \Halmoshere
\end{equation*}
\end{proof}
To control the marginal scores, we also need bounds on the affine threshold and the associated weights.
\begin{lemma}\label{lem:cons-threshold}
Let $\Omega$ be an event on which
\begin{equation*}
\sup_{t\ge1}\|\boldsymbol{\mu}_{t-1}\|_\infty<\infty,
\qquad
\max_{1\le i\le I}\sigma_{t-1,i}\le \bar\sigma
\end{equation*}
eventually for some finite $\bar\sigma$. Then, almost surely on $\Omega$, there exists $Q_0<\infty$ such that $|q_t|\le Q_0$ eventually. In addition, if $(i_t)_{t\ge1}$ is any sequence in $\{1,\ldots,I\}$ such that $\liminf_{t\to\infty}\sigma_{t-1,i_t}>0$ on $\Omega$, then there exists $w_*>0$ such that $w_{t-1,i_t}\ge w_*$ eventually on $\Omega$.
\end{lemma}

\begin{proof}{Proof.}
First consider rounds in case \textup{(ii)} of \cref{prop:affine}. Define
\begin{equation*}
B:=\sup_{t\ge1}\|\boldsymbol{\mu}_{t-1}\|_\infty<\infty,
\qquad
F_t(q):=\sum_{i=1}^I\Phi\!\FitParens{\frac{\mu_{t-1,i}-q}{\sigma_{t-1,i}}}.
\end{equation*}
Recall that $q_t$ is defined to satisfy $F_t(q_t)=k$. For each $t$, the map $F_t$ is continuous and strictly decreasing, with $F_t(q)\to I$ as $q\to-\infty$ and $F_t(q)\to0$ as $q\to+\infty$. Choose $Q_0>B$ such that
\begin{equation*}
I\Phi\!\FitParens{\frac{B-Q_0}{\bar\sigma}}
<k<
I\Phi\!\FitParens{\frac{Q_0-B}{\bar\sigma}}.
\end{equation*}
For all sufficiently large $t$,
\begin{equation*}
F_t(Q_0)
\le
I\Phi\!\FitParens{\frac{B-Q_0}{\bar\sigma}}
<k,
\end{equation*}
and similarly,
\begin{equation*}
F_t(-Q_0)
\ge
I\Phi\!\FitParens{\frac{Q_0-B}{\bar\sigma}}
>k.
\end{equation*}
Since $F_t$ is strictly decreasing and $F_t(q_t)=k$, it follows that
\begin{equation*}
-Q_0<q_t<Q_0
\end{equation*}
for all sufficiently large $t$. On a round in case \textup{(i)} of \cref{prop:affine}, $q_t=\mu_{t-1,j}$ for some $j$, so $|q_t|\le B<Q_0$. Thus, $|q_t|<Q_0$ eventually in either case.

Now suppose $\liminf_{t\to\infty}\sigma_{t-1,i_t}>0$. Then there exists $\underline\sigma>0$ such that
\begin{equation*}
\underline\sigma\le\sigma_{t-1,i_t}\le \bar\sigma
\end{equation*}
eventually. Together with $|\mu_{t-1,i_t}|\le B$ and $|q_t|\le Q_0$, this gives
\begin{equation*}
|z_{t-1,i_t}|
\le
\frac{B+Q_0}{\underline\sigma}
\end{equation*}
eventually. Therefore,
\begin{equation*}
w_{t-1,i_t}
=
\frac{\phi(z_{t-1,i_t})}{\sigma_{t-1,i_t}}
\ge
\frac{\phi((B+Q_0)/\underline\sigma)}{\bar\sigma}
=:w_*>0
\end{equation*}
eventually.
\end{proof}
We can now show that the score of any direction whose variance vanishes must itself vanish.
\begin{lemma}\label{lem:cons-recurrent-score}
Let $L\in\mathcal D$ be a direction such that
\begin{equation*}
L^\top\Sigma_{t-1}L\to0.
\end{equation*}
Then
\begin{equation*}
G_t(L)\to0.
\end{equation*}
\end{lemma}

\begin{proof}{Proof.}
For each $i$ with $\sigma_{t-1,i}>0$, Cauchy--Schwarz gives,
\begin{equation*}
\FitParens{\boldsymbol{g}_{t,i}^\top\Sigma_{t-1}L}^2
\le
\FitParens{\boldsymbol{g}_{t,i}^\top\Sigma_{t-1}\boldsymbol{g}_{t,i}}
\FitParens{L^\top\Sigma_{t-1}L}
=
\sigma_{t-1,i}^2L^\top\Sigma_{t-1}L.
\end{equation*}
Therefore,
\begin{equation*}
w_{t-1,i}
\FitParens{\boldsymbol{g}_{t,i}^\top\Sigma_{t-1}L}^2
\le
\phi(0)\sigma_{t-1,i}L^\top\Sigma_{t-1}L
\le
\phi(0)\bar\sigma L^\top\Sigma_{t-1}L
\end{equation*}
eventually, by \cref{lem:cons-bl}. Summing over $i$ yields
\begin{equation*}
\sum_{i:\,\sigma_{t-1,i}>0}
w_{t-1,i}
\FitParens{\boldsymbol{g}_{t,i}^\top\Sigma_{t-1}L}^2
\le
I\phi(0)\bar\sigma L^\top\Sigma_{t-1}L.
\end{equation*}
Also, Cauchy--Schwarz gives $\phi(z)^2\le\Phi(z)\Phi(-z)$, so $0\le\delta_t(L)<1$ and $\delta_t(L)/(L^\top\Sigma_{t-1}L)\le\beta_L^2$. Since $\phi(z/\sqrt{1-s})\le\phi(z)$ for $0\le s<1$ and $1-\sqrt{1-s}\le s$, integrating in \eqref{eq:cons-score} and substituting the bounds above gives
\begin{equation*}
0\le G_t(L)
\le
\frac{\delta_t(L)}{r_LL^\top\Sigma_{t-1}L}
\sum_{i:\,\sigma_{t-1,i}>0}
w_{t-1,i}
\FitParens{\boldsymbol{g}_{t,i}^\top\Sigma_{t-1}L}^2
\le
\frac{I\phi(0)\bar\sigma\beta_L^2}{r_L}
L^\top\Sigma_{t-1}L
\to0. \Halmoshere
\end{equation*}
\end{proof}
We next show the complementary result, that if the recurrent span does not contain $\boldsymbol{1}^\perp$, some score must remain bounded away from zero.
\begin{lemma}\label{lem:cons-positive-score}
Fix a deterministic set $\mathcal E\subseteq\mathcal D$, let $M:=\operatorname{span}(\mathcal E)$, and let $R_M$ and $Q$ be the orthogonal projectors onto $M$ and $M^\perp$. Suppose $\boldsymbol{1}^\perp\nsubseteq M$. On any event $\Omega$ on which
\begin{enumerate}[label=(\roman*)]
\item $Q\Sigma_{t-1}Q\succeq cQ$ eventually for some $c>0$,
\item $\|R_M\Sigma_{t-1}Q\|_{\mathrm{op}}\to0$, and
\item $\sup_t\|\boldsymbol{\mu}_{t-1}\|_\infty<\infty$ and $\Sigma_{t-1}\preceq\overline\Sigma$ eventually for some fixed positive semidefinite matrix $\overline\Sigma$,
\end{enumerate}
we have
\begin{equation*}
\liminf_{t\to\infty}\sum_{L\in\mathcal D}G_t(L)>0
\qquad\text{a.s. on $\Omega$.}
\end{equation*}
\end{lemma}

\begin{proof}{Proof.}
Because the vectors $\boldsymbol{e}_a-\boldsymbol{e}_b$, $a\ne b$, span $\boldsymbol{1}^\perp$, the assumption $\boldsymbol{1}^\perp\nsubseteq M$ implies that there exist $a\ne b$ such that $Q(\boldsymbol{e}_a-\boldsymbol{e}_b)\ne\boldsymbol{0}.$ Set $\delta := \frac{1}{2} \FitDelims{\|}{\|}{Q(\boldsymbol{e}_a-\boldsymbol{e}_b)}_2 >0.$ Since $Q(\boldsymbol{e}_a-\boldsymbol{e}_b) = Q(\boldsymbol{g}_{t,a}-\boldsymbol{g}_{t,b}),$ for every $t$, there exists $i_t\in\{a,b\}$ such that $\FitDelims{\|}{\|}{Q\boldsymbol{g}_{t,i_t}}_2\ge\delta.$ Using $\|\boldsymbol{g}_{t,i_t}\|_2\le2$, condition \textup{(i)}, and condition \textup{(ii)},
\begin{align*}
\sigma_{t-1,i_t}^2
&=
\boldsymbol{g}_{t,i_t}^{\top}
\Sigma_{t-1}
\boldsymbol{g}_{t,i_t}\\
&\ge
(Q\boldsymbol{g}_{t,i_t})^\top
\Sigma_{t-1}
(Q\boldsymbol{g}_{t,i_t})
-
2\FitDelims{|}{|}{
(R_M\boldsymbol{g}_{t,i_t})^\top
\Sigma_{t-1}
(Q\boldsymbol{g}_{t,i_t})
}\\
&\ge
c\FitDelims{\|}{\|}{Q\boldsymbol{g}_{t,i_t}}_2^2
-
2
\|R_M\Sigma_{t-1}Q\|_{\mathrm{op}}
\|R_M\boldsymbol{g}_{t,i_t}\|_2
\|Q\boldsymbol{g}_{t,i_t}\|_2\\
&\ge
c\delta^2
-
8\|R_M\Sigma_{t-1}Q\|_{\mathrm{op}}.
\end{align*}
Therefore,
\begin{equation*}
\liminf_{t\to\infty}
\sigma_{t-1,i_t}^2
\ge
c\delta^2
>0.
\end{equation*}
By \cref{lem:cons-threshold}, there exists $w_*>0$ such that eventually $w_{t-1,i_t}\ge w_*.$ The same bounds also imply that $z_{t-1,i_t}$ is bounded eventually.

Since $\mathcal D$ is finite and its span contains $\boldsymbol{1}^\perp$, there exists $\kappa>0$ such that, for every $\boldsymbol{y}\in\R^I$,
\begin{equation*}
\sum_{L\in\mathcal D}
(L^\top\boldsymbol{y})^2
\ge
\kappa\|P\boldsymbol{y}\|_2^2.
\end{equation*}
Applying this with $\boldsymbol{y}=\Sigma_{t-1}\boldsymbol{g}_{t,i_t}$, and using $\boldsymbol{g}_{t,i_t}\in\boldsymbol{1}^\perp$ and $\|\boldsymbol{g}_{t,i_t}\|_2\le2$, gives
\begin{equation*}
\sum_{L\in\mathcal D}
\FitParens{
\boldsymbol{g}_{t,i_t}^{\top}
\Sigma_{t-1}L
}^2
\ge
\kappa
\FitDelims{\|}{\|}{
P\Sigma_{t-1}\boldsymbol{g}_{t,i_t}
}_2^2\ge
\frac{\kappa}{4}
\FitParens{
\boldsymbol{g}_{t,i_t}^{\top}
P\Sigma_{t-1}\boldsymbol{g}_{t,i_t}
}^2=
\frac{\kappa}{4}
\sigma_{t-1,i_t}^4.
\end{equation*}

For $0\le s\le1/2$,
\begin{equation*}
\frac{1}{\sqrt{1-s}}
\phi\!\FitParens{\frac{z}{\sqrt{1-s}}}
\ge
\phi(z)\exp\!\FitParens{-\frac{z^2s}{2(1-s)}}
\ge
\phi(z)e^{-z^2/2}.
\end{equation*}
Since the upper limit in \eqref{eq:cons-score} is less than $1$, retaining only the $i_t$ summand and integrating over the first half of its interval gives
\begin{equation*}
G_t(L)
\ge
\frac{\delta_t(L)e^{-z_{t-1,i_t}^2/2}}
{4r_LL^\top\Sigma_{t-1}L}
w_{t-1,i_t}
\FitParens{
\boldsymbol{g}_{t,i_t}^{\top}
\Sigma_{t-1}L
}^2.
\end{equation*}
The definition of $\delta_t(L)$ gives
\begin{equation*}
\frac{\delta_t(L)}{L^\top\Sigma_{t-1}L}
=
\frac{\beta_L^2}{1+\beta_L^2L^\top\Sigma_{t-1}L}
\frac{\phi(\alpha_t(L))^2}
{\Phi(\alpha_t(L))\Phi(-\alpha_t(L))}.
\end{equation*}
By condition \textup{(iii)}, the boundedness of $z_{t-1,i_t}$, and continuity and strict positivity of the expression above, there exists $\tau_*>0$ such that eventually
\begin{equation*}
\frac{\delta_t(L)e^{-z_{t-1,i_t}^2/2}}
{4r_LL^\top\Sigma_{t-1}L}
\ge
\tau_*
\qquad
(L\in\mathcal D).
\end{equation*}
Consequently,
\begin{equation*}
\sum_{L\in\mathcal D}G_t(L)
\ge
\tau_*w_{t-1,i_t}
\sum_{L\in\mathcal D}
\FitParens{
\boldsymbol{g}_{t,i_t}^{\top}
\Sigma_{t-1}L
}^2\ge
\frac{\tau_*w_*\kappa}{4}
\sigma_{t-1,i_t}^4.
\end{equation*}
Thus
\begin{equation*}
\liminf_{t\to\infty}
\sum_{L\in\mathcal D}G_t(L)
\ge
\frac{\tau_*w_*\kappa}{4}
c^2\delta^4
>0. \Halmoshere
\end{equation*}
\end{proof}
Finally, after establishing sufficient exploration, we connect learning on $\boldsymbol{1}^\perp$ to the terminal opportunity cost.
\begin{lemma}\label{lem:cons-projection}
For every $\boldsymbol{x}\in\R^I$,
\begin{equation*}
S_k^*(\boldsymbol{x})=S_k^*(P\boldsymbol{x}),
\qquad
\sum_{i\in S_k^*(\boldsymbol{x})}x_i
=
\sum_{i\in S_k^*(P\boldsymbol{x})}(P\boldsymbol{x})_i
+k\overline x,
\qquad
\overline x:=\frac{1}{I}\boldsymbol{1}^\top\boldsymbol{x}.
\end{equation*}
Consequently, if
\begin{equation*}
\E\|P\boldsymbol{\mu}_T-P\boldsymbol{\theta}\|_\infty\to0,
\end{equation*}
then $\E[\mathrm{OC}_T]\to0$.
\end{lemma}

\begin{proof}{Proof.}
Since $x_i=(P\boldsymbol{x})_i+\overline x$ for every $i$, any size-$k$ subset $A$ satisfies
\begin{equation*}
\sum_{i\in A}x_i
=
\sum_{i\in A}(P\boldsymbol{x})_i+k\overline x.
\end{equation*}
The term $k\overline x$ does not depend on $A$, so $S_k^*(\boldsymbol{x})=S_k^*(P\boldsymbol{x})$. Evaluating the sum at $A=S_k^*(\boldsymbol{x})$ gives
\begin{equation*}
\sum_{i\in S_k^*(\boldsymbol{x})}x_i
=
\sum_{i\in S_k^*(P\boldsymbol{x})}(P\boldsymbol{x})_i
+k\overline x.
\end{equation*}

For the opportunity-cost bound, note that $S_k^*(\boldsymbol{\mu}_T)=S_k^*(P\boldsymbol{\mu}_T)$ by the identity just proved. Next, note that the sum of the $k$ largest elements is $k$-Lipschitz in the L-infinity norm: for any $\boldsymbol{x},\boldsymbol{y}\in\R^I$,
\begin{equation*}
\sum_{i\in S_k^*(\boldsymbol{x})}x_i
-
\sum_{i\in S_k^*(\boldsymbol{y})}y_i
\le
\sum_{i\in S_k^*(\boldsymbol{x})}(x_i-y_i)
\le
k\|\boldsymbol{x}-\boldsymbol{y}\|_\infty.
\end{equation*}
Interchanging $\boldsymbol{x}$ and $\boldsymbol{y}$ gives
\begin{equation*}
\FitDelims{|}{|}{
\sum_{i\in S_k^*(\boldsymbol{x})}x_i
-
\sum_{i\in S_k^*(\boldsymbol{y})}y_i
}
\le
k\|\boldsymbol{x}-\boldsymbol{y}\|_\infty.
\end{equation*}
Thus,
\begin{equation*}
0\le \mathrm{OC}_T
=
\sum_{i\in S_k^*(P\boldsymbol{\theta})}(P\boldsymbol{\theta})_i
-\sum_{i\in S_k^*(P\boldsymbol{\mu}_T)}(P\boldsymbol{\mu}_T)_i
+\sum_{i\in S_k^*(P\boldsymbol{\mu}_T)}
\FitParens{(P\boldsymbol{\mu}_T)_i-(P\boldsymbol{\theta})_i}
\le
2k\|P\boldsymbol{\mu}_T-P\boldsymbol{\theta}\|_\infty.
\end{equation*}
Taking expectations gives the result.
\end{proof}

\subsection{Proof of \cref{thm:consistency}}\label{app:consistency:proof}
We now have the tools to prove \cref{thm:consistency}.

\begin{proof}{Proof.}
Define the set of directions sampled infinitely often, $E_\infty:=\{L\in\mathcal D:L\text{ is sampled infinitely often}\}.$ For each $\mathcal E\subseteq\mathcal D$, let $\Omega_{\mathcal E}:=\{E_\infty=\mathcal E\}$. Fix $\mathcal E\subseteq\mathcal D$ and work on $\Omega_{\mathcal E}$. Let
\begin{equation*}
M:=\operatorname{span}(\mathcal E),
\qquad
R_M:=\text{orthogonal projector onto $M$,}
\qquad
Q:=\operatorname{Id}_I-R_M.
\end{equation*}
We claim that on $\Omega_{\mathcal E}$, $\boldsymbol{1}^\perp\subseteq M$. Assume for contradiction that $\boldsymbol{1}^\perp\nsubseteq M$. Because directions outside $\mathcal E$ are sampled only finitely often and $\mathcal D\setminus\mathcal E$ is finite, there exists an almost-surely finite round $T^*$ such that after round $T^*$ no direction outside $\mathcal E$ is ever sampled again.

\medskip\noindent\textbf{Step 1: learning on the recurrent space.} \Cref{lem:cons-learning} applies on $\Omega_{\mathcal E}$. Hence for every $\boldsymbol{v}\in M$,
\begin{equation*}
\boldsymbol{v}^\top\boldsymbol{\mu}_{t-1}^*\to\boldsymbol{v}^\top\boldsymbol{\theta},
\qquad
\boldsymbol{v}^\top\Sigma_{t-1}^*\boldsymbol{v}\to0
\qquad\text{a.s. on $\Omega_{\mathcal E}$.}
\end{equation*}
Choose a basis $L_1,\ldots,L_d\in\mathcal E$ of $M$ and let $A=[L_1\ \cdots\ L_d]$. By Cauchy--Schwarz,
\begin{equation*}
|L_r^\top\Sigma_{t-1}^*L_s|
\le
\sqrt{L_r^\top\Sigma_{t-1}^*L_r}
\sqrt{L_s^\top\Sigma_{t-1}^*L_s}
\to0,
\end{equation*}
so $A^\top\Sigma_{t-1}^*A\to0$ entrywise and hence in operator norm. It follows that
\begin{equation*}
\|R_M\Sigma_{t-1}^*R_M\|_{\mathrm{op}}\to0,
\qquad
R_M\boldsymbol{\mu}_{t-1}^*\to R_M\boldsymbol{\theta}
\qquad\text{a.s. on $\Omega_{\mathcal E}$.}
\end{equation*}

\medskip\noindent\textbf{Step 2: approximate variance stays positive off the recurrent span.} \Cref{lem:cons-offspan} applies with the conclusions of Step 1. Thus, almost surely on $\Omega_{\mathcal E}$, there exists $\eta>0$ such that $Q\Sigma_{t-1}^*Q\succeq\eta Q$ for all sufficiently large $t$. By \cref{lem:cons-bl,lem:cons-approx},
\begin{equation}\label{eq:cons-main-blocks}
\|R_M\Sigma_{t-1}R_M\|_{\mathrm{op}}\to0,
\qquad
\|R_M\Sigma_{t-1}Q\|_{\mathrm{op}}+\|Q\Sigma_{t-1}R_M\|_{\mathrm{op}}\to0,
\qquad
Q\Sigma_{t-1}Q\succeq \chi_-\eta Q
\end{equation}
almost surely on $\Omega_{\mathcal E}$, where the last inequality holds for all sufficiently large $t$.

\medskip\noindent\textbf{Step 3: the sum of scores stays bounded away from zero.} The conditions of \cref{lem:cons-positive-score} follow from \eqref{eq:cons-main-blocks}, \cref{ass:update}(ii), and \cref{lem:cons-bl}. Therefore,
\begin{equation}\label{eq:cons-main-positive}
\liminf_{t\to\infty}
\sum_{L\in\mathcal D}G_t(L)>0
\qquad\text{a.s. on $\Omega_{\mathcal E}$.}
\end{equation}

\medskip\noindent\textbf{Step 4: every recurrent direction's score goes to $0$.} Fix $L\in\mathcal E$. Since $L\in M$, Step 1 gives $L^\top\Sigma_{t-1}^*L\to0$. \Cref{ass:update}(i) then yields
\begin{equation*}
L^\top\Sigma_{t-1}L
\le
\chi_+L^\top\Sigma_{t-1}^*L
\to0.
\end{equation*}
\Cref{lem:cons-recurrent-score} therefore implies
\begin{equation*}
G_t(L)\to0
\qquad\text{a.s. on $\Omega_{\mathcal E}$.}
\end{equation*}
Thus
\begin{equation}\label{eq:cons-main-recurrent}
\sum_{L\in\mathcal E}G_t(L)\to0
\qquad\text{a.s. on $\Omega_{\mathcal E}$.}
\end{equation}
\medskip\noindent\textbf{Step 5: contradiction with the chosen evaluation.} By \eqref{eq:cons-main-positive}, \eqref{eq:cons-main-recurrent},
\begin{equation*}
\liminf_{t\to\infty}
\sum_{L\in\mathcal D\setminus\mathcal E}G_t(L)>0
\qquad\text{a.s. on $\Omega_{\mathcal E}$.}
\end{equation*}
Therefore,
\begin{equation*}
\liminf_{t\to\infty}
\max_{L\in\mathcal D\setminus\mathcal E}G_t(L)>0,
\qquad
\max_{L\in\mathcal E}G_t(L)\to0.
\end{equation*}
Hence there exists an almost-surely finite random time $T_1$ such that, for every $t\ge T_1$,
\begin{equation*}
\max_{L\in\mathcal D\setminus\mathcal E}G_t(L)
>
\max_{L\in\mathcal E}G_t(L).
\end{equation*}
Since \textsc{SA} maximizes the score \eqref{eq:cons-score} over $\mathcal D$, the direction chosen at round $t$ cannot be in $\mathcal E$ for all sufficiently large $t$. But after round $T^*$, every sampled direction lies in $\mathcal E$. Contradiction. Therefore,
\begin{equation*}
\boldsymbol{1}^\perp\subseteq\operatorname{span}(E_\infty)
\qquad\text{almost surely}.
\end{equation*}

\medskip\noindent\textbf{Step 6: convergence.} It remains to show that the full history identifies $P\boldsymbol{\theta}$, after which \cref{lem:cons-projection} implies the conclusion of the theorem. Fix $\mathcal E\subseteq\mathcal D$. On $\Omega_{\mathcal E}$, for each $i=1,\ldots,I$, we have $P\boldsymbol{e}_i\in\boldsymbol{1}^\perp\subseteq M,$ so \cref{lem:cons-learning} implies that $\1_{\Omega_{\mathcal E}}(P\boldsymbol{e}_i)^\top\boldsymbol{\theta}$ is $\mathcal H_\infty$-measurable. Since $(P\boldsymbol{\theta})_i = \boldsymbol{e}_i^\top P\boldsymbol{\theta} = (P\boldsymbol{e}_i)^\top\boldsymbol{\theta},$ it follows that $\1_{\Omega_{\mathcal E}}P\boldsymbol{\theta}$ is $\mathcal H_\infty$-measurable. Since $P\boldsymbol{\theta} = \sum_{\mathcal E\subseteq\mathcal D} \1_{\Omega_{\mathcal E}}P\boldsymbol{\theta} $ a.s., $P\boldsymbol{\theta}$ is $\mathcal H_\infty$-measurable.

Therefore, for each $i=1,\ldots,I$,
\begin{equation*}
(P\boldsymbol{\mu}_{t-1}^*)_i
=
\E[(P\boldsymbol{\theta})_i\mid\mathcal H_t]
\to
(P\boldsymbol{\theta})_i
\qquad\text{a.s. and in $L^1$,}
\end{equation*}
by Doob's theorem. Hence
\begin{equation*}
\E\|P\boldsymbol{\mu}_{t-1}^*-P\boldsymbol{\theta}\|_\infty
\le
\sum_{i=1}^I
\E\FitDelims{|}{|}{(P\boldsymbol{\mu}_{t-1}^*-P\boldsymbol{\theta})_i}
\to0.
\end{equation*}
This establishes convergence for the exact posterior mean along the run without a budget constraint. We now return to the procedure with budget $R$ and show that $\E\|P\boldsymbol{\mu}_T-P\boldsymbol{\theta}\|_\infty\to0$ as $R\to\infty$. For a run with budget $R$, take $t=\lfloor R/\max(r_0,r_1)\rfloor$ so that for the first $t$ evaluations this is equivalent to the unbudgeted run considered above. Its terminal history therefore contains the history through evaluation $t$. Jensen's inequality and the triangle inequality give
\begin{align*}
\E\|P\boldsymbol{\mu}_T^*-P\boldsymbol{\theta}\|_\infty
&\le
\E\|P\boldsymbol{\mu}_T^*-P\boldsymbol{\mu}_t^*\|_\infty
+
\E\|P\boldsymbol{\mu}_t^*-P\boldsymbol{\theta}\|_\infty\\
&=
\E\FitDelims{\|}{\|}{
\E\!\FitDelims{[}{]}{
P\boldsymbol{\theta}-P\boldsymbol{\mu}_t^*
\,\FitMid{|}\,
\mathcal H_{T+1}
}
}_\infty
+
\E\|P\boldsymbol{\mu}_t^*-P\boldsymbol{\theta}\|_\infty\\
&\le
2\E\|P\boldsymbol{\mu}_t^*-P\boldsymbol{\theta}\|_\infty
\to0
\qquad\text{as $R\to\infty$.}
\end{align*}
Also, for any $\boldsymbol{z}\in\R^I$, $\|P\boldsymbol{z}\|_\infty\le2\|\boldsymbol{z}\|_\infty$. Thus, by \cref{ass:update}(iii),
\begin{equation*}
\E\|P\boldsymbol{\mu}_T-P\boldsymbol{\theta}\|_\infty
\le
\E\|P\boldsymbol{\mu}_T-P\boldsymbol{\mu}_T^*\|_\infty
+
\E\|P\boldsymbol{\mu}_T^*-P\boldsymbol{\theta}\|_\infty
\le
2\E\|\boldsymbol{\mu}_T-\boldsymbol{\mu}_T^*\|_\infty
+
\E\|P\boldsymbol{\mu}_T^*-P\boldsymbol{\theta}\|_\infty
\to0.
\end{equation*}
\Cref{lem:cons-projection} gives
\begin{equation*}
\E[\mathrm{OC}_T]\to0
\qquad\text{as $R\to\infty$.} \Halmoshere
\end{equation*}
\end{proof}

\section{Supplement to \texorpdfstring{\cref{sec:numerics}}{Section 7}}\label{app:numerics}

The figures below are the counterparts of \Cref{fig:alloc}, \Cref{fig:endgame}a, and \Cref{fig:kgtraj} for all other configurations.

\subsection{Allocations over the budget}\label{app:numerics:alloc}

\begin{figure}[!htb]
\centering
\includegraphics[width=0.9\textwidth]{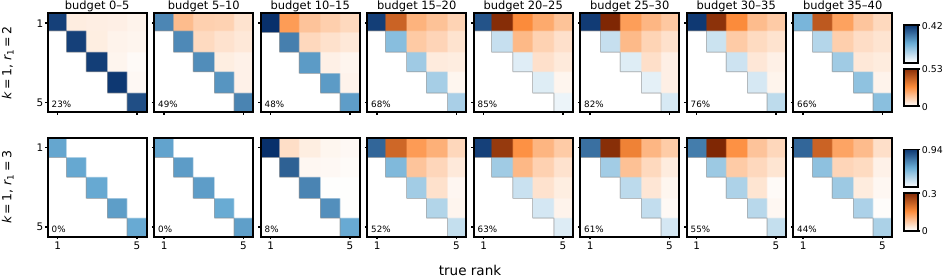}
\caption{Mean allocation of \textsc{SA} by true rank in each eighth of the budget, all $I=10$ configurations.}
\label{fig:app-alloc-10}
\end{figure}

\begin{figure}[!htb]
\centering
\includegraphics[width=0.9\textwidth]{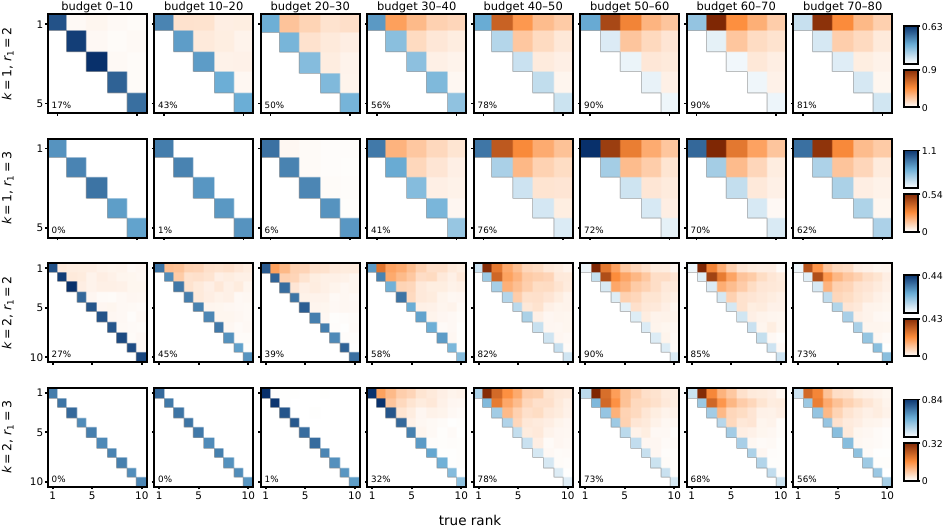}
\caption{Same as \Cref{fig:app-alloc-10} for $I=20$.}
\label{fig:app-alloc-20}
\end{figure}

\begin{figure}[!htb]
\centering
\includegraphics[width=0.9\textwidth]{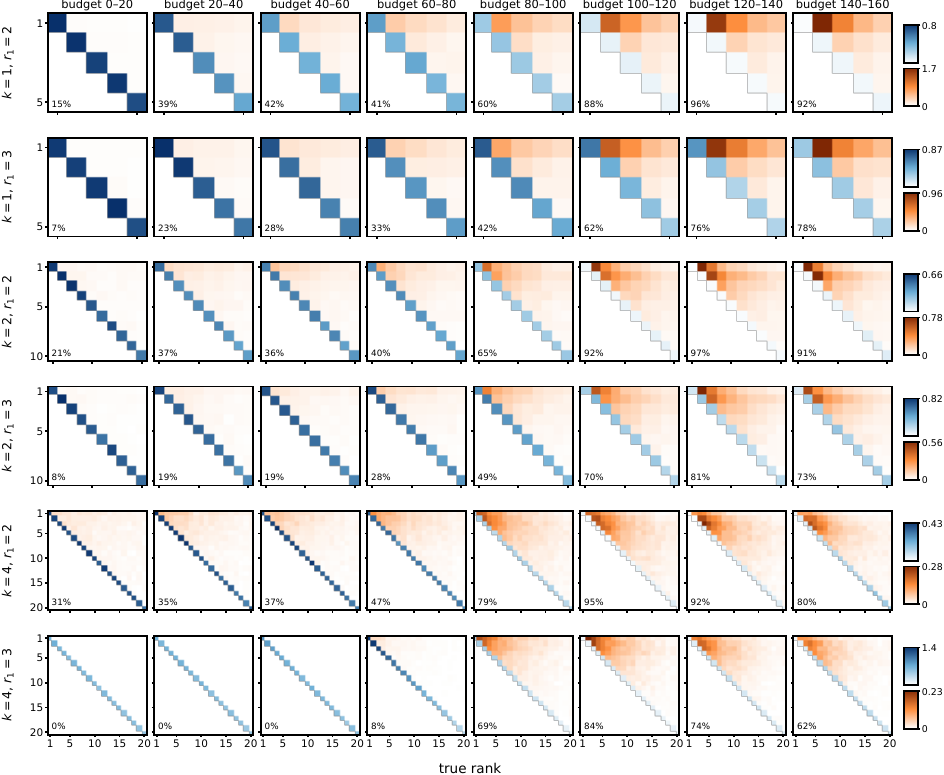}
\caption{Same as \Cref{fig:app-alloc-10} for $I=40$.}
\label{fig:app-alloc-40}
\end{figure}

\begin{figure}[!htb]
\centering
\includegraphics[width=0.9\textwidth]{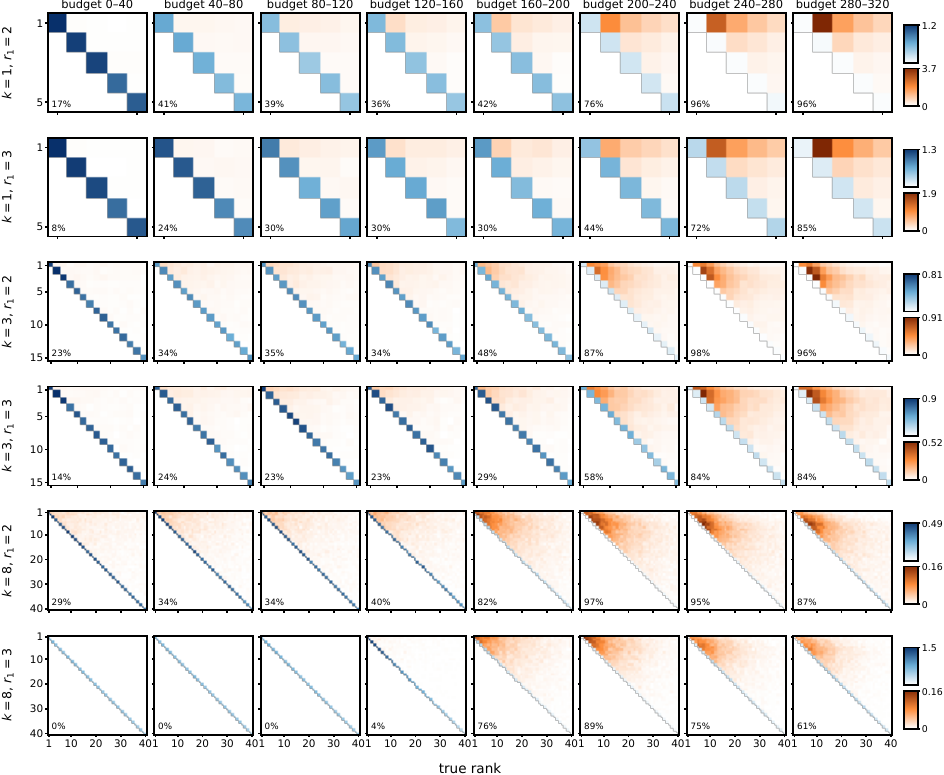}
\caption{Same as \Cref{fig:app-alloc-10} for $I=80$.}
\label{fig:app-alloc-80}
\end{figure}

\begin{figure}[!htb]
\centering
\includegraphics[width=0.9\textwidth]{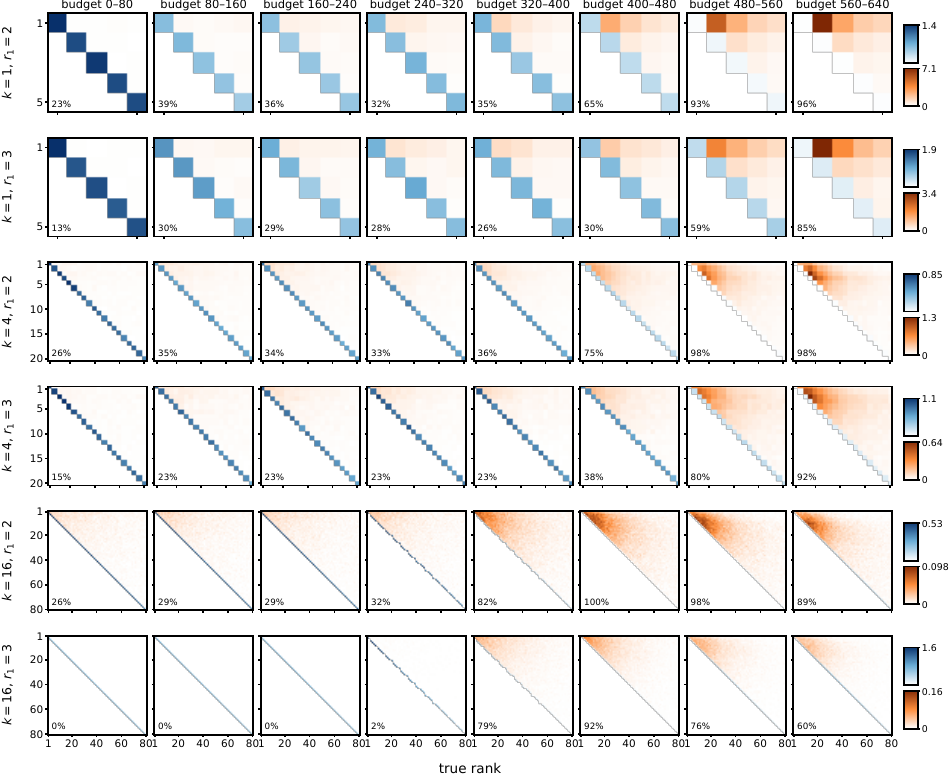}
\caption{Same as \Cref{fig:app-alloc-10} for $I=160$.}
\label{fig:app-alloc-160}
\end{figure}

\clearpage
\subsection{Allocation of \textsc{SA} versus \textsc{KG}}\label{app:numerics:endgame}

\begin{figure}[!htb]
\centering
\includegraphics[width=0.9\textwidth]{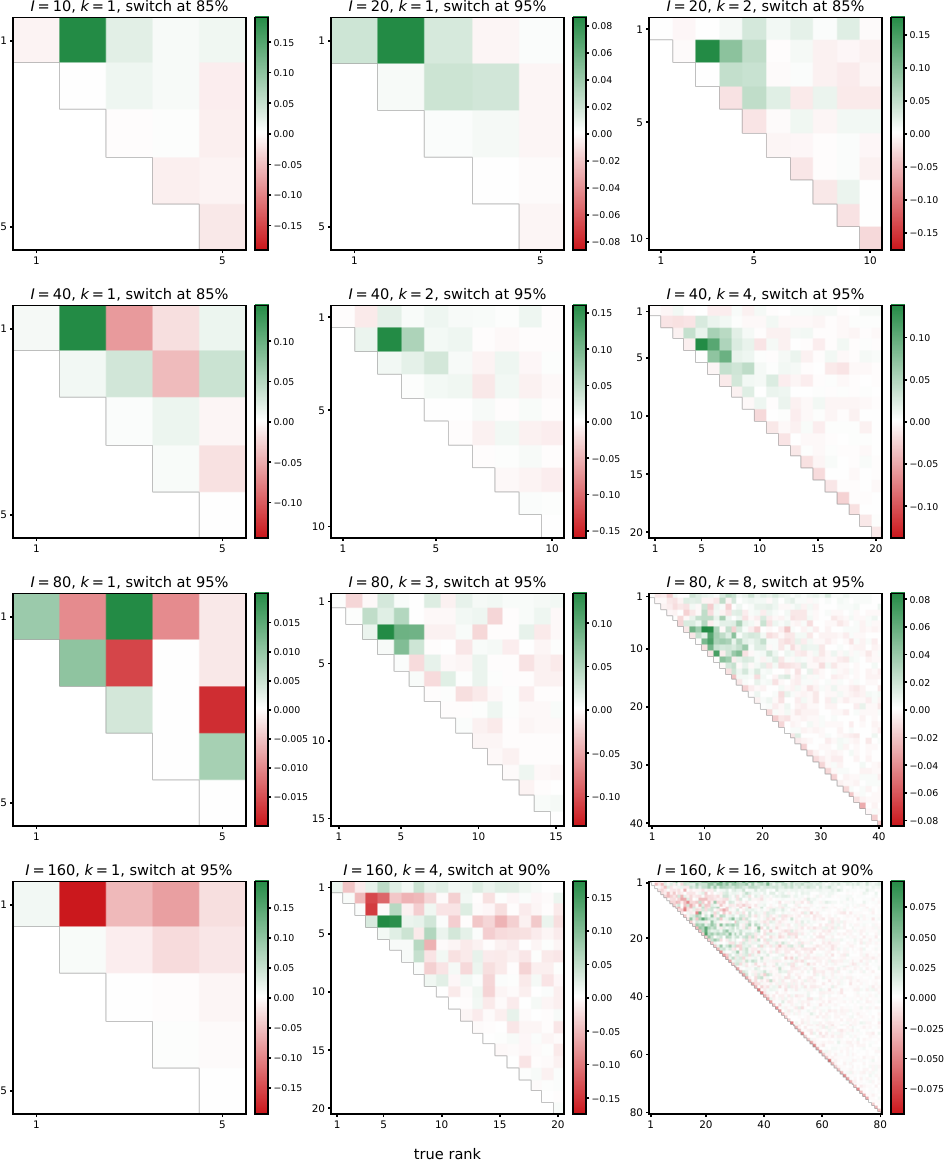}
\caption{Average allocation difference after the switch, SE minus \textsc{SA}, in cost units, all configurations at $r_1=2$. Green marks where SE spends more, red where it spends less.}
\label{fig:app-endgame-r2}
\end{figure}

\begin{figure}[!htb]
\centering
\includegraphics[width=0.9\textwidth]{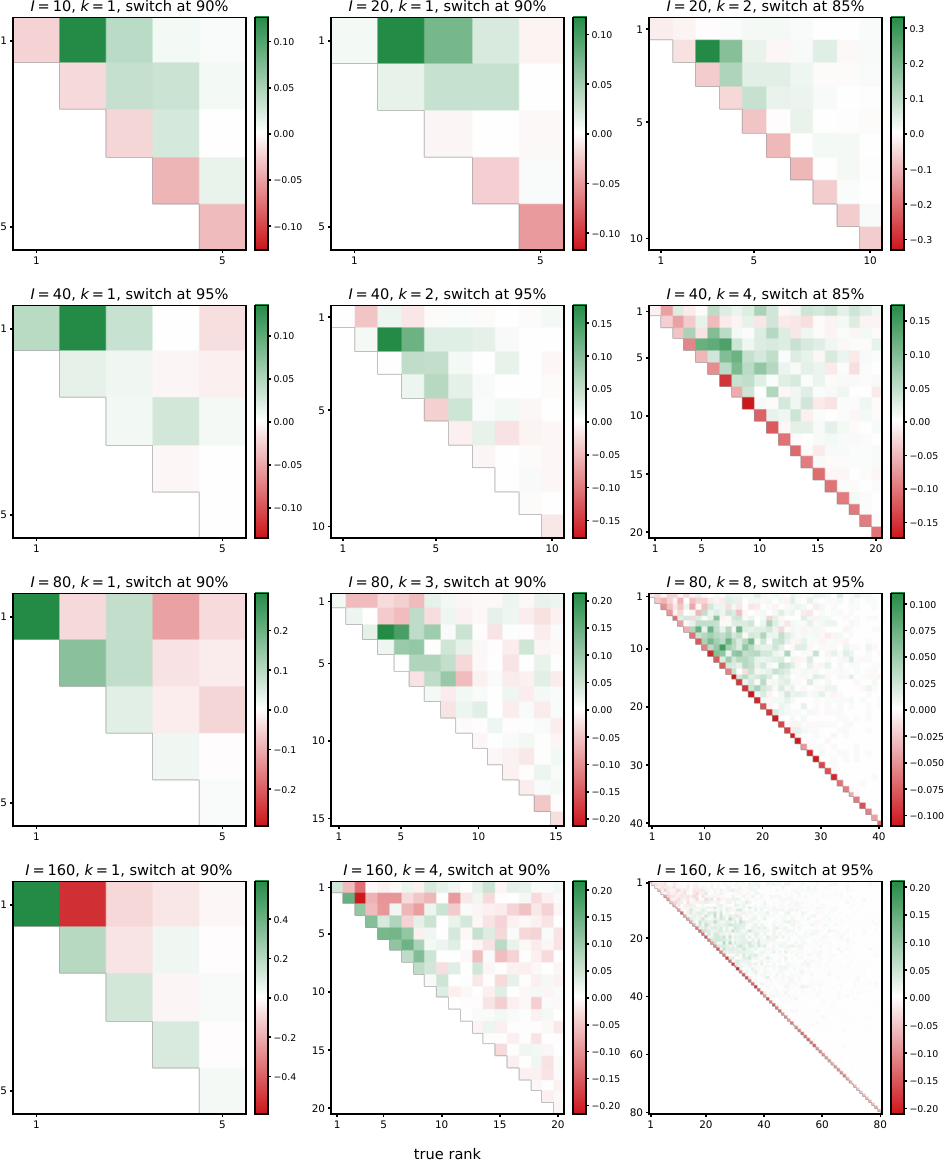}
\caption{Same as \Cref{fig:app-endgame-r2} for $r_1=3$.}
\label{fig:app-endgame-r3}
\end{figure}

\clearpage
\subsection{Comparison to external benchmarks}\label{app:numerics:external}

\begin{figure}[!htb]
\centering
\includegraphics[width=0.9\textwidth]{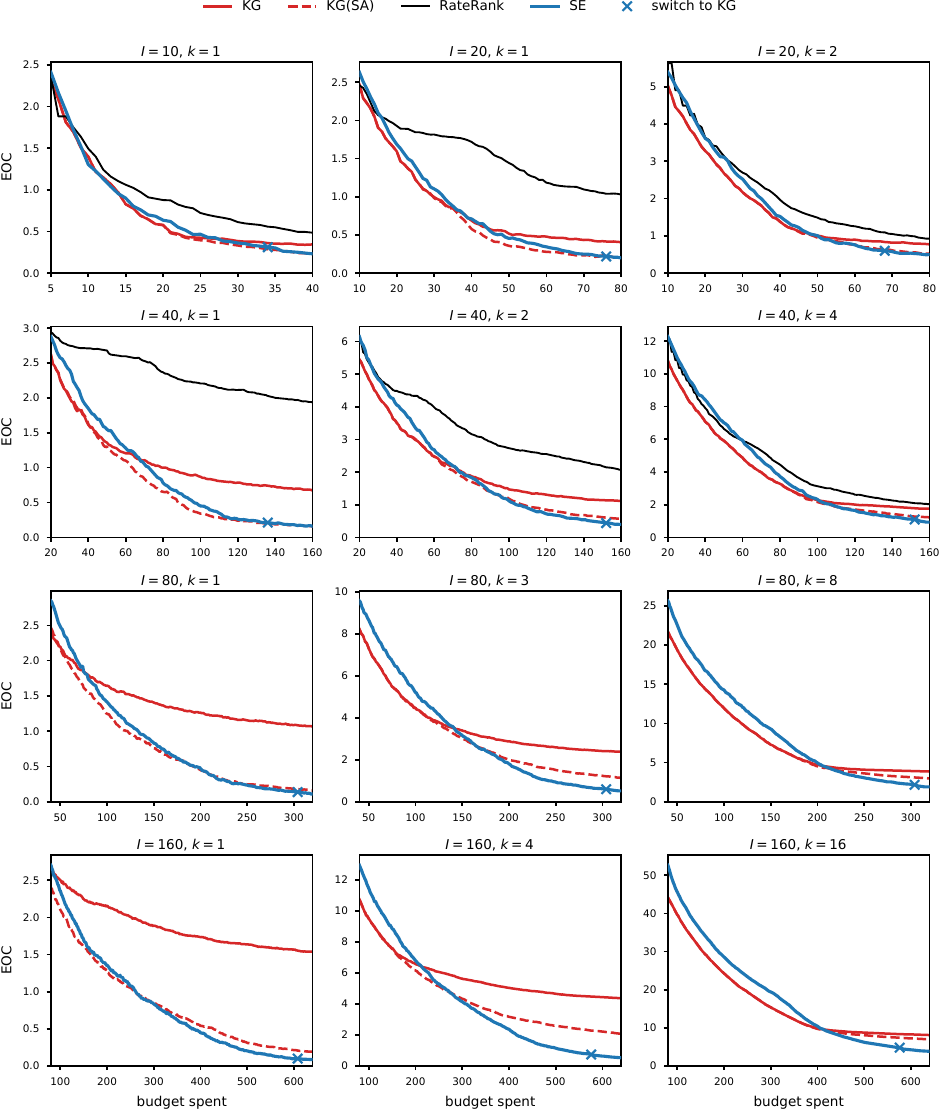}
\caption{EOC against the budget spent, all configurations at $r_1=2$.}
\label{fig:app-kgtraj-r2}
\end{figure}

\begin{figure}[!htb]
\centering
\includegraphics[width=0.9\textwidth]{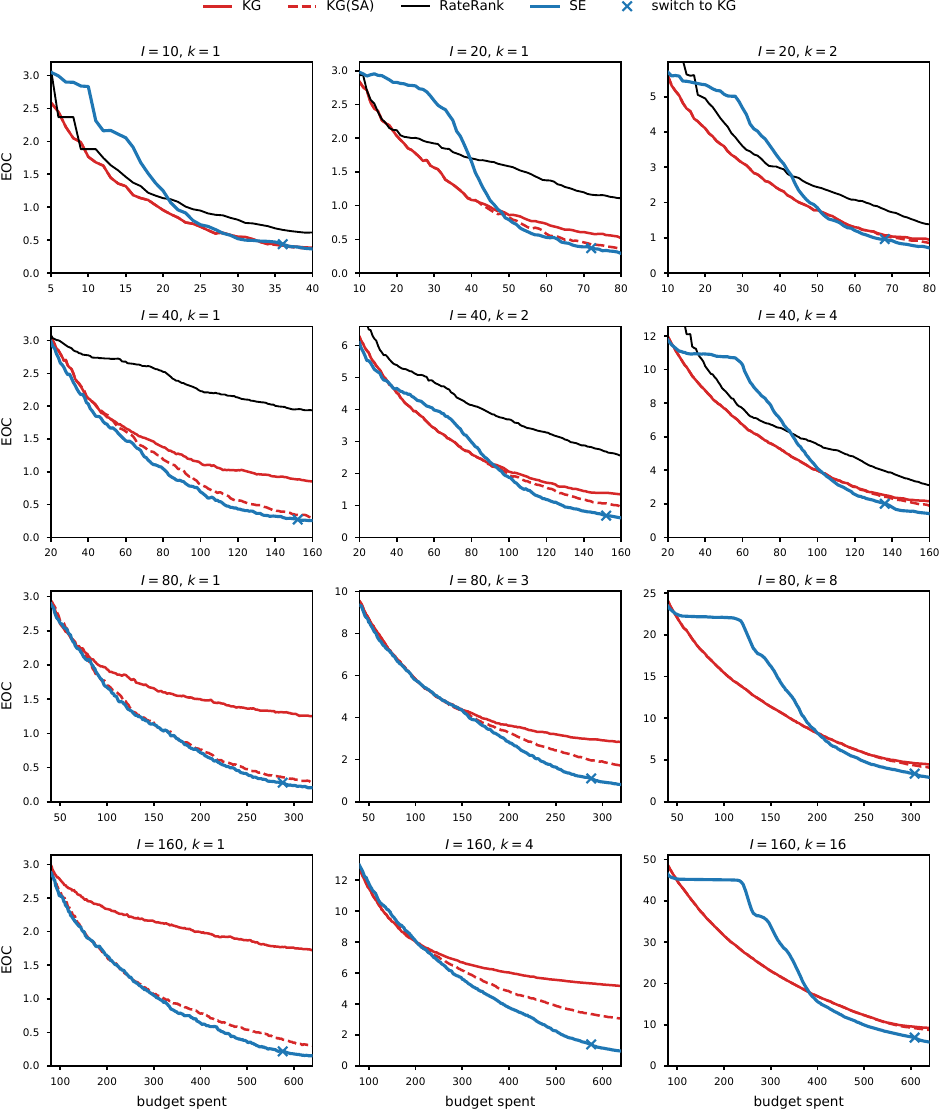}
\caption{EOC against the budget spent, all configurations at $r_1=3$.}
\label{fig:app-kgtraj-r3}
\end{figure}

\end{APPENDICES}

\end{document}

%% file: figs/table1_robust.tex
\begin{tabular}{lrcrcrc}
\toprule
misspecification &
\multicolumn{2}{c}{$I=20$, $k=2$}  &
\multicolumn{2}{c}{$I=80$, $k=3$}  &
\multicolumn{2}{c}{$I=160$, $k=4$} \\
\midrule
heterogeneous accuracies & $+3.7\%$  & $(5.0)$ & $+16.7\%$ & $(4.3)$ & $+14.8\%$ & $(3.8)$ \\
heterogeneous $\theta_0$ & $+2.5\%$  & $(4.6)$ & $-6.6\%$  & $(3.4)$ & $+1.7\%$  & $(3.4)$ \\
true accuracies 65/75    & $-3.1\%$  & $(2.6)$ & $-8.9\%$  & $(2.2)$ & $-4.6\%$  & $(1.8)$ \\
true accuracies 85/90    & $+5.3\%$  & $(6.6)$ & $+41.2\%$ & $(6.1)$ & $+73.7\%$ & $(6.5)$ \\
true $\theta_0 = +1.58$  & $+3.2\%$  & $(3.7)$ & $-0.4\%$  & $(3.6)$ & $+1.5\%$  & $(3.2)$ \\
true $\theta_0 = -1.58$  & $-1.2\%$  & $(4.0)$ & $+26.9\%$ & $(4.3)$ & $+21.5\%$ & $(3.5)$ \\
\bottomrule
\end{tabular}

%% file: figs/table2_allocation_85_90.tex
\begin{tabular}{lcccccccc}
\toprule
eighth of the budget & 1 & 2 & 3 & 4 & 5 & 6 & 7 & 8 \\
\midrule
SE & $22\%$ & $31\%$ & $32\%$ & $31\%$ & $30\%$ & $34\%$ & $47\%$ & $73\%$ \\
unbiased SE & $0\%$ & $0\%$ & $0\%$ & $1\%$ & $52\%$ & $99\%$ & $98\%$ & $94\%$ \\
\bottomrule
\end{tabular}